\documentclass[nohyperref]{article}

\usepackage{microtype}
\usepackage{subcaption}
\usepackage{booktabs} % for professional tables
\usepackage{xspace}
\usepackage{mwe}
\usepackage{multirow}
\usepackage{array}
\usepackage{makecell}
\usepackage{dblfloatfix}
\usepackage{hyperref}
\usepackage{tikz}

\usepackage{afterpage}

\usepackage[accepted]{icml2023_cr}
\usepackage{xcolor}
\usepackage{amsmath}
\usepackage{amssymb}
\usepackage{mathtools}
\usepackage{amsthm}
\usepackage{bm}

\usepackage{enumitem}

\usepackage{graphicx}

\usepackage[capitalize,noabbrev]{cleveref}

\theoremstyle{plain}

\theoremstyle{definition}

\theoremstyle{remark}

\DeclareMathOperator*{\argmin}{argmin}

\definecolor{col_human}{RGB}{148, 149, 152}
\definecolor{col_VAENS}{RGB}{12, 41, 255}
\definecolor{col_VAESTN}{RGB}{50, 191, 255}
\definecolor{col_DAGANUN}{RGB}{237, 139, 137}
\definecolor{col_DAGANRN}{RGB}{253, 217, 0}
\definecolor{col_CFGDM}{RGB}{77, 209, 21}
\definecolor{col_DDPM}{RGB}{167, 209, 12}
\definecolor{col_FSDM}{RGB}{155, 214, 179}

\newcommand{\VAENS}{\textbf{\textcolor{col_VAENS}{VAE-NS}}}
\newcommand{\VAESTN}{\textbf{\textcolor{col_VAESTN}{VAE-STN}}}
\newcommand{\DAGANUN}{\textbf{\textcolor{col_DAGANUN}{DA-GAN-UN}}}
\newcommand{\DAGANRN}{\textbf{\textcolor{col_DAGANRN}{DA-GAN-RN}}}
\newcommand{\human}{\textbf{\textcolor{black}{human}}}
\newcommand{\humans}{\textbf{\textcolor{black}{humans}}}
\newcommand{\Humans}{\textbf{\textcolor{black}{Humans}}}
\newcommand{\CFGDM}{\textbf{\textcolor{col_CFGDM}{CFGDM}}}
\newcommand{\DDPM}{\textbf{\textcolor{col_DDPM}{DDPM}}}
\newcommand{\FSDM}{\textbf{\textcolor{col_FSDM}{FSDM}}}

 \newcommand{\mquad}{\hspace{0.35em}}

\def\RevisionMode{0} 

 \if\RevisionMode1
 \newcommand{\revision}[1]{{\color{red}#1}}
 \else
 \newcommand{\revision}[1]{{\color{black}#1}}
 \fi
\usepackage[textsize=tiny]{todonotes}

\icmltitlerunning{Diffusion Models as Artists: Are we Closing the Gap between Humans and Machines?}

\begin{document}

\twocolumn[
% \icmltitle{Humans, unlike diffusion models, produce creative yet recognizable drawings}
%\icmltitle{Diffusion models vs humans: Who is the best one-shot drawer?}
%\icmltitle{Diffusion models as sketch artists: \\ How close are we to closing the gap between humans and machines?}
\icmltitle{Diffusion Models as Artists: \\ Are we Closing the Gap between Humans and Machines?}



% You can specify symbols, otherwise, they are numbered in order.
\icmlsetsymbol{equal}{*}

\begin{icmlauthorlist}
\icmlauthor{Victor Boutin}{aniti,brown}
\icmlauthor{Thomas Fel}{aniti,brown}
\icmlauthor{Lakshya Singhal}{brown}
\icmlauthor{Rishav Mukherji}{brown}
\icmlauthor{Akash  Nagaraj}{brown}
\icmlauthor{Julien Colin}{brown,ellisalicante}
\icmlauthor{Thomas Serre}{aniti,brown}
\end{icmlauthorlist}

\icmlaffiliation{aniti}{Artificial and Natural Intelligence Toulouse Institute,  Universit\'e de Toulouse, Toulouse, France.}
\icmlaffiliation{brown}{Carney Institute for Brain Science, Dpt. of Cognitive Linguistic \& Psychological Sciences, Brown University, Providence, RI 02912.}
\icmlaffiliation{ellisalicante}{ELLIS Alicante, Spain}
%\icmlaffiliation{sch}{School of ZZZ, Institute of WWW, Location, Country}

  %\textsuperscript{1} Artificial and Natural Intelligence Toulouse Institute, Universit\'e de Toulouse, France\\
   %\textsuperscript{2} Carney Institute for Brain Science, Dpt. of Cognitive Linguistic \& Psychological Sciences \\ 
   %Brown University, Providence, RI 02912\\
   %\texttt{\{victor\_boutin, thomas\_serre\}@brown.edu}\\
   
\icmlcorrespondingauthor{Victor Boutin}{victor\_boutin@brown.edu}
\icmlcorrespondingauthor{Thomas Serre}{thomas\_serre@brown.edu}

\icmlkeywords{Neuroscience, one-shot generative models, diffusion models, human creativity, diversity vs. accuracy trade-off, human drawings, human-machine comparison}

\vskip 0.3in
]



\printAffiliationsAndNotice{}  % leave blank if no need to mention equal contribution

\begin{abstract}

An important milestone for AI is the development of algorithms that can produce drawings that are indistinguishable from those of humans. %In particular, deep diffusion models are now able to generate near photo-realistic natural images. Are these algorithms also able to generate human-like drawings? 
% Are we closing the gap between humans and machines? 
Here, we adapt the ``diversity vs. recognizability'' scoring framework from \citealp{boutin2022diversity} and find that one-shot diffusion models have indeed started to close the gap between humans and machines.
%Here, we adapt the ``diversity vs. recognizability'' scoring framework from \citealp{boutin2022diversity} to compare the quality of drawings produced by one-shot generative models against those produced by human observers. We find that, compared to prior deep generative architectures, diffusion models have indeed started to close the gap between humans and machines. 
%However, with further refinement of the scoring metric to incorporate a finer measure of the originality of individual samples, we show that qualitative differences between state-of-the-art models and humans persist. Interestingly, strengthening the guidance of diffusion models helps their drawings look more human-like, but they still fall short of approximating the originality and recognizability of human drawings. 
However, using a finer-grained measure of the originality of individual samples, we show that strengthening the guidance of diffusion models helps improve the humanness of their drawings, but they still fall short of approximating the originality and recognizability of human drawings. 
%Using an online psychophysics experiment, we compare humans and diffusion models category diagnostic features, and we show that humans rely on fewer but more localized features than diffusion models. 
Comparing human category diagnostic features, collected through an online psychophysics experiment, against those derived from diffusion models reveals that humans rely on fewer and more localized features. 
%We further conduct an online psychophysics experiment to identify the most important image features for individual samples to be recognizable. Comparing these human-derived importance maps against those derived from diffusion models reveals remarkable differences. We suggest that these differences in visual strategies may be part of the reason for the remaining gap between machines and humans.
%At the sample level
%Using the class-level measures 
%Our results suggest that (i) diffusion models have started to close the gap between humans and machines; (ii) strengthening the conditioning signal helps improve the humanness of their drawings; but (iii) they still fall short of approximating the inventiveness and recognizability of human drawings, and (iv) qualitative differences between humans and machines may in part be explained by the fact that they emphasize different object features.
Overall, our study suggests that diffusion models have significantly helped improve the quality of machine-generated drawings; however, a gap between humans and machines remains -- in part explainable by discrepancies in visual strategies.
 
\end{abstract}

\section{Introduction}
%% Drawings : an open window on the human mind
Drawing is a fundamental human skill; from paintings on cave walls to the finest pieces of art, generations of humans have expressed their creative skills and imagination through drawings~\cite{donald1991origins}. Drawings are so deeply rooted in human cognition that they are routinely used in a variety of clinical settings -- from evaluating emotional trauma~\cite{koppitz1968psychological} and developmental disorders~\cite{ryan2001use} to intellectual deficits~\cite{goodenough1926measurement}. Cognitive psychologists and computer scientists have also used drawing tasks to probe the human ability to learn new visual concepts from just a single example~\cite{feldman1997structure,lake2015human}. From a computational perspective, such one-shot drawing tasks offer an unprecedented challenge -- to solve the seemingly impossible task of estimating entire probability distributions for novel image categories from a unique sample. Nevertheless, humans can effortlessly produce drawings that are original (i.e., sufficiently different from the shown exemplar), yet, easily recognizable~\cite{tiedemann2022one} -- suggesting strong inductive biases~\cite{tenenbaum1999bayesian, ullman2020bayesian} that are yet to be discovered.

\begin{figure}[t!]
%\vskip -0.1in
\begin{center}
\centerline{
\includegraphics[width=\columnwidth]{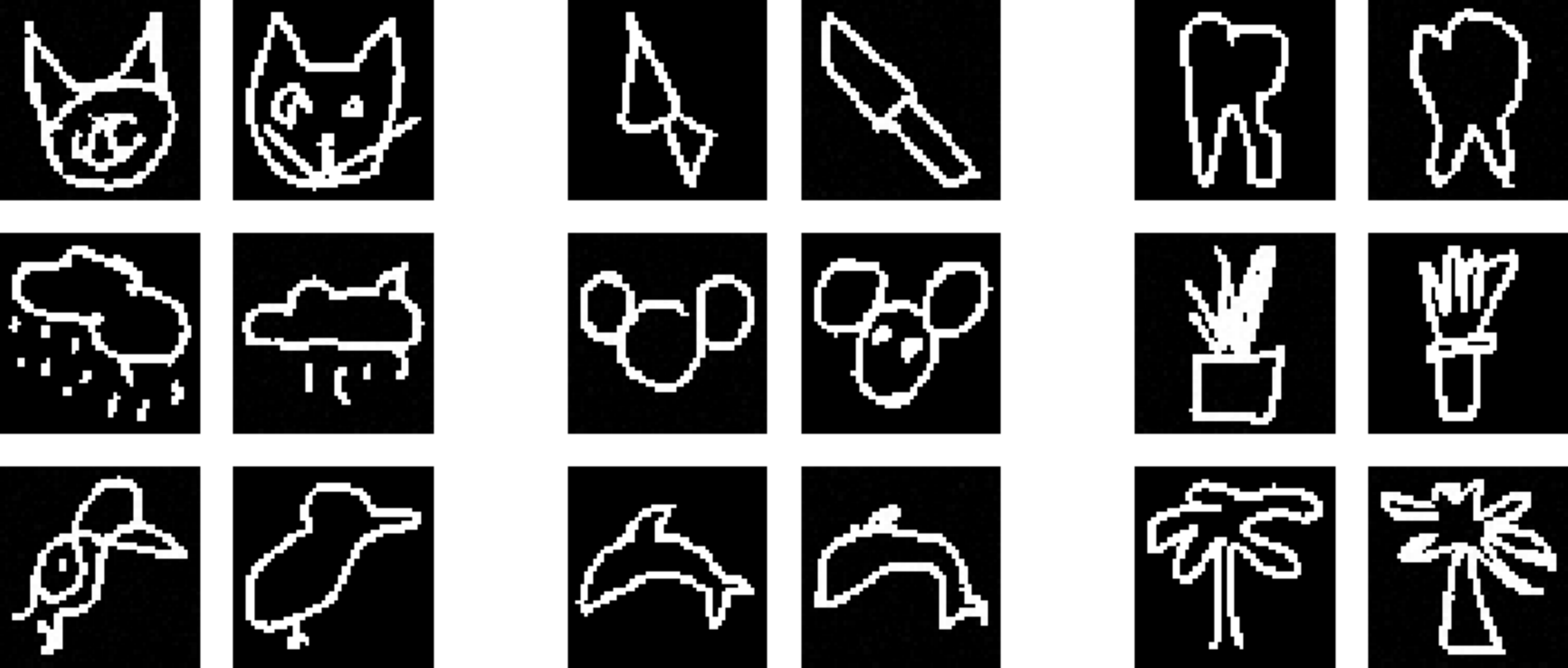}
}
\caption{ \setcounter{footnote}{0}
Can you tell apart human from machine-generated samples in each pair\footnotemark? Machine samples are generated using a \CFGDM{} (see \cref{related:one_shot_models}). }
\label{fig:fig0}
\end{center}
\vskip -0.4in
\end{figure}
\footnotetext{For each pair the human samples are (by row): right - left - left; right - right - left; right - right - left.}
%% One shot generation models
%While a common criticism of modern AI approaches is their reliance on large training datasets, significant progress has been made in the few-shot image generation. Generative models based on Variational Auto-Encoders (VAE) have exploited inductive biases such as spatial attention~\cite{rezende2016one} or context integration~\cite{edwards2016towards, giannone2021hierarchical} to generate novel variations from a single visual exemplar. Another approach involves conditioning the latent space of a Generative Adversarial Network (GANs) with a compressed representation of the exemplar to generate new samples belonging to the same category~\cite{antoniou2017data}. The recent breakthrough of diffusion models~\cite{song2019generative, sohl2015deep} probably offers a most promising avenue for one-shot modeling. Whether conditioned on a context vector~\cite{giannone2021hierarchical} or directly guided by the exemplar~\cite{ho2022classifier}, diffusion models exhibit state-of-the-art performance in one-shot image generation (see~\cref{related:one_shot_models} for more details on one-shot diffusion models).

While a common criticism of modern AI approaches is their appetite for large training datasets, significant progress has been made in the field of few-shot image generation. In particular, one-shot generative models based on Generative Adversarial Networks (GANs) or Variational Auto-Encoders (VAEs) have started to take advantage of various inductive biases %including in the form 
via specific forms of spatial attention~\cite{rezende2016one} or 
%contextual associations
the integration of contextual information~\cite{edwards2016towards, giannone2021hierarchical, antoniou2017data}. Furthermore, the recent breakthrough achieved using diffusion models~\cite{song2019generative, sohl2015deep} makes them a particularly promising class of models for one-shot generation. Clever methods for conditioning on a context vector~\cite{giannone2021hierarchical} or using direct guidance from the exemplar~\cite{ho2022classifier} has led to diffusion models that can produce near photo-realistic samples -- consistently outperforming the state-of-the-art in one-shot image generation (see~\cref{related:one_shot_models} for more details on one-shot diffusion models). %Nevertheless, since humans cannot produce photo-realistic images, such one-shot generative models need to be trained on more human-compatible datasets to allow for a fair comparison.

The ability of diffusion models to synthesize photo-realistic images speaks to their great expressivity -- which begs the question: \revision{``Can diffusion algorithms also serve as effective models of the human creative process for sketch drawing task?". Addressing this query necessitates a departure from photo-realistic image generation, as humans are incapable of producing such images. Instead, retraining generative models on more human-compatible datasets becomes imperative. Therefore, we consider the one-shot drawing task as it offers a leveled playing field for both humans and machines.}

%``Are diffusion algorithms good models of the human creative process?''. 
%To answer this question, we focus on one-shot drawing tasks as they offer a seemingly leveled playing field to compare humans and machines. Shown in~\cref{fig:fig0} are drawings produced by humans and a diffusion model. Are human and machine samples distinguishable or are diffusion models already able to produce human-like drawings?

%We argue that the metrics and datasets used by current human-machine comparison frameworks are not adequate to answer these questions for two main reasons (see \cref{related:metric_framework} for more details). First, these comparisons are often made on datasets that do not reflect the human ability to produce original samples. Second, as the current metrics tend to aggregate scores for entire classes, they do not have the required granularity to evaluate the quality of individual drawings. We propose a new dataset to address the first limitation and a more fine-grained human-machine comparison framework to overcome the second limitation.
%First, the comparison is often made on Omniglot~\cite{lake2015human}, a dataset that does not reflect the human ability to produce original drawings. Second, the comparison metrics tend to aggregate samples regardless of their originality. These limitations urge us to propose a new experimental protocol to draw a more fair comparison between humans and machines in the one-shot drawing scenario.

%contribution
In this article, we introduce QuickDraw-FewShot (QuickDraw-FS), a dataset built on the \textit{Quick, Draw!} challenge~\cite{jongejan2016quick}, specifically designed for the few-shot image generation challenge. Building on the ``diversity vs. recognizability'' scoring framework by \citealp{boutin2022diversity}, we systematically compare humans with one-shot generation algorithms based on VAEs, GANs, and diffusion models. We show that diffusion models provide a better approximation of human drawing ability compared to VAEs and GANs. \revision{We further introduce the \textit{originality} metric to measure the similarity between an individual sample and the corresponding exemplar. For a given sample, the originality score is computed as its $\ell_2$ distance to the exemplar in the SimCLR latent space.}
%In this article, we introduce QuickDraw-FewShot (QuickDraw-FS), a dataset built on the \textit{Quick, Draw!} challenge~\cite{jongejan2016quick}, and specifically designed for the few-shot image generation challenge. At the level of object categories, we adapt the diversity vs recognizability framework by \citealp{boutin2022diversity} to systematically compare humans with one-shot generation algorithms based on VAEs, GANs, and diffusion models. We found that diffusion models provide a better approximation of human behavior compared to VAEs and GANs. We further introduce an \textit{originality} metric as the distance between a sample and the training exemplar. 
We analyze the evolution of samples' recognizability as a function of their originality using \textit{generalization curves}. Our results suggest that strengthening the guidance of diffusion models helps improve the humanness of their drawings, but they still fall short of approximating the originality and recognizability of human drawings. We further conduct an online psychophysics experiment to identify the most important image features for individual samples to be recognizable. Comparing these human-derived importance maps against those derived from diffusion models reveals a remarkable difference: humans tend to rely on fewer and more localized features than diffusion models. We suggest that these differences in visual strategies may be part of the reason for the remaining gap between machines and humans.

%and (iii) qualitative differences between humans and machines may in part be explained by different attentional strategies.

\section{Related Works}
\subsection{Humans-Machines Comparison Frameworks}
\label{related:metric_framework}

Researchers have historically used the Omniglot dataset to compare the generalization abilities of humans and machines on drawing tasks~\cite{lake2019omniglot}. To collect the Omniglot samples, human participants were presented with exemplars of novel handwritten characters and asked to reproduce them as accurately as possible (see Appendix 1 in \cite{lake2015human}). A limitation of this experimental protocol is that it hinders human creativity by prompting subjects to literally copy the exemplar. In \cref{dataset_comp}, we confirm this limitation by comparing the distributions of intra-class variability for the Omniglot and our proposed dataset QuickDraw-FS (see \cref{methods:dataset} for more details on the datasets). Another limitation of the Omniglot dataset is that it contains a reduced number of samples per category ($n=20$ samples). This prevents us from performing intra-class analyses that are statistically meaningful.

Different methods have been proposed to compare humans and machines in the one-shot generation task. \citealp{lake2015human} use the visual Turing test in which participants are asked to distinguish human drawings from those generated by the models (similar to \cref{fig:fig0}). Another approach consists in asking human participants or classifiers~\cite{lake2015human} to evaluate the recognizability of individual samples. However, neither of these methods helps quantify the degree to which models are able to produce samples that are sufficiently diverse -- at least compared to humans. More recently, a ``diversity vs. recognizability'' framework was proposed to circumvent this limitation via an additional critic network to evaluate samples' intra-class variability~\cite{boutin2022diversity}. This diversity metric provides a single measure for an entire class, and hence, it does not provide the necessary granularity needed to evaluate individual samples. Here, we refine this diversity metric to systematically compare the originality of individual drawings produced by humans and models~\cite{tiedemann2022one}.

\subsection{One-Shot Generative Models}
\label{related:one_shot_models}
The one-shot image generation task involves synthesizing variations of a visual concept that has not been seen during training. Let $\mathbf{x} \in \mathbb{R}^D$ be the image data to model, and $\mathbf{y} \in \mathbb{R}^D$ be the exemplar. Mathematically, the task involves learning the conditional probability distribution $p(\mathbf{x}|\mathbf{y})$. Herein, we mainly focus on diffusion models to learn $p(\mathbf{x}|\mathbf{y})$~\cite{song2019generative, sohl2015deep}, but VAEs~\cite{kingma2013auto} or GANs~\cite{goodfellow2014generative} models are also succinctly described afterwards. 

A diffusion process describes the transformation of an observed data $\mathbf{x_0} \in \mathbb{R}^{D}$ to  a pure noise $\mathbf{x}_{T} \in \mathbb{R}^{D}$ using a sequence of latent variables $\{\mathbf{x_i}\}_{i=1}^{T-1} \in \mathbb{R}^{D\times(T-1)}$. This transformation is parameterized by the approximated transition probability $p_{\theta}(\mathbf{x_{t-1}}|\mathbf{x_{t}})$ (see \cref{sup:parametrization_diffusion} for more mathematical details). The first diffusion model we consider in this article is the conditional Denoising Diffusion Probabilistic Model (\DDPM{}), introduced by \citealp{ho2020denoising}. The \DDPM{} reduces the learning of $p_{\theta}(\mathbf{x_{t-1}}|\mathbf{x_{t}})$ to the optimization of a simple conditional auto-encoder $\epsilon_{\theta}$ (with $\epsilon_{\theta} : \mathbb{R}^{D} \times \mathbb{R}^{D} \to {R}^{D}$, see \cref{sup:ddpm_loss}):
\begin{align}
\argmin_\theta \mathbb{E}_{(\mathbf{x_{t}},\mathbf{y}), \mathbf{\epsilon}}\Bigr[\left\|\mathbf{\epsilon}_{\theta}(\mathbf{x}_{t}, \mathbf{y}) - \mathbf{\epsilon} \right\|_{2}^{2}\Bigr] \mquad \text{s.t.} \mquad \mathbf{\epsilon} \sim \mathcal{N}(\mathbf{0}, \mathbf{I})
\label{eq:ddpm}
\end{align}
Said differently, $\mathbf{\epsilon}_{\theta}$ is trained to predict the noise $\mathbf{\epsilon}$ using a degraded sample $\mathbf{x}_t$ and the exemplar $\mathbf{y}$. \cref{eq:ddpm} is a denoising score matching objective~\cite{song2020score}, so the optimal model $\epsilon_{\theta^{*}}$ matches the following score function:
\begin{equation}
\nabla_{\mathbf{x}_{t}}\log p_{\theta^{*}}(\mathbf{x}_{t}|\mathbf{y}) \approx -\frac{1}{\sqrt{1-\bar{\alpha}_t}}\mathbf{\epsilon}_{\theta^{*}}(\mathbf{x}_t, \mathbf{y})
\label{eq:score}
\end{equation}
In \cref{eq:score}, $\bar{\alpha}_t$ is used to schedule the noise degradation of $\mathbf{x}_t$ (see \cref{sup:reparametrization} in \cref{sup:parametrization_diffusion}). Training information, details on the architecture, and samples generated by the \DDPM{} are available in~\cref{sup:DDPM_omniglot,sup:DDPM_qd}. 

\citealp{dhariwal2021diffusion} have shown that one could improve the conditioning signal of the \DDPM{} by guiding the forward process with a classifier. The second diffusion model we consider, the Classifier Free Guided Diffusion Model (\CFGDM{}), adopts a similar idea but replaces the classifier with a conditional generative model~\cite {ho2022classifier}. The score function of the \CFGDM{} can be expressed using the \DDPM{} one (see \cref{sup:CFGDM_loss} for more details):
\begin{align}
%\nabla_{\mathbf{x}_{t}}\log p_{\theta^{*}, \gamma}(\mathbf{x}_{t}|\mathbf{y})  = (1-\gamma)\nabla_{\mathbf{x}_{t}} \log p_{\theta^{*}}(\mathbf{x}_{t})  \nonumber\\ 
%+ \,\gamma \nabla_{\mathbf{x}_{t}} \log p_{\theta^{*}}(\mathbf{x}_{t}|\mathbf{y})
\nabla_{\mathbf{x}_{t}}\log p_{\theta^{*}, \gamma}(\mathbf{x}_{t}|\mathbf{y})  =  (1 + \gamma) \nabla_{\mathbf{x}_{t}} \log p_{\theta^{*}}(\mathbf{x}_{t}|\mathbf{y})\nonumber\\ 
- \, \gamma \nabla_{\mathbf{x}_{t}} \log p_{\theta^{*}}(\mathbf{x}_{t})
\label{eq:CFGDM_score}
\end{align}
This formulation introduces a guidance scale $\gamma$ to tune the part of the distribution that captures the influence of the conditioning signal. Note that in \cref{eq:CFGDM_score}, the two terms on the right hand side are parametrized by the same neural networks and are trained together (with $ \nabla_{\mathbf{x}_{t}} \log p_{\theta^{*}}(\mathbf{x}_{t}) \propto \mathbf{\epsilon}_{\theta^{*}}(\mathbf{x}_{t}, \mathbf{\varnothing})$ and $\nabla_{\mathbf{x}_{t}} \log p_{\theta^{*}}(\mathbf{x}_{t}|\mathbf{y}) \propto \mathbf{\epsilon}_{\theta^{*}}(\mathbf{x}_{t}, \mathbf{y})$, see \cref{sup:CFGDM_practical}). In the \CFGDM{}, $\gamma$ is set to $1$, except if specified otherwise. Training information, details on the architecture, and samples generated by the \CFGDM{} are available in ~\cref{sup:CFGDM_omniglot,sup:CFGDM_qd}.

The third diffusion model we consider is the Few-shot Diffusion Model (\FSDM{}, \citealp{giannone2022few}). In the \FSDM{}, the feature maps of the auto-encoder $\mathbf{\epsilon}_{\theta}$ are conditioned with a context vector $\mathbf{c}=h(\mathbf{y})$ using a FiLM-like mechanism~\citep{ethan2018film}. Note that this is different from the \DDPM{} and the \CFGDM{} that are conditioned by stacking the degraded samples $\mathbf{x}_t$ with the exemplar $\mathbf{y}$. We refer the reader to \cref{sup:FSDM_omniglot,sup:FSDM_qd} for more information on the conditioning mechanism, the architecture, and the samples of the \FSDM{}.

For the sake of comparison, we also include in this study the one-shot generative models presented in \citealp{boutin2022diversity}: the \VAENS{}, the \VAESTN{}, the \DAGANRN{} and the \DAGANUN{}. Both \VAENS{} and \VAESTN{} belong to the family of conditional Variational Auto-Encoders (VAE). The \VAENS{}, also called the Neural Statistician~\cite{edwards2016towards, giannone2021hierarchical} is conditioned on a \textit{context} set (similar to \FSDM{}, see \cref{sup:VAENS_qd}). The \VAESTN{} is a sequential VAE which includes an attention mechanism that learns to focus on important locations of the exemplar image~\cite{rezende2016one}. The \VAESTN{} iteratively generates images using a recurrent network~(see \cref{sup:VAESTN_qd}). Both \DAGANRN{} and \DAGANUN{} are Data Augmentation Generative Adversarial Networks, that are conditioned on a compressed representation of the exemplar image~\cite{antoniou2017data}. The \DAGANUN{} is based on the U-Net architecture and the \DAGANRN{} leverages the ResNet architecture (see \cref{sup:DAGAN_qd}). The code to train these models and to reproduce all the results of this paper is available on \url{https://github.com/serre-lab/diffusion_as_artist}.

\section{Methods}
\subsection{Datasets}
\label{methods:dataset}
\textbf{Omniglot }is composed of binary images representing $1,623$ classes of handwritten letters and symbols (extracted from $50$ alphabets) with only $20$ samples per class~\cite{lake2015human}. We have downsampled the original dataset to be $50\times50$ pixels. In this article, we use the weak generalization split, in which the training set is composed of all available symbols minus $3$ symbols per alphabet left aside for the test set~\cite{rezende2016one}. It is called \textit{weak} because all the alphabets are shown during the training (but not all symbols).

\textbf{QuickDraw-FewShot} (QuickDraw-FS) is built on the \textit{Quick, Draw !} challenge~\cite{jongejan2016quick}, in which human subjects are presented with object names and are asked to draw them in less than $20$ seconds. The original dataset is not suitable for the one-shot generation task because some object categories include more than one visual concept. For example, the `clock' visual concept includes digital and analog clocks (see~\cref{fig_sup:QD_concepts} for more examples). We use a clustering method on the original QuickDraw dataset to isolate distinct visual concepts for each object category (see~\cref{sup:qd_fs_processing} for a step-by-step description of the clustering and the filtering method). The resulting dataset, called QuickDraw-FS, is fully compatible with the one-shot scenario. It is composed of black \& white images representing $665$ distinct visual concepts. Each of the visual concepts is described by an exemplar and $500$ variations. The training set is made of $550$ randomly sampled visual concepts. The remaining $115$ visual concepts constitute the test set. We have downsampled the images to be $48\times48$ pixels so that it could be fed into ResNet blocks without resizing. 

Note that the dissimilarity between the training and test visual concepts is higher in the QuickDraw-FS dataset than in the weak generalization split of the Omniglot dataset. Consequently, the one-shot generation task requires a greater generalization ability in the QuickDraw-FS dataset.

\begin{figure*}[ht!]
%\vskip 0.2in
\begin{subfigure}{0.5\textwidth}
\includegraphics[width=\columnwidth]{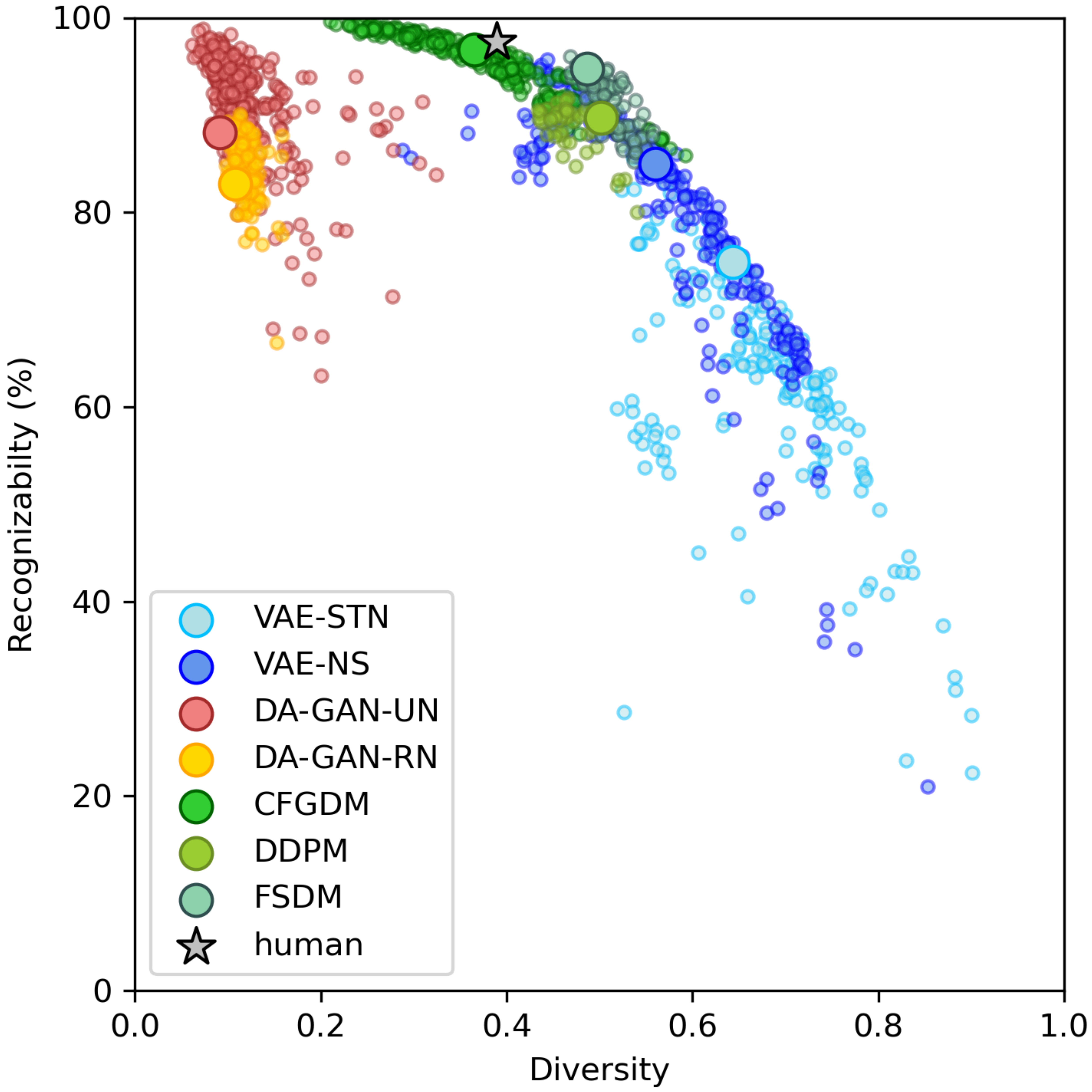}
\caption{Omniglot}
\label{fig:fig1a}
\end{subfigure}  
\begin{subfigure}{0.5\textwidth}
\includegraphics[width=\columnwidth]{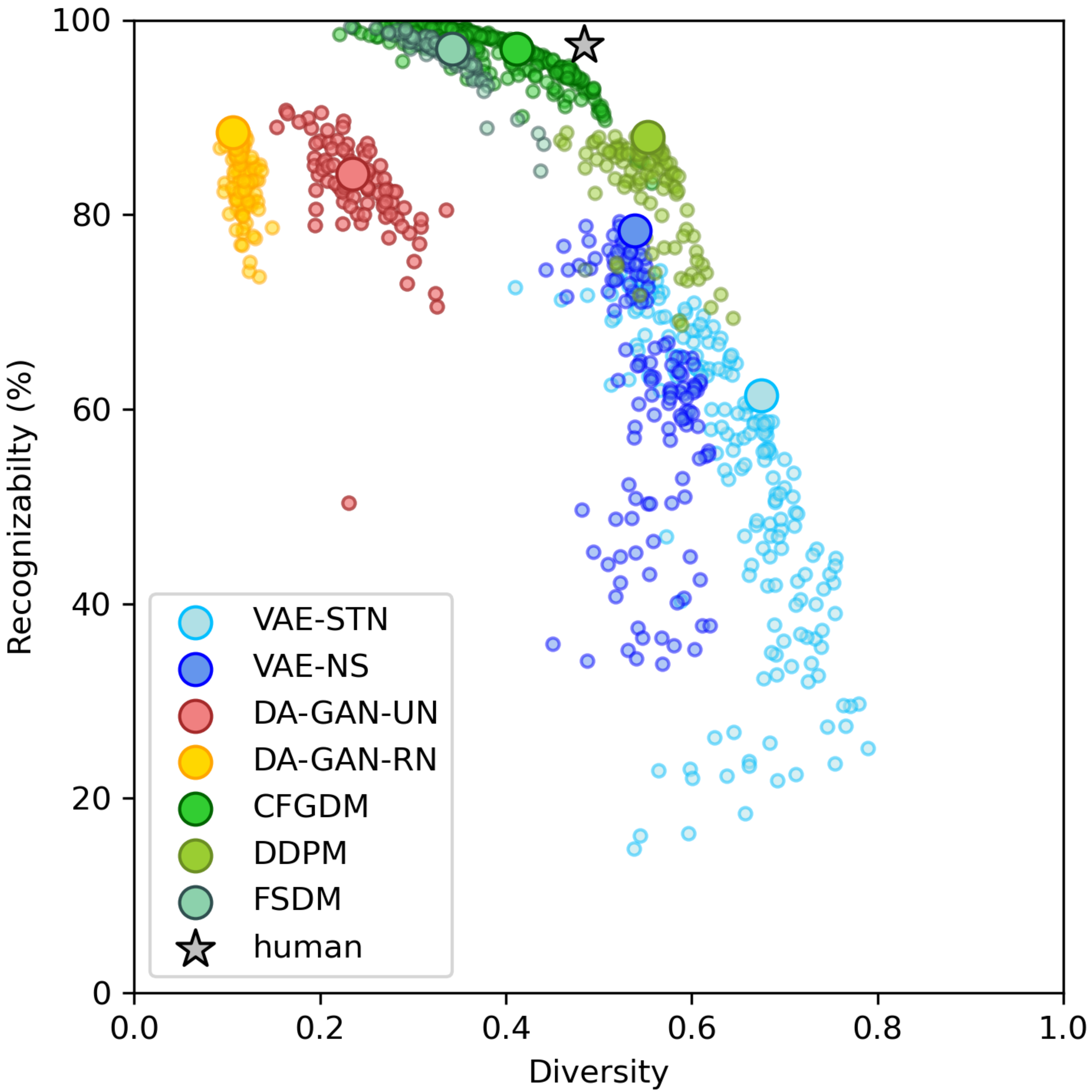}
\caption{QuickDraw-FS}
\label{fig:fig1b}
\end{subfigure}
\caption{\setcounter{footnote}{1}
Diversity vs. recognizability plots for models (colored data points) and humans (black/grey star) for \textbf{(a)} the Omniglot dataset ($1\,320$  models tested) and
\textbf{(b)} the QuickDraw dataset ($1\,212$  models tested). Data points for \VAENS{}, \VAESTN{}, \DAGANRN{} and \DAGANUN{} on Omniglot were computed using code from~\citealp{boutin2022diversity}\footnotemark. Each data point corresponds to the mean diversity and recognizability computed over all classes on the test set. Larger circles correspond to  base architectures for which we controlled the number of parameters ($\approx6-7$M for Omniglot and $\approx12-13$M for QuickDraw-FS). The \human~data point is computed based on the test samples of the Omniglot and the QuickDraw-FS datasets.}
\vskip -0.1in
\end{figure*}

%\textbf{Fleming shape: }
\subsection{The Diversity vs. Recognizability Framework}
The ``diversity vs. recognizability'' framework was initially proposed by~\citealp{boutin2022diversity} to evaluate the performance of humans and machines on the one-shot generation task. \revision{Within this framework, $2$ distinct scores are assigned to the variations generated from a given exemplar: the diversity, which assesses the intra-class variability of all variations of a given class, and the recognizability, which measures whether the variations fall within the same class as the exemplar. Intuitively, an ideal model should fall in the top-right region of the diversity vs recognizability space (see Figure $1$ of ~\citealp{boutin2022diversity}) : it should produce recognizable samples (high y-axis values) that are as diverse as possible (high x-axis values). On the contrary, a model that can only produce ``copy'' of the exemplar would fall in the top-left corner, and a model producing random nonsensical samples would be located on the bottom-right corner. } %The diversity score quantifies the intra-class variability of the generated samples and is computed with a standard deviation across samples from the same class (see \cref{sup:creativity and diveristy} for more details). 
\revision{The diversity score is computed with a standard deviation across samples from the same class (see \cref{sup:creativity and diveristy} for more details). }
\revision{Note that the intra-class variability is not directly computed in the pixel space, but rather in the feature space of a SimCLR network~\citep{chen2020simple}. The SimCLR features offer greater invariance to transformations such as rotation, translation and scaling.} \revision{We call ``recognizability'', the classification accuracy as evaluated by a one-shot classifier. Herein, we use a Prototypical Net to perform one-shot classification~\citep{snell2017prototypical}. The choice of the $2$ critic networks (i.e. the SimCLR and the Prototypical) is validated and discussed in the Annex S3 of~\citealp{boutin2022diversity}.} On the QuickDraw-FS dataset, we have adapted the architectures of the critic networks (see \cref{diversity_recognizability_qd}). In this work, we normalize the diversity metric such that the average standard deviation across features is equal to one. Such a normalization allows us to more faithfully compare the diversity and recognizability scores across different datasets or different critic networks (see \cref{diversity_normalization}). 
\footnotetext{\url{https://github.com/serre-lab/diversity_vs_recognizability} }

\section{Results}
\subsection{Diversity vs Recognizability}

\cref{fig:fig1a,fig:fig1b} show diversity vs. recognizability plots for all algorithms described in~\cref{related:one_shot_models}. We  conducted an extensive hyper-parameter exploration with a total of $1\,320$ and $1\,212$ models trained for Omniglot and QuickDraw-FS, respectively. Each data point represents a single model whereby the diversity and recognizability scored were averaged over all classes of the test set. The black star in each plot corresponds to the human ideal model, and the colored points are the one-shot generative models. Large data points represent models' base architectures, with a comparable number of parameters: $\approx$ 6-7M parameters for Omniglot (\cref{fig:fig1a}) and $\approx$ 12-13M parameters for QuickDraw-FS (\cref{fig:fig1b}). The \VAENS{}, the \VAESTN{}, the \DAGANUN{} and the \DAGANRN{} trained on Omniglot are the same exact architectures as reported in~\citealp{boutin2022diversity} using original code from these authors. 
The different hyper-parameters we have varied to obtain the point cloud for each model are described in \cref{sup:DDPM_omniglot,sup:CFGDM_omniglot,sup:FSDM_omniglot,sup:VAENS_qd,sup:VAESTN_qd,sup:DAGAN_qd,sup:DDPM_qd,sup:CFGDM_qd,sup:FSDM_qd}.

\begin{figure*}[ht!]
%\vskip 0.2in
\begin{subfigure}{0.5\textwidth}
\includegraphics[width=\columnwidth]{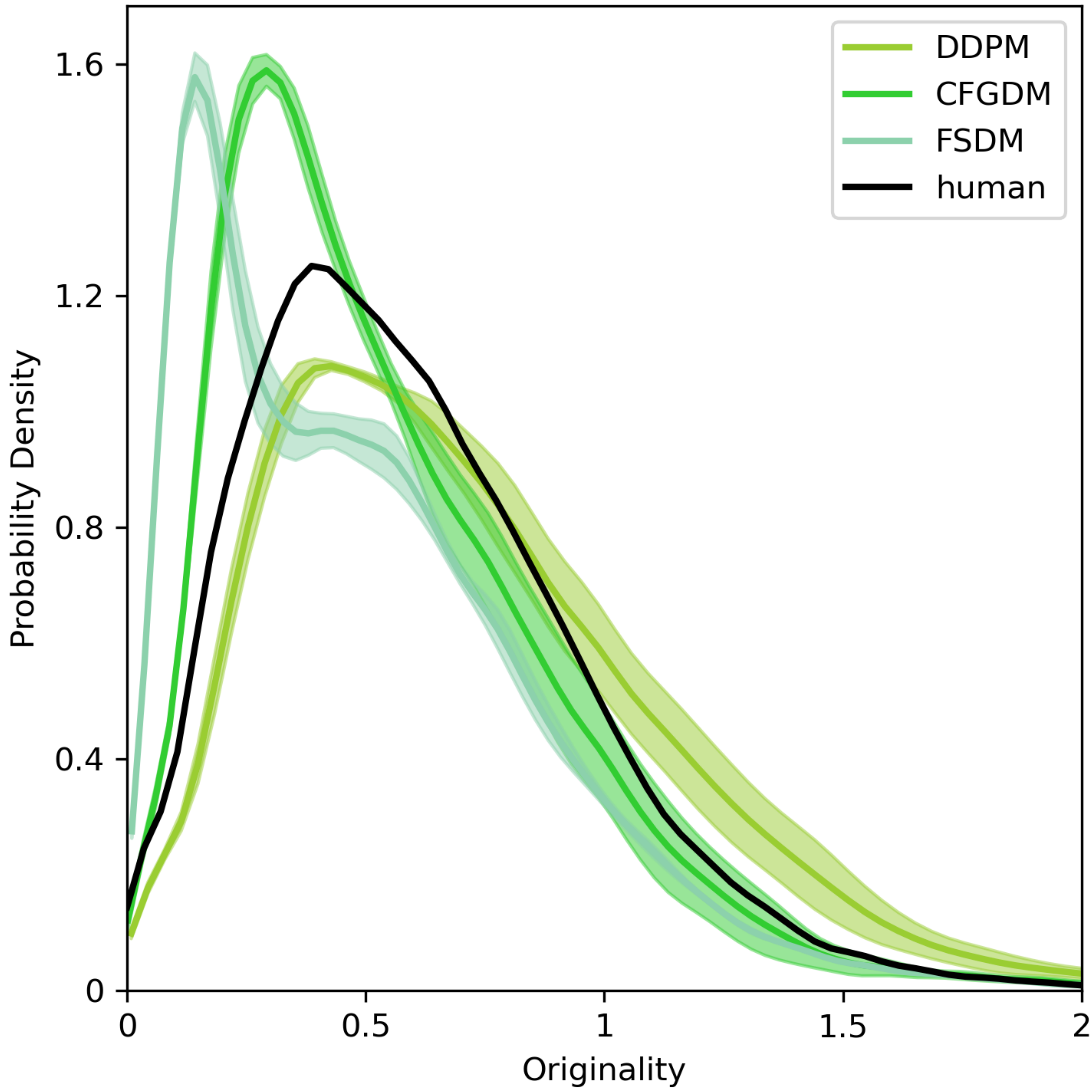}
\caption{Probability density}
\label{fig:fig2a}
\end{subfigure}  
\begin{subfigure}{0.5\textwidth}
\includegraphics[width=\columnwidth]{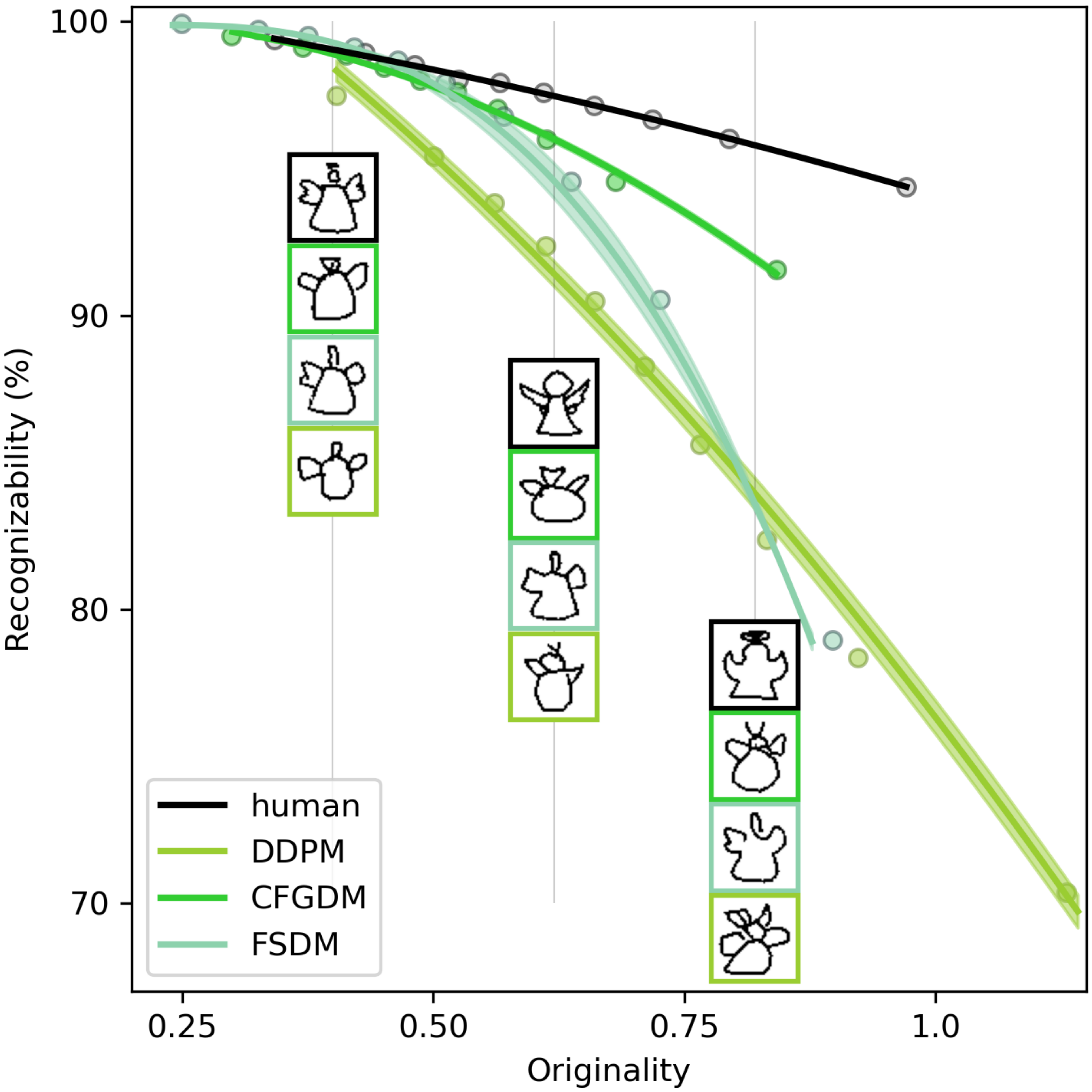}
\caption{Generalization curves}
\label{fig:fig2b}
\end{subfigure}
\caption{\textbf{(a)}: Distribution of the samples' \textit{originality} computed for the \DDPM{}, \CFGDM{}, \FSDM{} and \humans{}. Originality scores were computed as the $\ell_2$-distance, in the SimCLR feature space, between a sample and its exemplar. Distributions were estimated from histograms using a Gaussian density kernel estimation approach. \textbf{(b)}: Generalization curves for the \DDPM{}, \CFGDM{}, \FSDM{} and \humans{}. Each data point corresponds to the average originality and recognizability over the samples in each of the $10$ originality bins. Plain lines are smooth interpolations (polynomial regression) between data points. Thumbnails show human and model-generated samples for $3$ different levels of originality. For both panels, base architectures (corresponding to the larger markers in \cref{fig:fig1a,fig:fig1b}) trained on the QuickDraw-FS dataset  were used for all models. Shaded areas are computed using the standard deviation over $3$ different runs.} 
\vskip -0.2in
\end{figure*}

Overall, GANs (\DAGANRN{} and \DAGANUN{}) tend to exhibit low diversity for both the Omniglot (\cref{fig:fig1a}) and the QuickDraw-FS datasets (\cref{fig:fig1b}).VAEs (\VAENS{} and \VAESTN{}) display a higher diversity but also slightly lower recognizability for both datasets. This observation has already been made on the Omniglot dataset by~\citealp{boutin2022diversity}; Here, we generalize this result to a more complex dataset (QuickDraw-FS). Furthermore, we also observe a drop in recognizability between the VAEs trained on Omniglot and those trained on QuickDraw-FS: from $86\%$ to $78\%$ for the \VAENS{} and from $74\%$ to $61\%$ for the \VAESTN{}. This phenomenon is less pronounced for GANs and diffusion models. This suggests that GANs (\DAGANRN{}, \DAGANUN{}) and diffusion models (\DDPM{}, \CFGDM{} and \FSDM{}) are easier to scale up, without major architecture changes, to more complex datasets than VAEs (\VAENS{} and \VAESTN{}). This observation tends to corroborate the scaling difficulties already reported for the VAEs~\citep{bond2021deep, vahdat2020nvae}.

We notice that with identical numbers of parameters to VAEs and GANs, diffusion models (\DDPM{}, \CFGDM{} and \FSDM{}) can produce more recognizable samples on Omniglot and QuickDraw-FS. This observation aligns with the latest findings suggesting that diffusion models beat other models in terms of sample quality~\citep{dhariwal2021diffusion}. In terms of diversity, the diffusion models are in between the GANs and the VAEs. The \CFGDM{}, the \DDPM{}, and \FSDM{} data points consistently fall in a close neighborhood of the human ideal observer (black star) for both datasets. We conclude that diffusion models provide the best approximation of human-level drawings. Henceforth, we will focus on the diffusion models and the human data on the QuickDraw-FS dataset.

\subsection{Generalization Curves}
\label{generaliation_curves}

Here, we introduce the \textit{originality} metric to quantify the distance between an individual sample and the corresponding exemplar. This distance is computed using a $\ell_2$-norm in the feature space of a SimCLR network~(see \cref{diversity_recognizability_qd} for more details on the SimCLR architecture). Intuitively, the higher the distance to the exemplar, the more original the sample and the higher the inventiveness and creativity of the corresponding model. We have validated our originality measure through a series of control experiments with different feature extractor networks and different distance metrics (see~\cref{SI:originality_metric}). In~\cref{fig_sup:Samples_sorted_originality}, one can see that our originality metric is qualitatively similar to human judgments.

We draw the reader's attention to the fact that the originality and diversity metrics are different. Even though both metrics tell us something about the creative process of models, the "diversity" is an aggregate per-class measure (the intra-class variability, hence one model as one value of diversity per class), while the "originality" is a per-sample measure (the distance of one sample generated to the exemplar, hence we can probe the model at different levels of "originality" for a given class). The former assesses the mean distance between samples of the category and the corresponding exemplar, and the latter evaluates the mean distance to the center of the category cluster (see~\cref{sup:creativity and diveristy} for more details).  In \cref{fig:fig2a}, we plot the distribution of the samples' originality for the \human{}, the \DDPM{}, the \CFGDM{} and the \FSDM{} on the QuickDraw-FS dataset. The \FSDM{} distribution is highly concentrated in the low-originality region, which suggests that the model tends to produce samples that are similar to the exemplar. On the contrary, the \DDPM{} has a distribution spreading towards higher originality (positive skewness). This indicates that the \DDPM{} has the ability to generate samples that are more dissimilar to the exemplar. The originality metric informs us about the inventiveness of the corresponding model, through the distance to the exemplar, but it does not tell us anything about how faithfully the sample represents the visual concept conveyed by the exemplar.

To overcome this limitation, we introduce the \textit{generalization curve}. For a given model, the generalization curve quantifies the evolution of the recognizability at different levels of originality. It is important to emphasize that a generalization curve describes samples that are all generated by the same model. For each class, we sort the samples into originality bins, such that the samples belonging to the same bin have a relatively similar distance to the exemplar. In particular, we arrange the samples into $10$ bins with $50$ samples in each bin. As a result, the QuickDraw-FS dataset is split into $10$ sub-sets, each containing samples with comparable originality levels. We then compute the average originality and recognizability for each bin (see data points in \cref{fig:fig2b}). We ultimately derive generalization curves by smoothly interpolating between data points  (using polynomial regression). For each model, we report the regression error in~\cref{sup:interp_error}. These generalization curves are shown with plain lines for \human{}, the \DDPM{}, the \CFGDM{} and the \FSDM{} on the QuickDraw-FS dataset in \cref{fig:fig2b}. Note that such curves would not be statistically meaningful on the Omniglot dataset due to the reduced number of samples per class ($20$ samples). \revision{Intuitively, an agent with superior generalization capabilities would be expected to exhibit a generalization curve that is shifted towards the upper-right corner. Such an agent would be able to evaluate so accurately the decision boundary that it could produce samples lying in its close vicinity.}

In \cref{fig:fig2b}, we observe that the \FSDM{} model is able to produce samples with the highest recognizability ($\approx 100 \%$) albeit with the lowest originality ($\approx 0.25$). The \FSDM{} recognizability falls sharply as the originality score increases. The \DDPM{} generalization curve spans the longest range in terms of originality (from $0.40$ to $1.15$) and recognizability (from $70 \%$ to $97 \%$). The \DDPM{} can produce samples that are different from the exemplar but are also poorly recognizable. The \CFGDM{} samples are less original but also more recognizable than the \DDPM{} samples.~\Humans{} maintain the best generalization curve in the high-originality regime: the \human{} curve is above all others for originality values greater than $0.53$. It suggests that \humans{} can produce samples that are simultaneously more original and more recognizable than all the models. For an originality score of $0.75$, the recognizability of \human{} samples is $96\%$, that of \CFGDM{} is around $92\%$ and that of \DDPM{} drops to $87\%$. Among all tested models, the \CFGDM{}  best approximates the human generalization curve.

\begin{figure}[t!]
%\vskip 0.2in
\begin{center}
\centerline{
\includegraphics[width=\columnwidth]{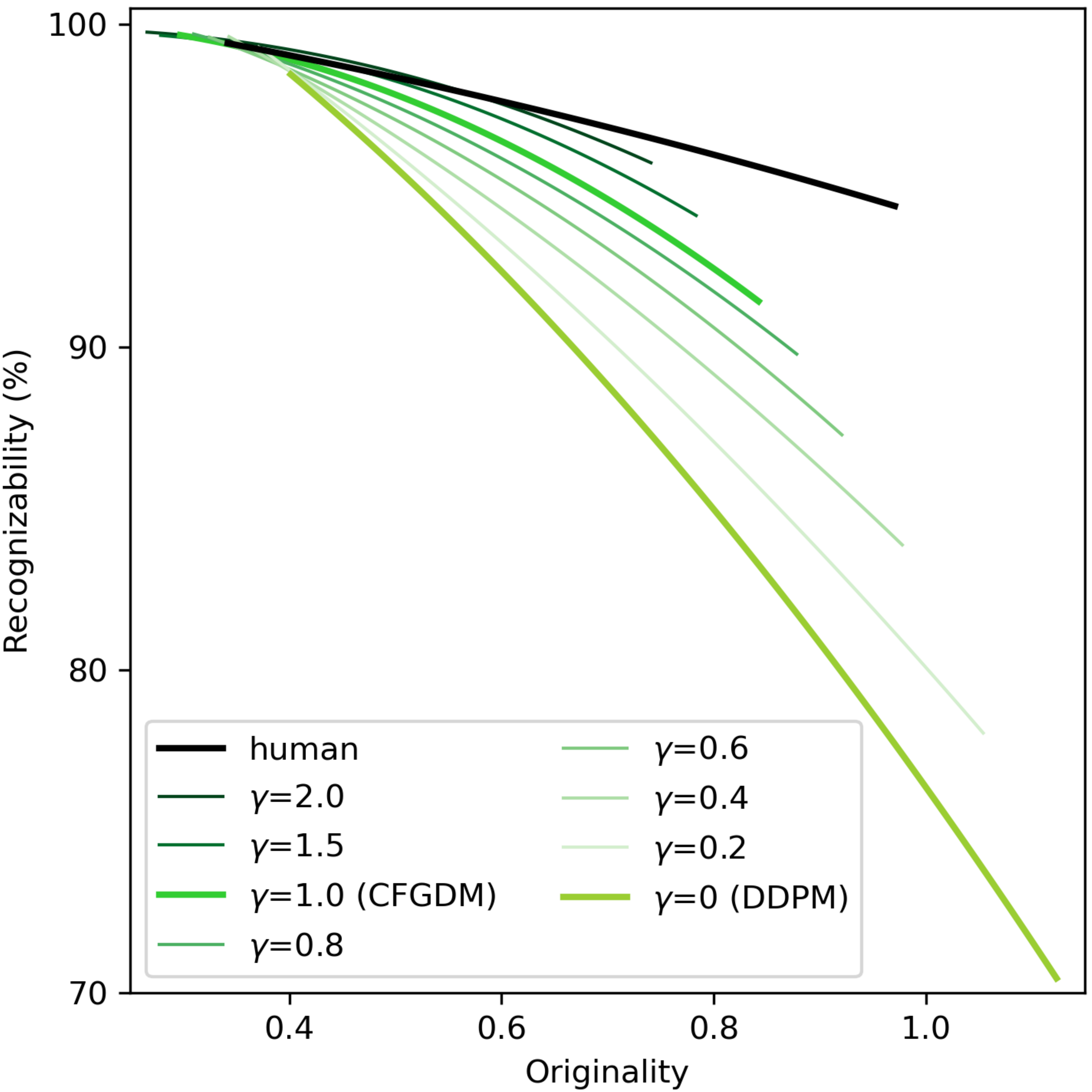}
}
\caption{Generalization curves for \humans{}, the \DDPM{} and the \CFGDM{} with different levels of guidance ($\gamma$) on the QuickDraw-FS dataset. Each curve represents a different model. The gray dashed line is the best possible generalization curve for the \CFGDM{} models. For readability, we have omitted the data points (only  smooth interpolation curves are shown; see~\cref{sup:interp_error} for the corresponding interpolation errors).}
\label{fig:fig3}
\end{center}
\vskip -0.2in
\end{figure}

\begin{figure*}[t!]
\vskip 0.2in
\begin{subfigure}{0.5\textwidth}
\includegraphics[width=0.98\columnwidth]{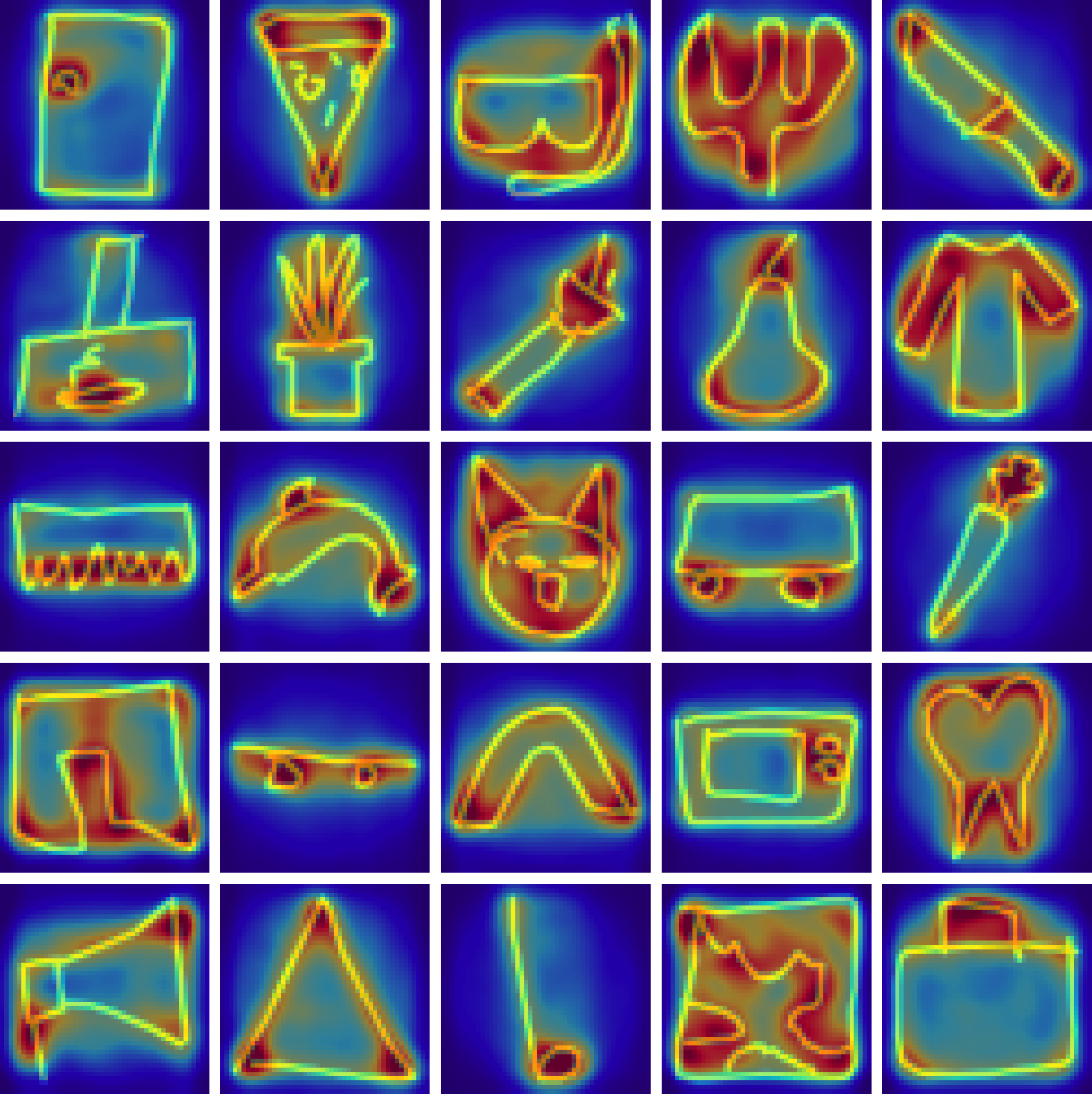}
\caption{\CFGDM{}}
\label{fig:fig4a}
\end{subfigure}  
\begin{subfigure}{0.5\textwidth}
\includegraphics[width=0.98\columnwidth]{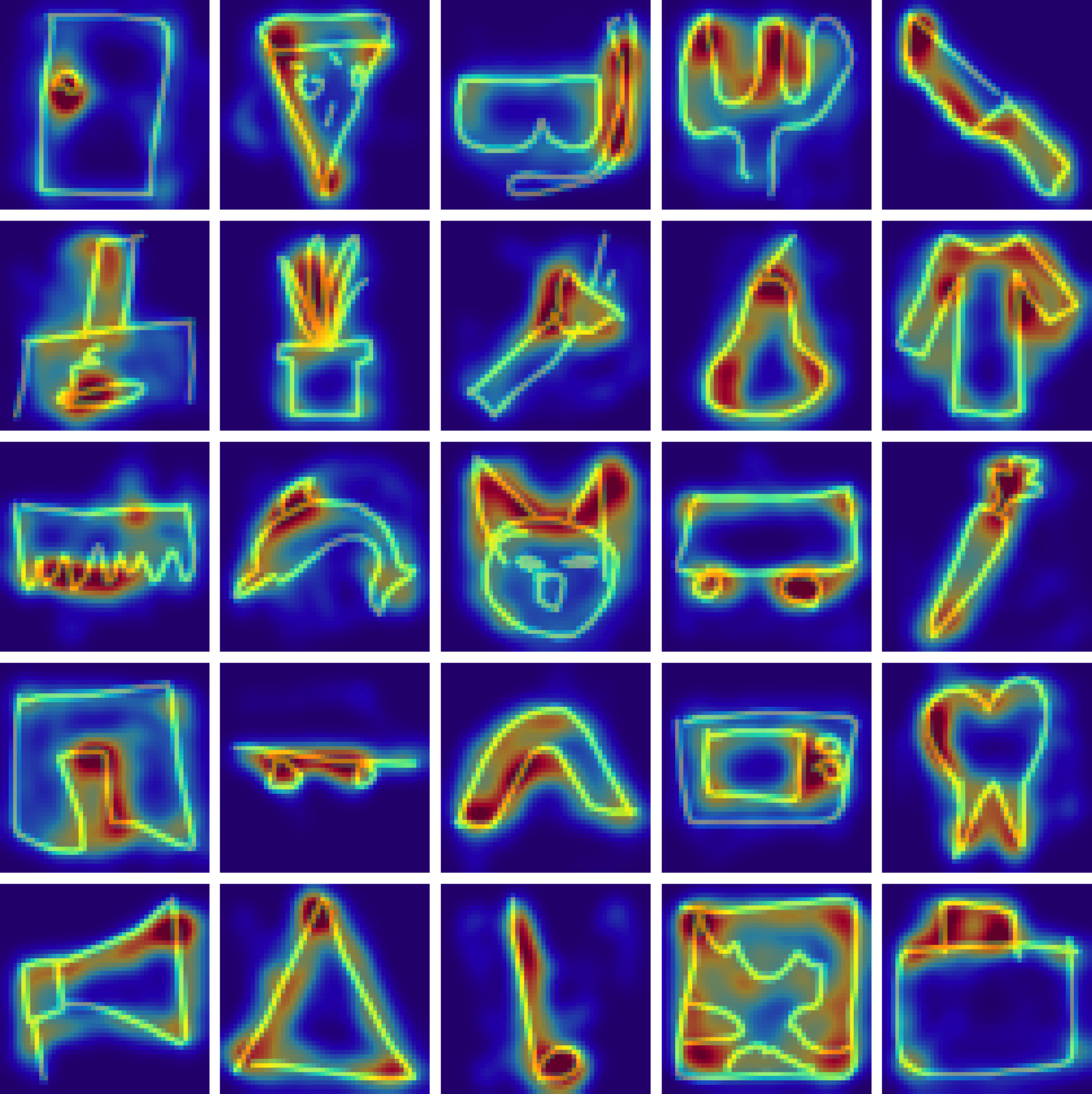}
\caption{\human{}}
\label{fig:fig4b}
\end{subfigure}
\caption{Importance maps (overlaid on \revision{exemplars}) derived for \textbf{(a)} \CFGDM{} and \textbf{(b)} \human{} observers for $25$ representative visual concepts. Hot vs. cold pixels indicate image locations that are more vs. less important.   Maps for \textbf{(a)} \CFGDM{} were obtained by averaging over $n=10$ misalignment maps $\bm{\phi}(\mathbf{x}, \mathbf{y})$ as defined in \cref{eq:attribution_score}. Maps for \human{} observers were obtained using our ClickMe-QuickDraw online game. } 
\label{fig:fig4}
\vskip -0.2in
\end{figure*}

In \cref{fig:fig3}, we study the impact of the guidance scale (the $\gamma$ coefficient in~\cref{eq:CFGDM_score}) on  generalization curves. Increasing the guidance scale has a double effect on the \CFGDM{} score: i) it encourages the conditional term (first term of the RHS of~\cref{eq:CFGDM_score}) and ii) it penalizes the unconditional term (second term of the RHS of~\cref{eq:CFGDM_score}). When $\gamma=0$, the unconditional term in~\cref{eq:CFGDM_score} disappears, the \CFGDM{} score becomes then strictly equivalent to the \DDPM{} one. The base architecture of the \CFGDM{} is obtained with $\gamma=1$. We observe improved recognizability and a decrease in originality as we increase the guidance scale from $0$ to $2$. This observation is even more pronounced for high originality values (see the right end of the curves in \cref{fig:fig3} for different guidance scales). Overall, we observe a progressive shrinkage of the originality and the recognizability range as we increase $\gamma$. Interestingly, a similar phenomenon has also been reported on natural images~\citep{dhariwal2021diffusion}. The \DDPM{} (i.e., when $\gamma=0$) spans an originality range of $0.75$ (from $0.4$ to $1.15$) and a recognizability range of $27\%$ (from $70\%$ to $97\%$) whereas the \CFGDM{}, with $\gamma=2$, spans an originality range of $0.5$ (from $0.25$ to $0.75$) and a recognizability range of $3\%$ (from $97\%$ to $100\%$).

We note that the generalization curve of the \CFGDM{}, with $\gamma=2$, provides a good approximation to the \human{} generalization curve for low originality values (below $0.6$).  But this model fails to account for \human{} generalization in higher originality regimes. In~\cref{fig:fig3}, we highlighted the best possible generalization curve for the \CFGDM{} models with the gray dashed line. This curve is obtained by selecting the model with the highest recognizability for all originality levels. We observe that this curve still shows a severe drop in recognizability as the samples get more original. Among all tested models, we did not find one that is able to reach the recognizability of \humans{} for a high level of sample originality.

\subsection{Comparing Human and Machine Visual Features}

To delve deeper into the differences observed between \CFGDM{} and \humans{}, we study the diagnosticity of individual features for each category.
%To understand these differences  %As we delve deeper into the question of whether humans and machines share a similar approach to solving the one-shot drawing task,
%we turn our attention to the examination of
%category diagnostic features. % for both the \CFGDM{} and the \human{}. 
% These features are meantcharacteristic of a given category and they might hold the key to understanding generalization in the one-shot drawing task.

% Intuitively, the diagnostic features 
%The diagnostic features are those that are very specific to the class
%the unconditional score encode for the features that are common to all classes
%The conditional score encodes for the features that are 
%The features that are very specific to the class, should see the biggest misalignement between the conditional and the unconditional part of the score.

We draw inspiration from attribution methods~\cite{zeiler2014visualizing,sundararajan2017axiomatic,smilkov2017smoothgrad,fel2021sobol,novello2022making,fel2022don} and use the score function decomposition of the \CFGDM{} to visualize diagnostic features. We denote $\bm{\delta}_{t}^{\mathbf{y}}(\mathbf{x}, \mathbf{y})$ the part of the score conditioned on the exemplar at each time step $t$, and $\bm{\delta}_{t}(\mathbf{x})$ the unconditional part of the score. $\bm{\delta}_{t}^{\mathbf{y}}(\mathbf{x}, \mathbf{y})$ conveys information specific to the $\mathbf{y}$ exemplar, while $\bm{\delta}_{t}(\mathbf{x})$ encodes more general properties (e.g., background color, stroke size, etc). As previously observed in \cref{fig:fig3}, the misalignment between the conditional and unconditional signals is strongly related to image recognizability. Therefore, such a misalignment could be used to identify the most discriminative features. For a given sample $\mathbf{x}$ exemplified by the exemplar $\mathbf{y}$, we propose a metric, denoted $\bm{\phi}(\mathbf{x}, \mathbf{y})$, to evaluate the misalignment at every position in the image.
%We propose a metric, denoted $\bm{\phi}(\mathbf{x}, \mathbf{y})$, to evaluate the misalignment at every position in the image for a given sample $\mathbf{x}$ that represents a variation of the exemplar $\mathbf{y}$. 
\vspace{-2pt}
\begin{flalign}
\bm{\phi}(\mathbf{x}, \mathbf{y}) = \sum_{t=1}^{T} \Big \lvert \frac{\bm{\delta}_{t}^{\mathbf{y}}(\mathbf{x}, \mathbf{y})}{\lVert \bm{\delta}_{t}^{\mathbf{y}}(\mathbf{x}, \mathbf{y})\rVert_{2}} - \frac{\bm{\delta}_{t}(\mathbf{x})}{\lVert \bm{\delta}_{t}(\mathbf{x})\rVert_{2}} \Big \rvert \quad \quad  \label{eq:attribution_score} \\ 
\text{s.t} \quad \left\{ \begin{array}
{ll}
\bm{\delta}_{t}^{\mathbf{y}}(\mathbf{x}, \mathbf{y}) & = \nabla_{\mathbf{x}_{t}}\log p_{\theta^{*}}(\mathbf{x}_{t}|\mathbf{y}) \\
\bm{\delta}_{t}(\mathbf{x}) &= \nabla_{\mathbf{x}_{t}}\log p_{\theta^{*}}(\mathbf{x}_{t})
\end{array}
\right. \nonumber
\end{flalign}
This metric is computed by accumulating over all time steps the absolute value of the difference between the normalized conditional and unconditional scores. For each category, we average over $10$ misalignment maps to obtain the final feature importance map. In \cref{fig:fig4a}, we show representative feature importance maps for the \CFGDM{}, trained on QuickDraw-FS, for $25$ different categories (see \cref{SI:additionalImportanceMap} for more feature importance maps).

We conducted an online experiment, called ClickMe-QuickDraw, to get feature importance maps for comparison with \humans{}. The ClickMe-QuickDraw experiment follows a similar protocol to the ClickMe experiment initially used by~\citealp{linsley2018learning} to derive human importance maps for ImageNet. In ClickMe-QuickDraw, participants are asked to locate features in an image that they believe are important for categorizing it. As the participant selects important image regions, those regions are gradually revealed starting from a blank canvas and passed iteratively to a classifier. The participant gets rewarded whenever the classifier correctly classifies the canvas before the round time is up. At the end of each round, we obtain a ClickMe map: a map in which pixel intensities represent the probability of the pixel being painted by the participant. To obtain the importance feature map of a category, we average the ClickMe maps over all participants and images for that category. To keep a fair comparison with the \CFGDM{} importance feature maps, the same images were used as those used to compute the misalignment maps for the models. Crucially, previous studies have shown that the ClickMe experimental protocol produces feature importance maps that are perceptually meaningful~\cite {linsley2017visual,linsley2018learning}.
For the ClickMe-QuickDraw experiment, we collected $1,050$ ClickMe maps from $102$ participants. We refer the reader to \cref{SI:click} for more details on the ClickMe-QuickDraw experimental protocol as well as the statistics used to assess the reliability of the results. In \cref{fig:fig4b}, we show human feature importance maps for comparison with the model.
%As we gaze upon the feature importance maps, we cannot help but wonder if humanity and machines truly share a common strategy for solving the one-shot drawing task. Only further investigation will reveal the answer.

We observe that \CFGDM{} importance maps are more diffuse than those of \humans{}. The \CFGDM{} gives importance (albeit weak) to the background in the close vicinity of the object while \humans{} tend to focus only on sparser features of the object itself. In general, the category diagnostic features of \humans{} are also highlighted in the \CFGDM{} feature importance maps. %The striking exception is the knife ($1^{st}$ row, $5^{th}$ column in \cref{fig:fig4}): for \humans{} the blade is a diagnostic feature while the \CFGDM{} seems to rely on the tip and the handle of the knife. 
In short, \humans{} rely on fewer and more localized features to identify the object category. For example, \humans{} consider that the ears are diagnostic of the ``cat head'' while the \CFGDM{} tends to highlight the full head contour ($3^{rd}$ row, $3^{rd}$ column in \cref{fig:fig4}). One interesting exception is ``golf club'', the \CFGDM{} emphasizes the club's head while \humans{} consider that the club's shaft is also important. %More surprisingly, humans seem to agree together that the more important feature is the same single wheel for the ``truck'' and the ``skateboard'' classes ($3^{rd}$ row, $4^{th}$ column and $4^{th}$ row, $2^{nd}$ column in \cref{fig:fig4}, respectively), and the same corner for the ``triangle'' ($5^{th}$ row, $2^{nd}$ column in \cref{fig:fig4}). 
The comparison between the \CFGDM{} and \humans{} feature importance maps suggests that they rely on different visual strategies.

\section{Discussion}

In this article, we compared humans and machines on a one-shot drawing task. We extended the ``diversity vs. recognizability'' framework of~\citealp{boutin2022diversity} to compare samples produced by various generative models against those produced by humans. We found that diffusion models (\DDPM{}, \CFGDM{} and \FSDM{}) offer a better approximation to human drawings than VAEs (\VAENS{}, \VAESTN{}) and GANs (\DAGANRN{}, \DAGANUN{}). Other studies have also reported state-of-the-art performances for diffusion models~\cite{chahal2022exploring, dhariwal2021diffusion, peebles2022scalable}. We hypothesize that this success comes from the fact that diffusion models have to evaluate a simpler mathematical object than VAEs and GANs. The former learns the progressive transition between $2$ noisy states, while VAEs and GANs have to encode the direct mapping from pure noise to the data distribution. %\revision{In Fig, we have observed that diffusion models produce samples that are more diverse than GANs. It suggests that the diffusion model are not suffering from the mode-dropping phenomenon observed in GANs. In addition, the}

%Interestingly, such progressive learning has also been proven successful in solving stochastic differential equations~\citep{song2020score}.%Such an argument is echoing the works on neural Fourrier operator

We further introduced the \textit{originality} metric to evaluate the distance to exemplar for each individual sample. We found that the \CFGDM{} provides a better approximation to human drawings in low-originality regimes compared to the \DDPM{} and the \FSDM{}. One unique feature of the \CFGDM{} is that it penalizes the features that are common to all classes in favor of more diagnostic features. Interestingly, forcing the model to move further away from non-category specific features improves the humanness of generated drawings in a low-originality regime (see~\cref{fig:fig3}). This suggests that humans might rely on a few features that are strongly discriminative to solve the one-shot drawing task. This hypothesis seems to align with the human data we have collected with the ClickMe-QuickDraw experiment: human feature importance maps tend to be sparser and tend to emphasize strongly localized features (see~\cref{fig:fig4b}). Additionally, this result is in line with other psychophysics experiments that have shown that small but specific fragments of an image are sufficient to humans for correct categorization~\cite{hegde2008fragment, ullman2002visual}.

Nevertheless, a question remains: how can humans produce drawings that are both highly original and recognizable? We speculate that this aspect of human drawings could be the consequence of their attentional strategy. Since humans seem to focus on a few localized features, non-important features may vary more freely to provide room for creativity. This hypothesis is supported by recent human experiments that have suggested that the alteration of non-discriminative features has little to no impact on their recognizability~\cite{tiedemann2022one}. 

\revision{Recent research has established the exceptional zero-shot generalization capabilities of diffusion models guided by expressive language embeddings~\citep{saharia2022photorealistic, ramesh2021zero, nichol2021glide, rombach2022high}. Nevertheless, psychophysics studies have shown that humans demonstrate impressive generalization abilities in drawing tasks even when the drawings are semantically nonsensical~\citep{tiedemann2022one}. It suggests that semantic knowledge is not a prerequisite for human to achieve good generalization, and our results suggest a similar pattern in one-shot generative models. However, given the inherently compositional nature of language, we speculate that semantic knowledge may greatly enhance the model's generalization capabilities.}

\revision{The metrics we have leveraged in this work provide a rigorous framework to probe the humans and machines generalization abilities in the one-shot drawing task. Those metrics have been carefully designed to avoid the pitfalls that are inherently related to the task: the law-data regime, and the dissimilarity between
the training and testing visual concepts (see section 2.1 in \citealp{boutin2022diversity}). Nevertheless, the evaluation metrics are based on critic networks (e.g. Prototypical Net or SimCLR) that might not be well aligned with the human perception. In future work, we plan to circumvent this issue using human-alignment methods~\citep{fel2022harmonizing,muttenthaler2022human}.}

 By introducing and quantifying samples' recognizability and originality, we wanted to shed light on the relationship between generalization and creativity. We hope the generalization curves of the types introduced in this work may help provide better benchmarks for future models and help further close the gap between machines and humans.

\section*{Aknowledgement}
We thank Roland W. Fleming and his team for the insightful feedback and discussion about the diversity vs. recognizability framework. This work was funded by ANITI (Artificial and Natural Intelligence Toulouse Institute) and the French National Research Agency,  under the grant agreement number : ANR-19-PI3A-0004. Additional funding was provided by ONR (N00014-19-1-2029) and NSF (IIS-1912280 and EAR-1925481).  Computing hardware supported by NIH Office of the Director grant S10OD025181 via the Center for Computation and Visualization (CCV). J.C. has been partially supported by funding from the Valencian Government (Conselleria d'Innovació, Universitats, Ciència i Societat Digital) by virtue of a 2022 grant agreement (convenio singular 2022).
%Recently, diffusion models guided by expressive language embeddings have demonstrated outstanding zero-shot performance~\citep{saharia2022photorealistic, ramesh2021zero, nichol2021glide, rombach2022high}. 
%However, this success in terms of systematic generalization is somewhat expected because of the naturally composable structure of language~\cite{singh2021illiterate, lake2018generalization, bahdanau2018systematic}. In contrast, humans can generalize even when they are dealing with purely visual tasks~\citep{lake2015human, tiedemann2022one, feldman1992constructing}. Where does such a generalization ability come from ? 

%By using the originality metr

%decomposing the performance of the one-shot generation task along the recognizability vs. diversity axes we wanted to shed light on the relationship between generalization and creativity
%(quantified by the samples diversity in our framework). We hope one can make use of our framework
%to validate key hypotheses about human generalization abilities so that we can better understand
%the brain. We argue that the best way to reach human-like generalization abilities is to unleash the
%algorithms’ creativity.

%\cleardoublepage
\clearpage
\newpage
\bibliography{biblio}
\bibliographystyle{icml2023_cr}

%%%%%%%%%%%%%%%%%%%%%%%%%%%%%%%%%%%%%%%%%%%%%%%%%%%%%%%%%%%%%%%%%%%%%%%%%%%%%%%
%%%%%%%%%%%%%%%%%%%%%%%%%%%%%%%%%%%%%%%%%%%%%%%%%%%%%%%%%%%%%%%%%%%%%%%%%%%%%%%
% APPENDIX
%%%%%%%%%%%%%%%%%%%%%%%%%%%%%%%%%%%%%%%%%%%%%%%%%%%%%%%%%%%%%%%%%%%%%%%%%%%%%%%
%%%%%%%%%%%%%%%%%%%%%%%%%%%%%%%%%%%%%%%%%%%%%%%%%%%%%%%%%%%%%%%%%%%%%%%%%%%%%%%

\appendix

\counterwithin{figure}{section}
\onecolumn

\newpage
\section{Construction of the QuickDraw-FS dataset}
\label{sup:QD_dataset}

\subsection{Different visual concepts for the same object category}

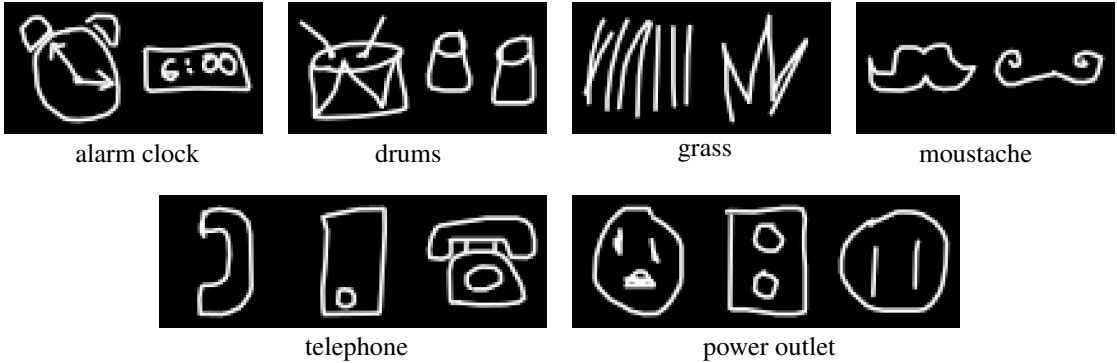
\begin{figure}[h!]
%\vskip 0.2in
\begin{center}

\begin{tikzpicture}
\draw [anchor=north west] (0\linewidth, 1\linewidth) node {\includegraphics[width=0.2\linewidth]{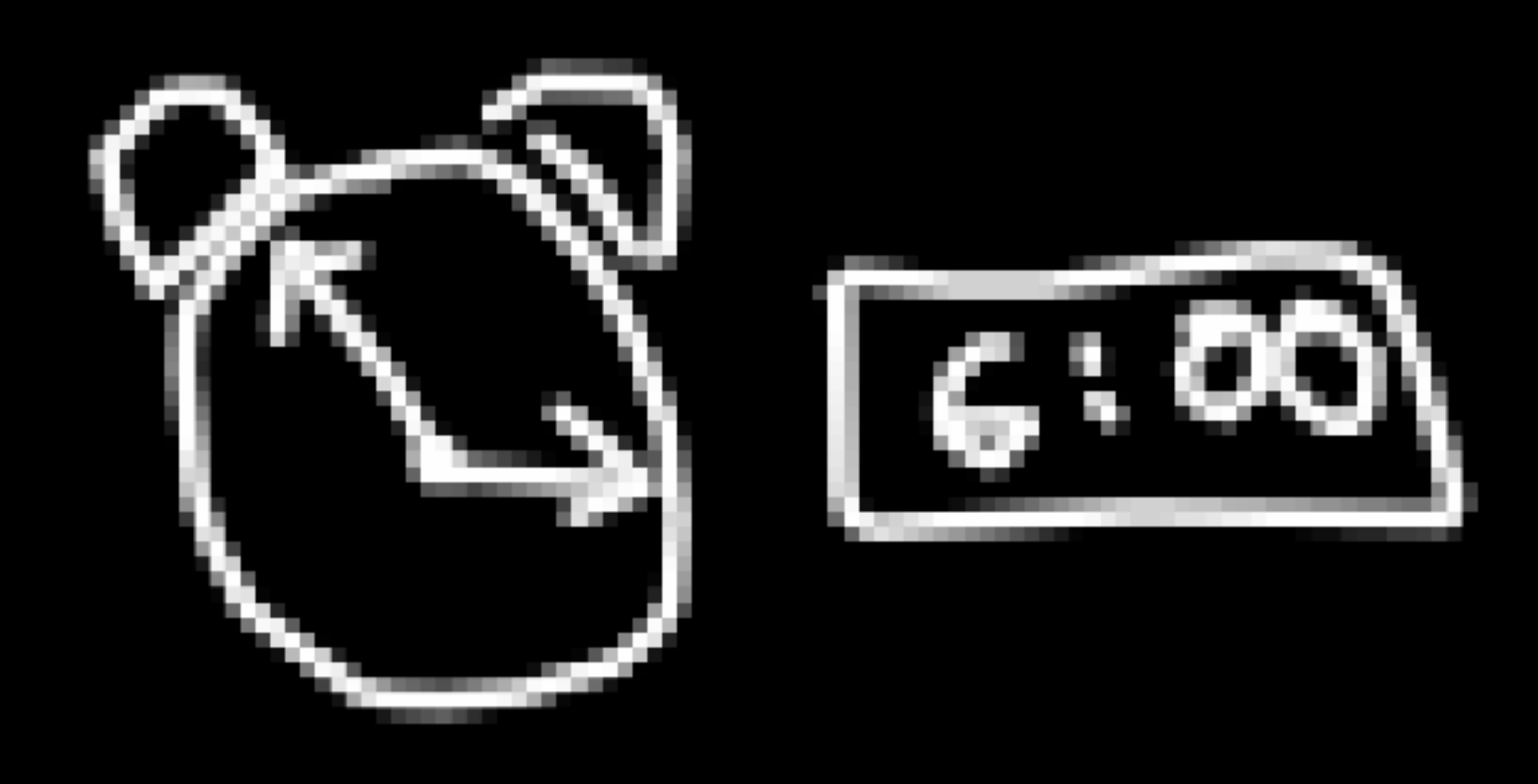}};
\draw [anchor=north west] (0.22\linewidth, 1\linewidth) node {\includegraphics[width=0.2\linewidth]{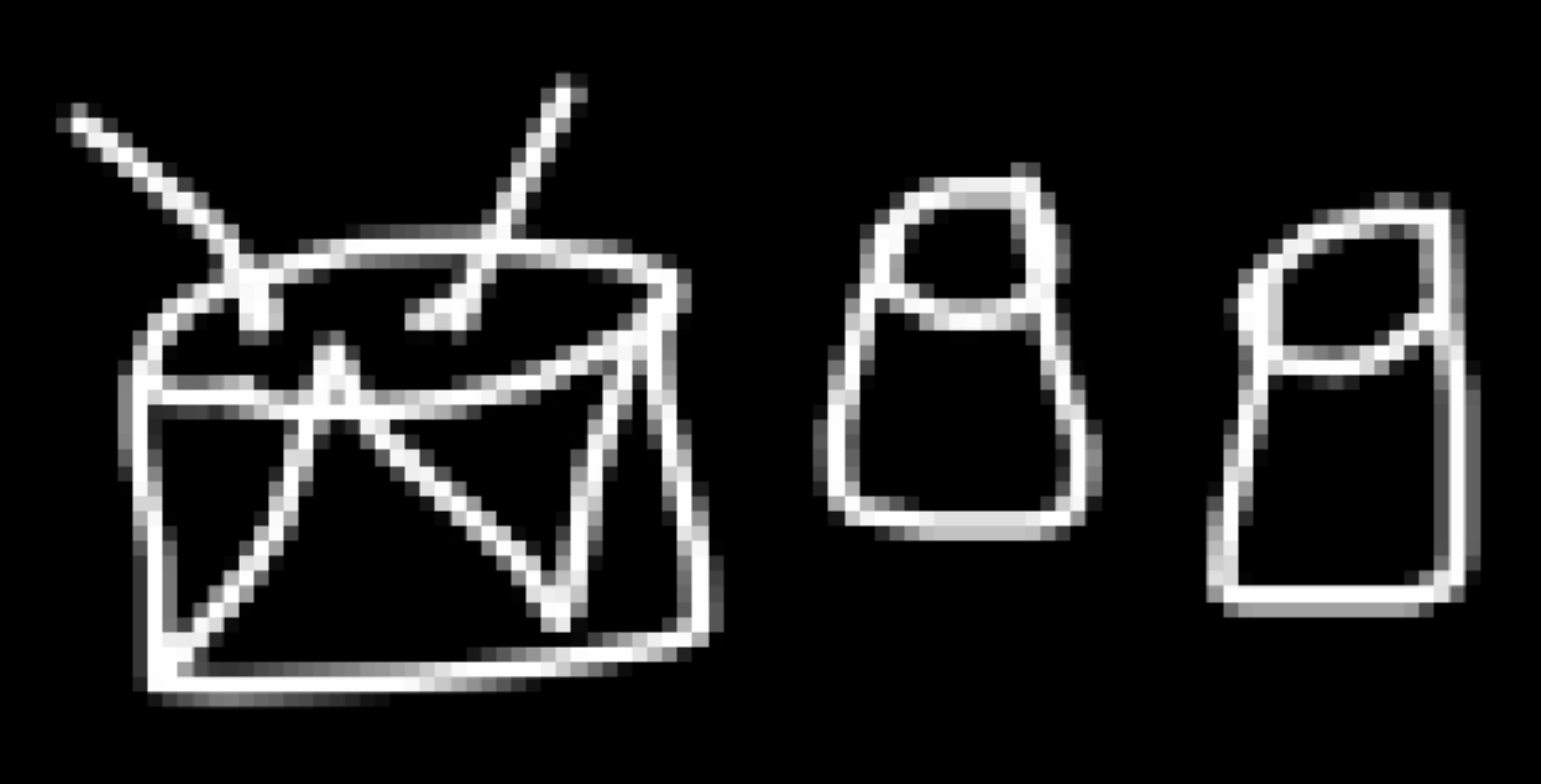}};
\draw [anchor=north west] (0.44\linewidth, 1\linewidth) node {\includegraphics[width=0.2\linewidth]{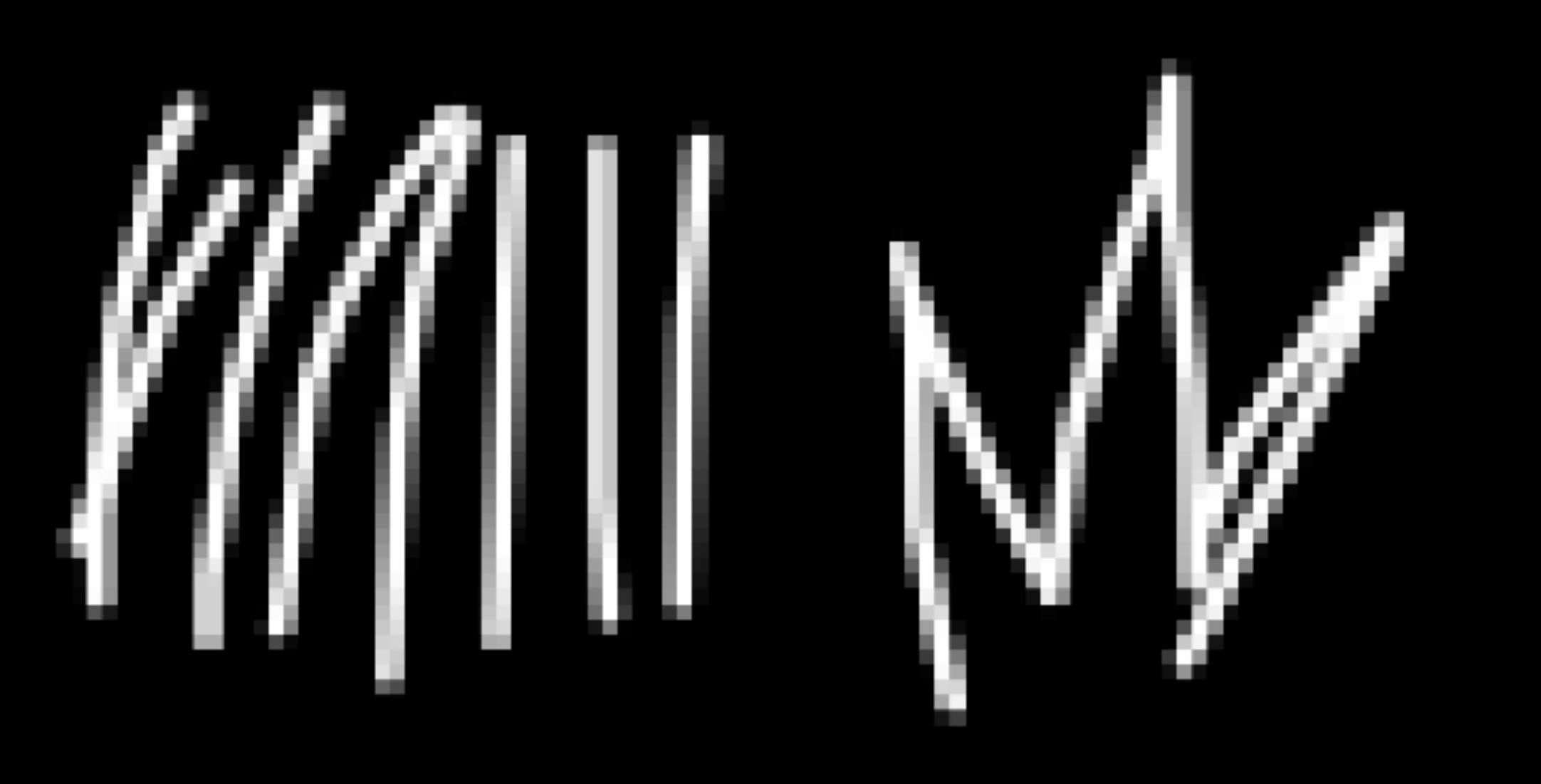}};
\draw [anchor=north west] (0.66\linewidth, 1\linewidth) node {\includegraphics[width=0.2\linewidth]{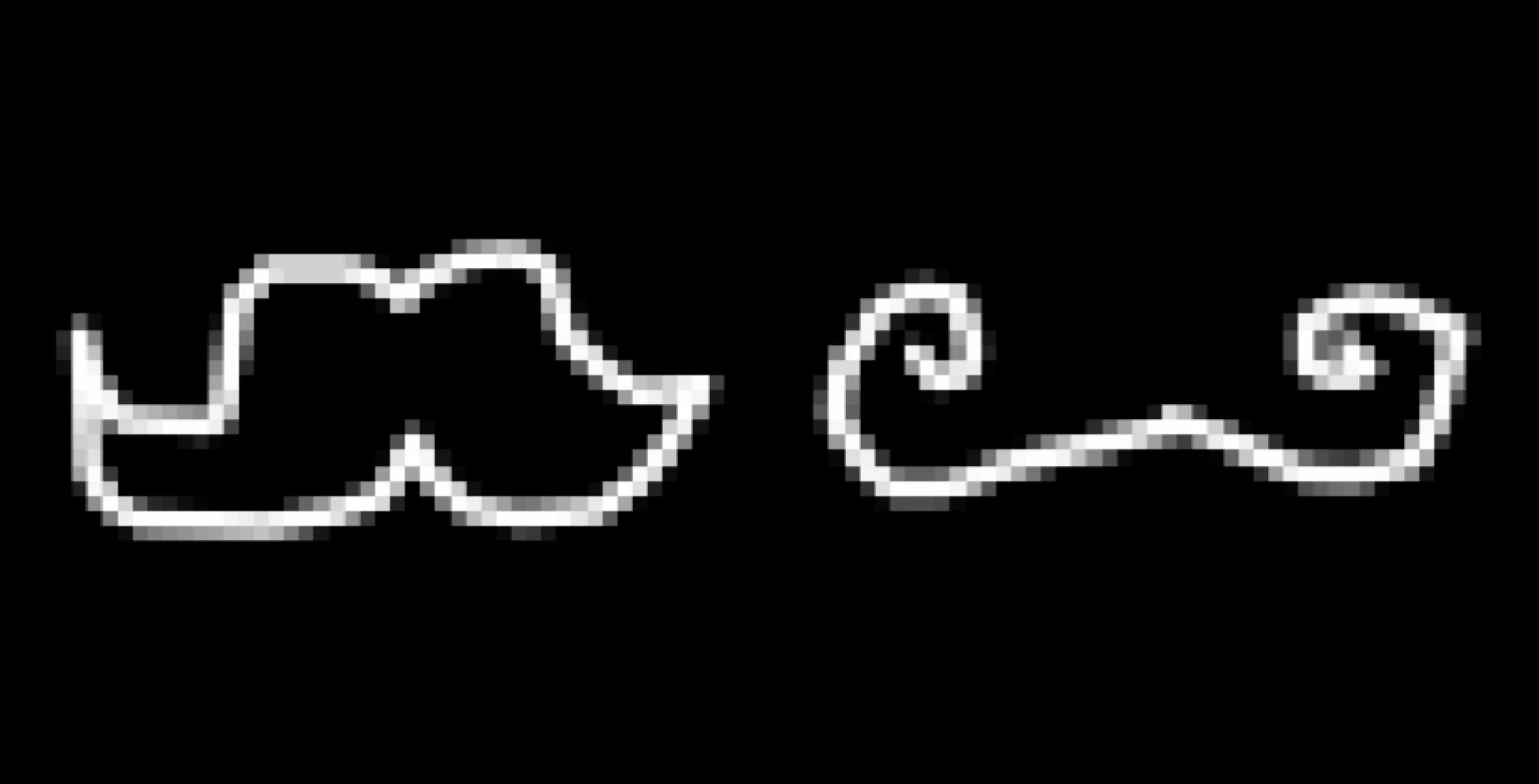}};
\draw [anchor=north west] (0.12\linewidth, 0.85\linewidth) node {\includegraphics[width=0.3\linewidth]{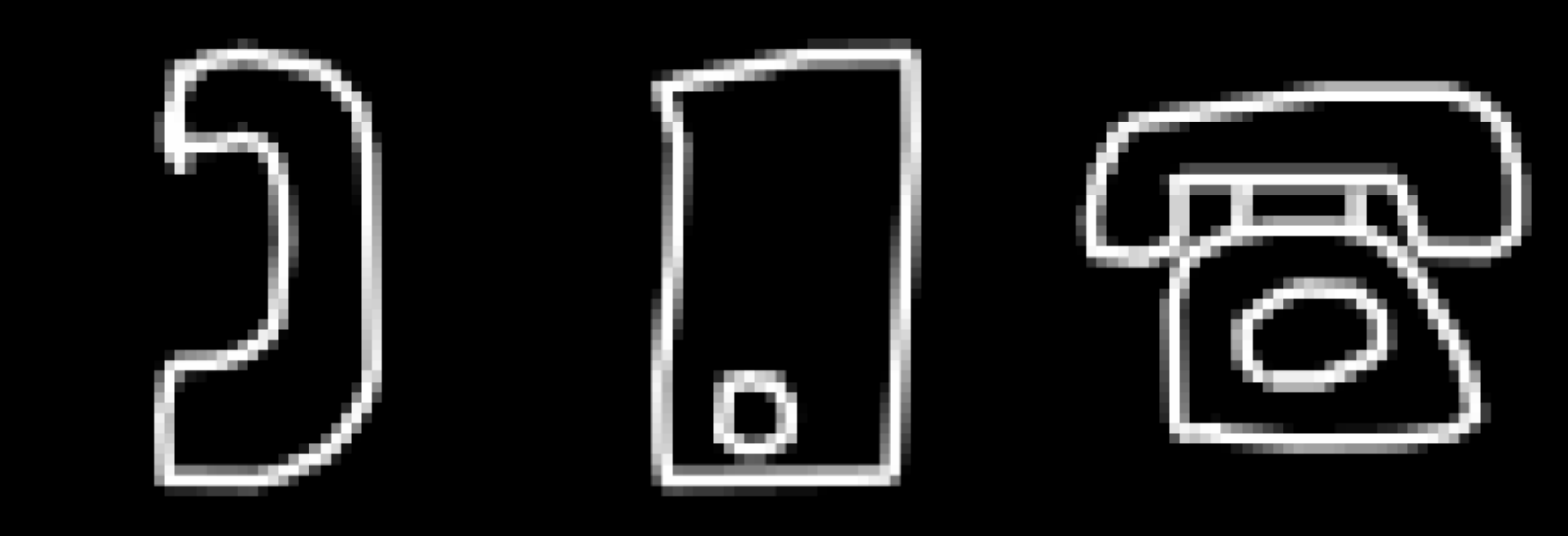}};
\draw [anchor=north west] (0.44\linewidth, 0.85\linewidth) node {\includegraphics[width=0.3\linewidth]{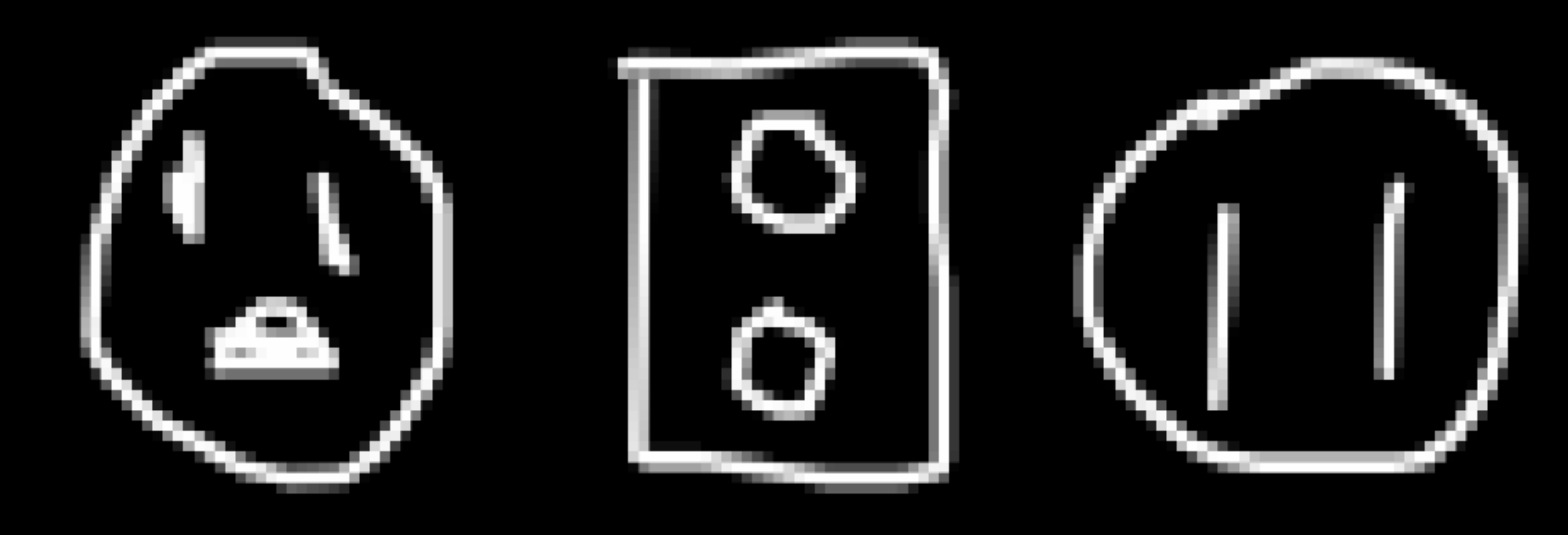}};
\draw [anchor=north,fill=white] (0.11\linewidth, 0.89\linewidth) node {alarm clock};
\draw [anchor=north,fill=white] (0.32\linewidth, 0.89\linewidth) node {drums};
\draw [anchor=north,fill=white] (0.55\linewidth, 0.89\linewidth) node {grass};
\draw [anchor=north,fill=white] (0.76\linewidth, 0.89\linewidth) node {moustache};
\draw [anchor=north,fill=white] (0.28\linewidth, 0.74\linewidth) node {telephone};
\draw [anchor=north,fill=white] (0.6\linewidth, 0.74\linewidth) node {power outlet};
\end{tikzpicture}

\caption{Examples of distinct visual concepts belonging to the same object category}
\label{fig_sup:QD_concepts}
\end{center}
%\vskip -0.5in
\end{figure}

In the \textit{Quick, Draw !} challenge, participants have to draw objects belonging to a specific object category in less than 20 seconds~\cite{jongejan2016quick}. The dataset represents $345$ object categories with approximately $150,000$ samples per category. The total number of drawings in the dataset exceeds $50$ million. The participant is instructed on the object category using words, the object category actually includes more than one unique visual concept. We illustrate this phenomenon in~\cref{fig_sup:QD_concepts} with some object categories composed of more than one visual concept. This property of the original QuickDraw dataset makes it incompatible with the one-shot image generation task because it requires all the samples of a given category to represent the same visual concept.

\subsection{QuickDraw-FS processing steps}
\label{sup:qd_fs_processing}

To circumvent this issue, we have created a new dataset, called QuickDraw-FewShot (QuickDraw-FS), built on the drawings of the \textit{Quick, Draw !} challenge. More specifically, we have re-defined the original QuickDraw object categories so that they correspond to unique visual concepts. Here are the steps we have followed to create the QuickDraw-FS dataset :
\begin{enumerate}
    \item We split the original QuickDraw dataset in a training set made of $5\,175\,000$ drawings (i.e., $345$ categories, $15\,000$ samples each) and a testing set composed of $1\,725\,000$ samples (i.e., $345$ categories, $5\,000$ samples each).

    \item We train a SimCLR feature extractor on the QuickDraw training set. We refer the reader to~\cref{SI:table_SimCLR} in \cref{diversity_recognizability_qd} for more details on the SimCLR architecture.
    
    \item For each object category, we project all the testing samples in the feature space of the SimCLR network.

    \item We then apply a K-Means clustering algorithm on the features extracted previously. More specifically, we extract 6 clusters per object category

    \item We filter out the clusters with less than $500$ samples. This filter prevents us to choose clusters that are not representative enough of the object category.
    
    \item We filter out the clusters with the largest spreading. The spreading size is computed as the mean $\ell_2$-distance between the samples and the cluster center. When the spreading size is above $1\,800$, the cluster is filtered-out. This filter allows us to discard the \textit{junk} clusters, composed of exuberant drawings that are all very different from each other. Note that this filter is triggered very occasionally.

    \item We discard the clusters with centers that are not distant enough from the centers of other clusters. We do so by imposing a minimum $\ell_2$-distance between clusters (set to $700$). This rule prevents us to select $2$ clusters that represent the same visual concept.

    \item For each cluster, we pick an exemplar. The exemplar is selected as being the closest sample to the center of the cluster.

\end{enumerate}

\begin{figure}[ht]
%\vskip 0.2in
\begin{center}

\begin{tikzpicture}
\draw [anchor=north west] (0.2\linewidth, 1\linewidth) node {\includegraphics[width=0.57\linewidth]{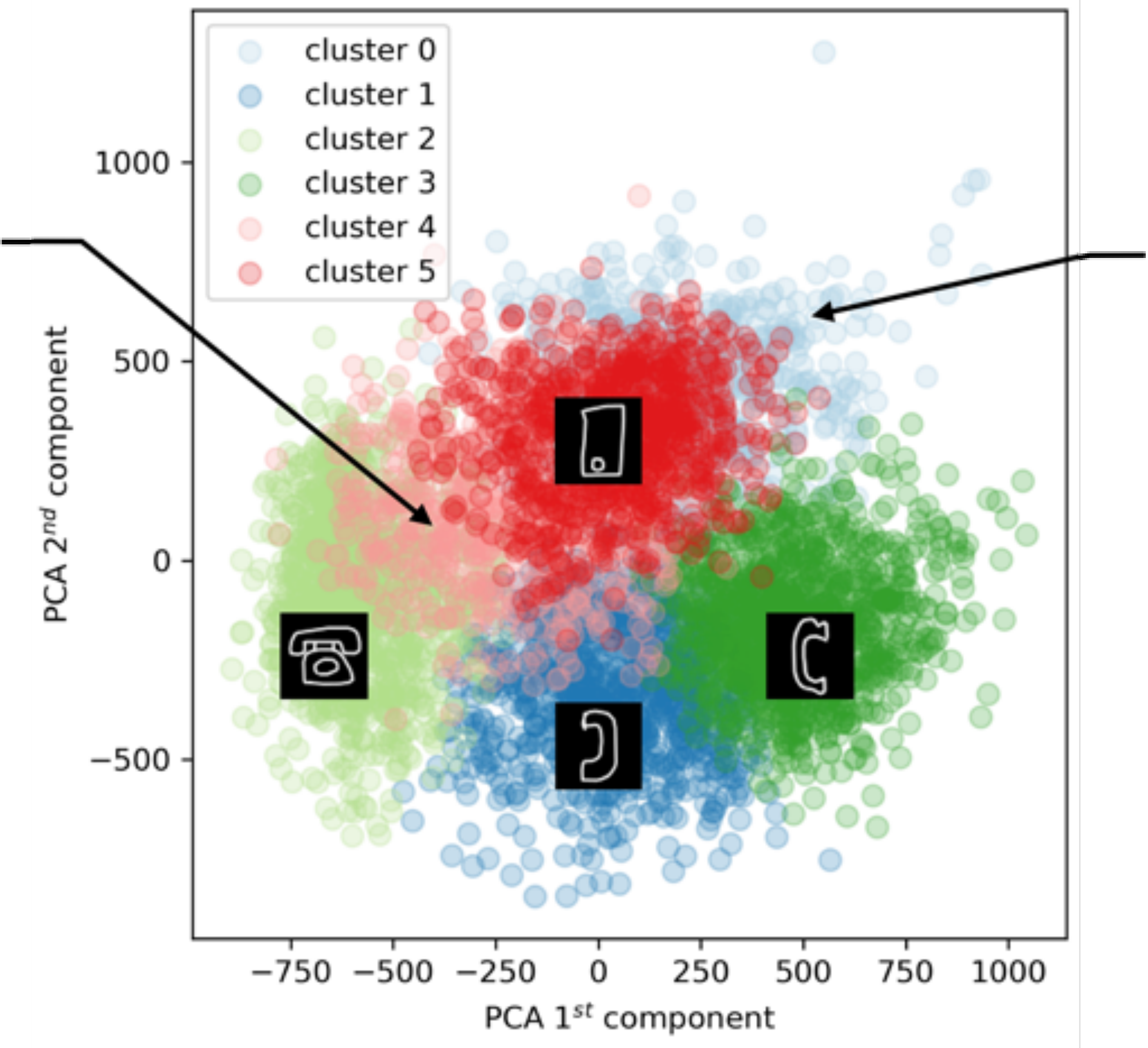}};
\draw [anchor=north,fill=white] (0.1\linewidth, 0.92\linewidth) node[align=left] {The cluster 4 has been \\ filtered-out because it is \\ too close to other cluster' \\ centers. See bullet point 7.};
\draw [anchor=north,fill=white] (0.88\linewidth, 0.92\linewidth) node[align=left] {The cluster 0 has been \\ filtered-out because it is \\ too wide. \\ See preprocessing step 6.};
\end{tikzpicture}

\caption{Illustration of the cluster selection process for the samples of the \textit{phone} object category. The plot represents the PCA coordinates of the samples in the SimCLR feature space. Two clusters were filtered out by the selection process (clusters 4 and 0). The center of the remaining clusters corresponds to the exemplar of distinct visual concepts, as illustrated by the thumbnails.}
\label{fig_sup:QD_cluster}
\end{center}
%\vskip -0.5in
\end{figure}

The filtering and clustering methods described in bullet points $3$ to $8$ are repeated for all object categories of the original QuickDraw dataset.~\cref{fig_sup:QD_cluster} illustrates the selection process for a single object category. 

\begin{figure*}[h!]
%\vskip 0.2in
\begin{center}

\begin{tikzpicture}
\draw [anchor=north west] (0.096\linewidth, 1\linewidth) node {\includegraphics[width=0.048\linewidth]{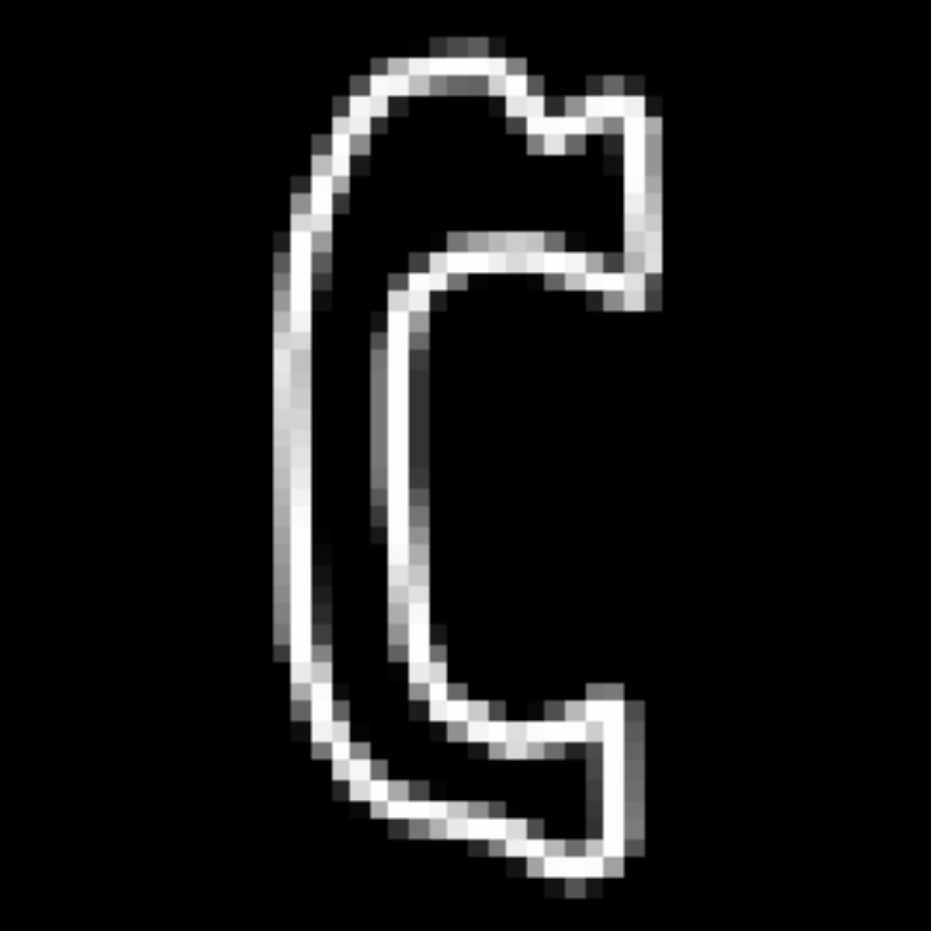}};
\draw [anchor=north west] (0.346\linewidth, 1\linewidth) node {\includegraphics[width=0.048\linewidth]{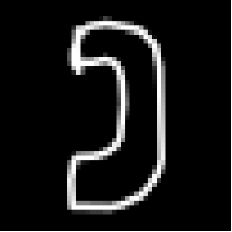}};
\draw [anchor=north west] (0.596\linewidth, 1\linewidth) node {\includegraphics[width=0.048\linewidth]{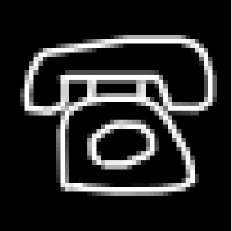}};
\draw [anchor=north west] (0.846\linewidth, 1\linewidth) node {\includegraphics[width=0.048\linewidth]{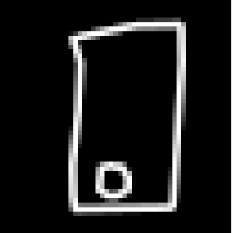}};

\draw [anchor=north west] (0\linewidth, 0.95\linewidth) node {\includegraphics[width=0.24\linewidth]{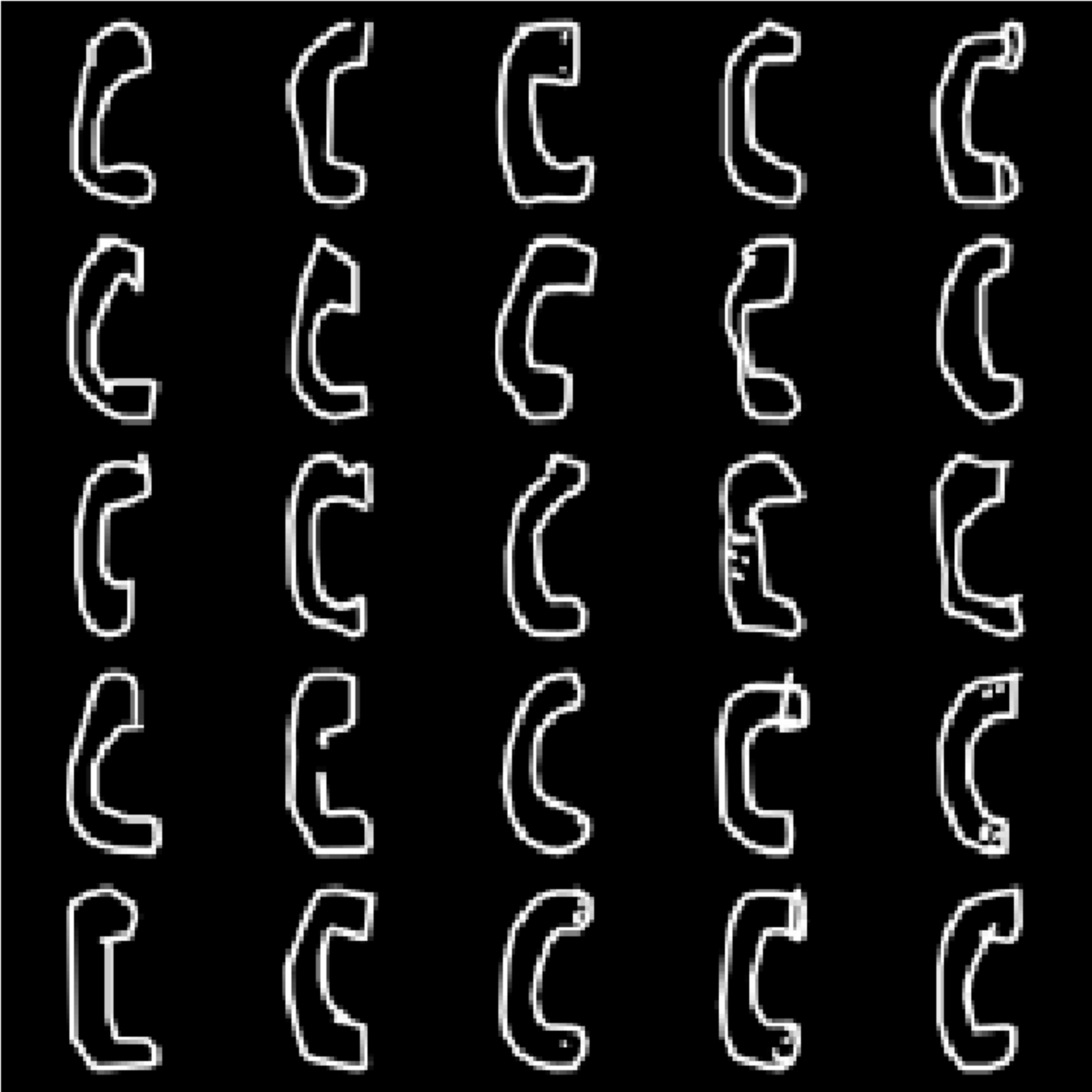}};
\draw [anchor=north west] (0.25\linewidth, 0.95\linewidth) node {\includegraphics[width=0.24\linewidth]{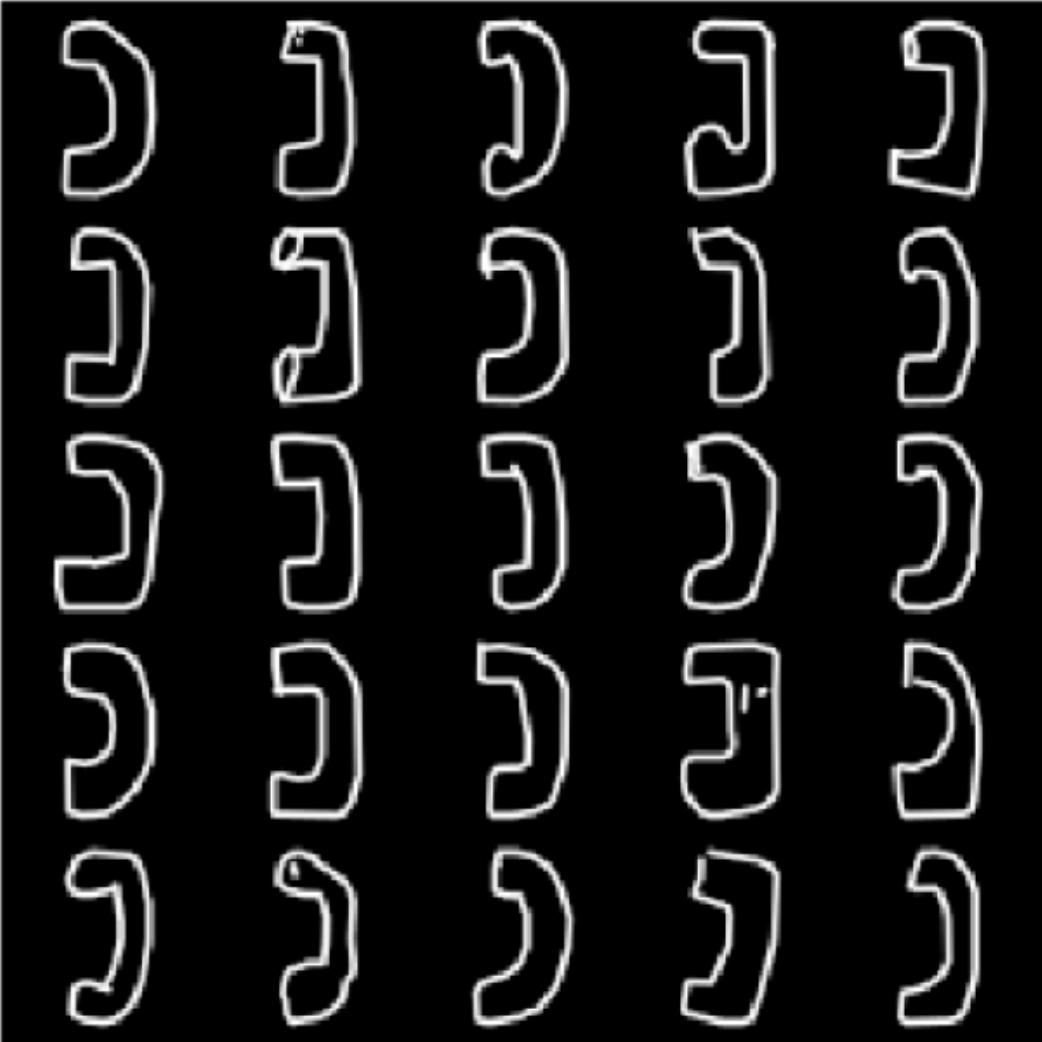}};
\draw [anchor=north west] (0.50\linewidth, 0.95\linewidth) node {\includegraphics[width=0.24\linewidth]{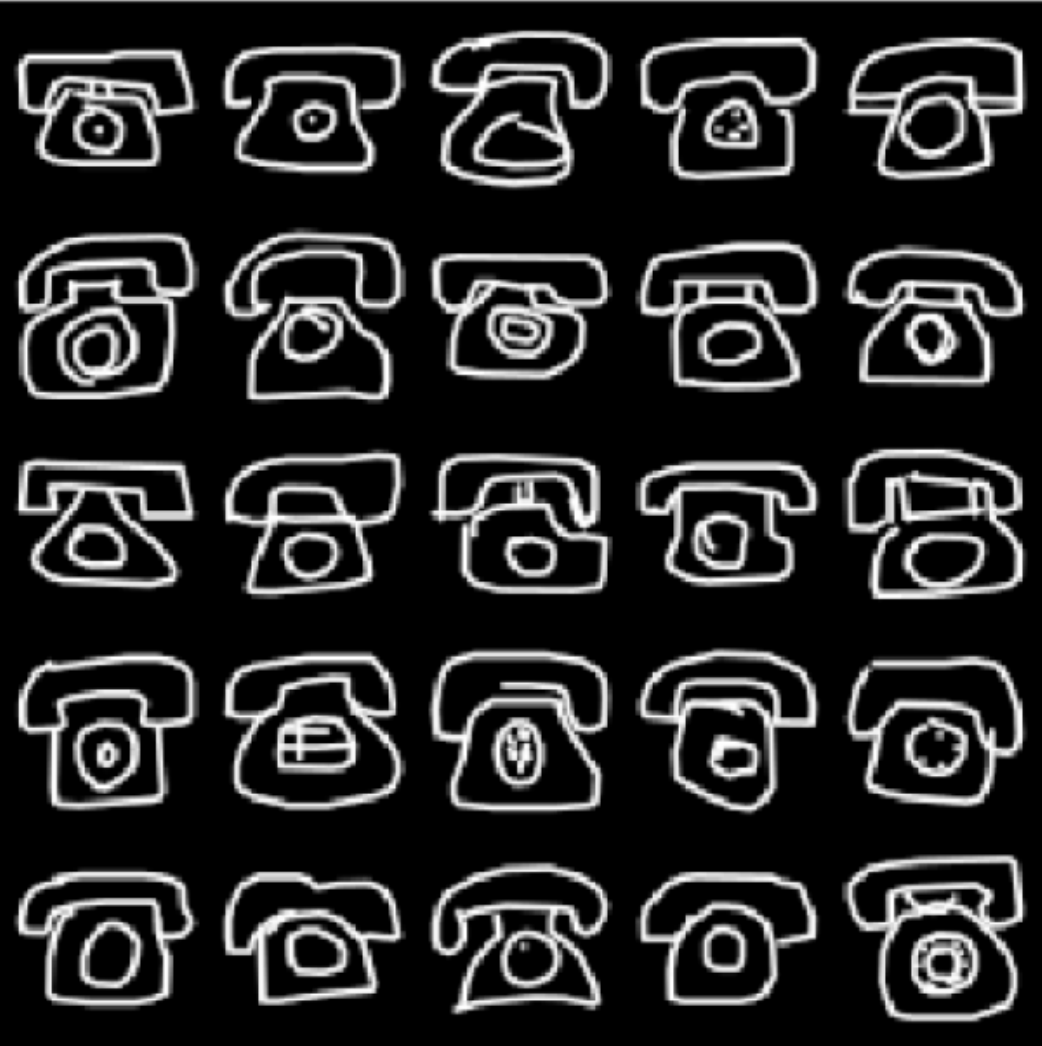}};
\draw [anchor=north west] (0.75\linewidth, 0.95\linewidth) node {\includegraphics[width=0.24\linewidth]{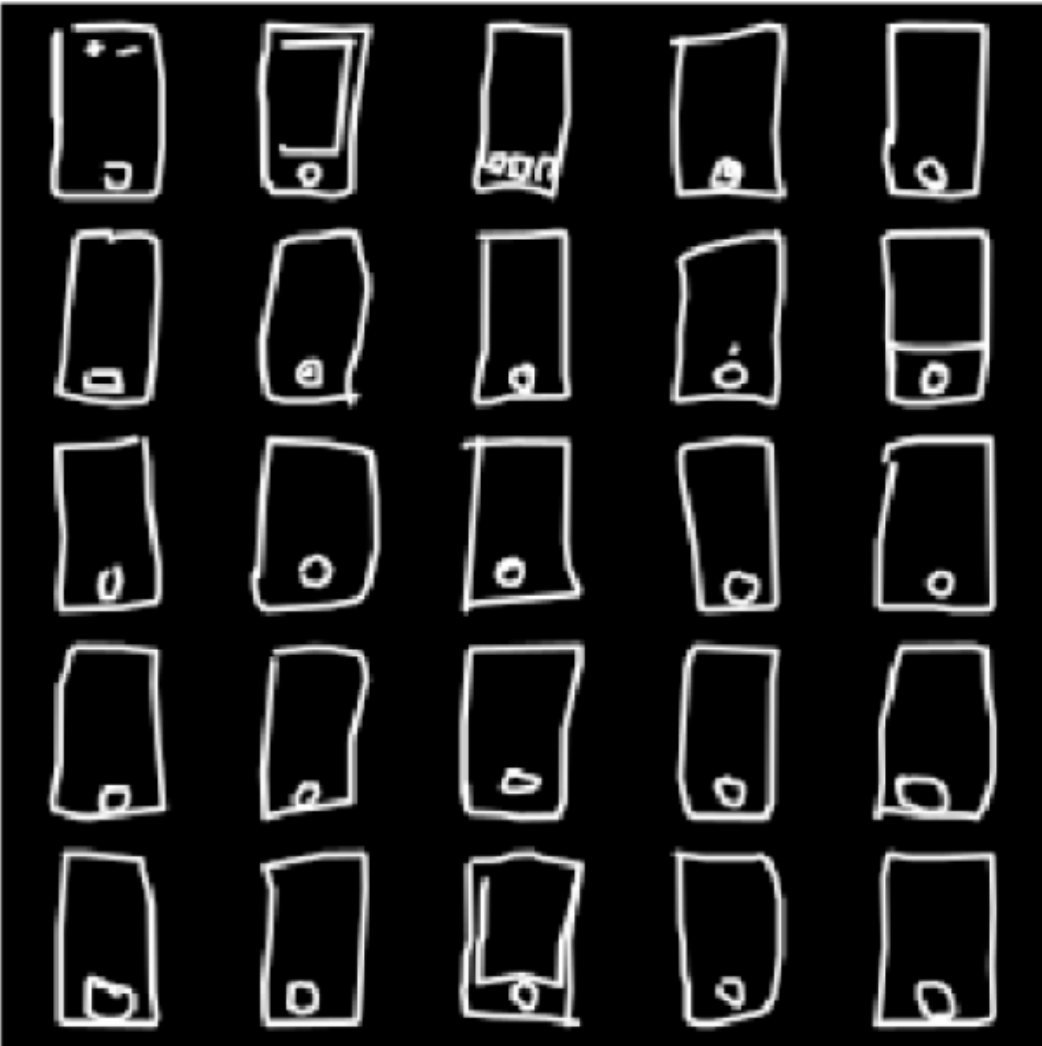}};
\end{tikzpicture}

\caption{Samples and exemplars of the distinct visual concepts extracted through the cluster selection process of the QuickDraw-FS dataset. The top thumbnails represent the exemplars describing the different visual concepts. The exemplars are picked so that they are approximately located at the center of the clusters (in the SimCLR feature space). The $5\times5$ grid of images showcases variations of the corresponding visual concepts. Those variations have been randomly sampled in the clusters. }
\label{fig_sup:QD_samples}
\end{center}
%\vskip -0.5in
\end{figure*}

At the end of the cluster selection process, we perform a visual inspection in which we have filtered out $2$ \textit{junk} clusters (see bullet point 6). The obtained QuickDraw-FS dataset is composed of $332\,000$ drawings ($500$ samples for each of the $665$ distinct visual concepts). The train set is obtained by randomly sampling $550$ visual concepts. The remaining visual concepts constitute the test set. In~\cref{fig_sup:QD_samples}, we showcase randomly selected samples and their corresponding exemplar.

\newpage
\section{Comparison between Omniglot and QuickDraw-FS}
\label{dataset_comp}
\begin{figure}[h!]
%\vskip 0.2in
\begin{center}

\begin{tikzpicture}
\draw [anchor=north west] (0\linewidth, 1\linewidth) node {\includegraphics[width=0.5\linewidth]{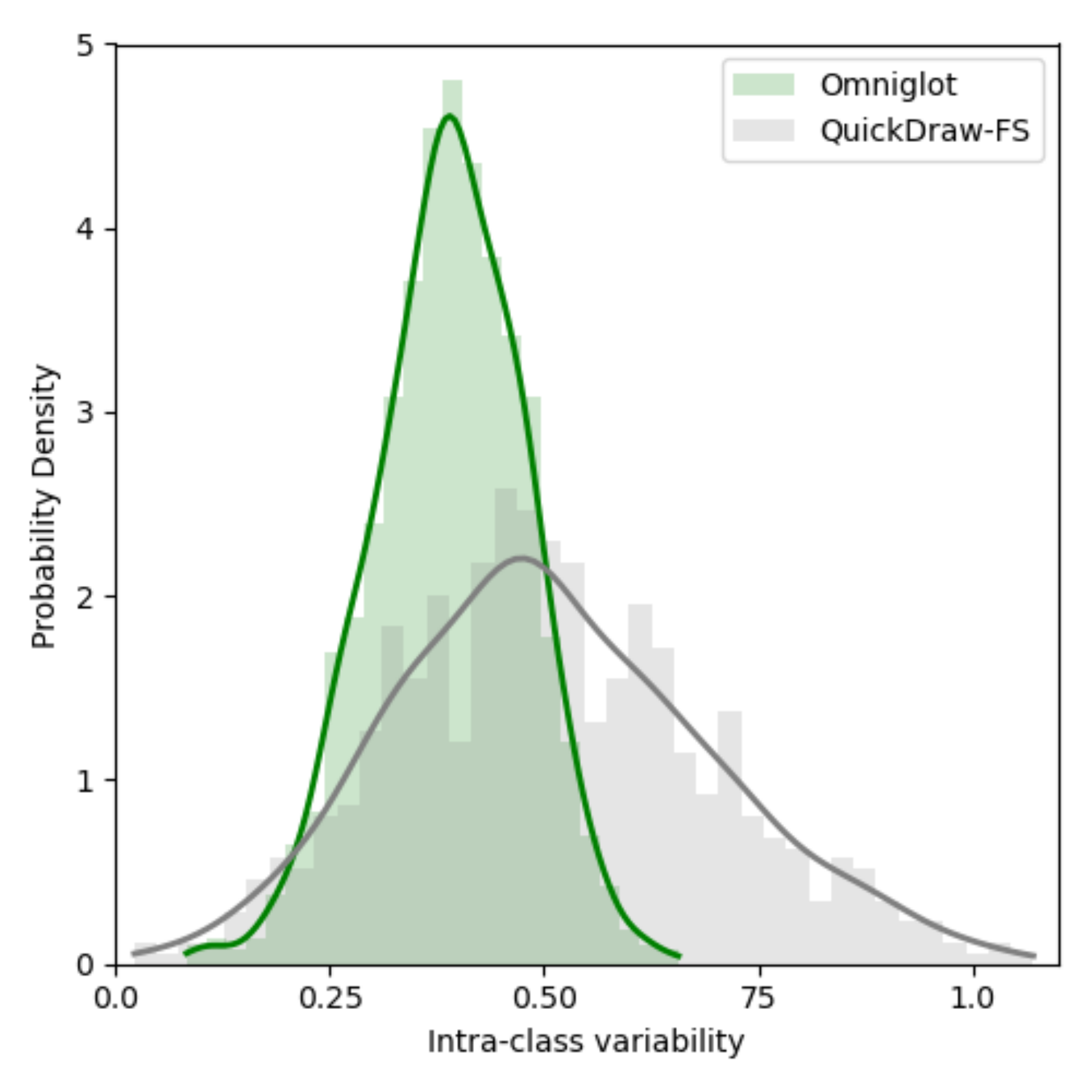}};
\end{tikzpicture}
\caption{Probability density of the (normalized) intra-class variability of the Omniglot (green) and the QuickDraw-FS dataset (grey). The intra-class variability is computed in the latent space of a SimCLR network, as the intra-class standard deviation.}
\label{fig_sup:comp_QD_om}
\end{center}
%\vskip -0.5in
\end{figure}

Herein we compare the intra-class variability of the Omniglot samples with the intra-class variability of the QuickDraw-FS samples. We refer the reader to \cref{sup:QD_dataset} for more information on the QuickDraw-FS dataset. We show the distributions of the intra-class variability in \cref{fig_sup:comp_QD_om}. To obtain such a distribution, we (i) pass the samples into a feature extractor network (a SimCLR network), and (ii) compute the standard deviation for samples belonging to the same class. To have a faithful comparison between the $2$ datasets, we normalized the features vector so that the standard deviation, across features, is set to one (see \cref{diversity_normalization} for the reasons of such a normalization).

We observe that the grey distribution covers a wider range of intra-class variability compared to the green distribution. The average intra-class variability is $0.37$ for Omniglot and $0.48$ for QuickDraw-FS. The maximum intra-class variability is $0.66$ for Omniglot and $1.10$ for QuickDraw-FS. It suggests that (i) the QuickDraw-FS samples are more diverse than those of the Omniglot dataset and (ii) the Omniglot samples do not reflect
the human ability to produce original and diverse drawings.

This phenomenon could be explained by the way the Omniglot samples have been collected. Participants are presented with category exemplars and asked to draw, as accurately as possible, a replica of the exemplar (see Appendix 1 in \citealp{lake2015human}). This experimental protocol implicitly reduces the variability of the samples. For the QuickDraw dataset, the participants are presented with a word describing the object and are asked to draw the objects without other constraints. Even if the filtering process we have performed to obtain the QuickDraw-FS dataset should reduce the samples' intra-class variability (see \cref{sup:QD_dataset}), it is still higher than the Omniglot intra-class variability.

\newpage
\section{Diversity vs Recognizability framework for the QuickDraw-FS dataset}
\label{diversity_recognizability_qd}
The diversity versus recognizability framework leverages $2$ critic networks to i) extract the features to compute the diversity metric, and ii) evaluate the one-shot classification accuracy for the recognizability. Similarly to~\citealp{boutin2022diversity},  we use a SimCLR network~\cite{chen2020simple} to extract features and we leverage a Prototypical Net~\cite{snell2017prototypical} for the computation of the recognizability. We have increased the size of the critics' network architecture compared to~\citealp{boutin2022diversity} to adapt to the complexity of the QuickDraw-FS dataset. Here we describe the new architectures of both the SimCLR and Prototypical Network

\subsection{SimCLR on QuickDraw-FS}
\cref{SI:table_SimCLR} describes the architecture of the SimCLR network~\citep{chen2020simple} trained on the QuickDraw-FS dataset. We use the Pytorch convention to describe the layers of the network.

\begin{table}[h!]
  \caption{Description of the SimCLR Architecture}
  \centering
  \begin{tabular}{ccc}
    \toprule
    Network & Layer & \# params \\
    \midrule
    \multirow{4}{*}{ConvBlock(In$_{c}$, Out$_{c}$)} & Conv2d(In$_{c}$, Out$_{c}$, 3, padding=1)    &    In$_{c}$ $\times$ Out$_{c}$ $\times$ 3 $\times$ 3 +  Out$_{c}$ \\
    & BatchNorm2d(Out$_{c}$)  & 2 x Out$_{c}$ \\
    & ReLU & - \\
    & MaxPool2d(2, 2)  & - \\
    \midrule
    \multirow{8}{*}{SimCLR} & ConvBlock(1, 64) & 0.7 K \\ 
    & ConvBlock(64, 128) & 74 K\\
    & ConvBlock(128, 128) & 149 K\\
    & Flatten & - \\
    & ReLU & - \\
    & Linear(4608, 256) & 1 179 K\\
    & ReLU & \\
   & Linear(256, 128) & 32 K\\
   \bottomrule
  \end{tabular}
\label{SI:table_SimCLR}
\end{table}

The overall number of parameters of the SimCLR network we are using is around 1.4 M parameters. The features are extracted on the first fully-connected layer after the last convolutional layer (i.e., of size $256$).

All other implementation details (augmentation, training parameters) are the same as those described in Appendix S2 of ~\citealp{boutin2022diversity}.

\newpage
\subsection{Prototypical Net on QuickDraw-FS}
For the Prototypical Net trained on QuickDraw-FS, we leverage a ResNet-like architecture~\cite{he2016deep}. This architecture is described in~\cref{SI:table_ProtoNet}.

\begin{table}[h!]
  \caption{Description of the Prototypical Net Architecture}
  \centering
  \begin{tabular}{ccc}
    \toprule
    Network & Layer & \# params \\
    \midrule
    \multirow{9}{*}{ResNetBlock(In$_{c}$, Out$_{c}$)} & Conv2d(In$_{c}$, Out$_{c}$, 3, padding=1)    &    In$_{c}$ $\times$ Out$_{c}$ $\times$ 3 $\times$ 3 +  Out$_{c}$ \\
    & BatchNorm2d(Out$_{c}$)  & 2 x Out$_{c}$ \\
    & ReLU & - \\
    & Conv2d(Out$_{c}$, Out$_{c}$, 3, padding=1) &    Out$_{c}$ $\times$ Out$_{c}$ $\times$ 3 $\times$ 3 +  Out$_{c}$ \\
    & BatchNorm2d(Out$_{c}$)  & 2 x Out$_{c}$ \\
    & \textcolor{gray}{ \#\# Shortcut connection} & \\
    & Conv2d(In$_{c}$, Out$_{c}$, 1, padding=1) &    In$_{c}$ $\times$ Out$_{c}$ +  Out$_{c}$ \\
    & BatchNorm2d(Out$_{c}$)  & 2 x Out$_{c}$ \\
    & ReLU & - \\
    \midrule
    \multirow{14}{*}{Prototypical Net} & Conv2d(1, 64, 3, padding=1)  &    0.6 K \\
    & BatchNorm2d(128)  & 0.12 K \\
    & ReLU & - \\
    & ResNetBlock(64, 64) & 78 K \\
    & ResNetBlock(64, 64) & 78 K \\
    & ResNetBlock(64, 128) & 230 K \\
    & ResNetBlock(128, 128) & 312 K \\
    & ResNetBlock(128, 256) & 919 K \\
    & ResNetBlock(256, 256) & 1 246 K \\
    & ResNetBlock(256, 512) & 3 673 K \\
    & ResNetBlock(512, 512) & 4 984 K \\
    & AvgPool2d(6, 6 ) & - \\
    %& Flatten() & - \\
    & Linear(512, 256) & 131 K \\
   & Linear(256, 128) & 32 K\\
   \bottomrule
  \end{tabular}
\label{SI:table_ProtoNet}
\end{table}

The overall number of parameters of the Prototypical Net we are using is around $11.6$ M parameters. The loss of the Prototypical Net is applied to the output of the last fully connected layers (of size $128$).

To prevent over-fitting and to adapt to the variability of the QuikDraw-FS dataset, we have randomly applied the following augmentation: first a horizontal flip, then a vertical flip, and last an affine transformation. The affine transformation is a combination of a rotation (with an angle randomly selected in the range $[-180^{\circ}, 180^{\circ}]$), a translation (randomly selected in $[-10\textnormal{px}, 10 \textnormal{px}]$), a zoom (with a ratio randomly selected $[0.5, 1.5]$). Note that the augmentation is applied similarly to all the samples belonging to the same class so that it is virtually increasing the number of classes of the dataset.

All other training parameters are similar to those described by~\citealp{boutin2022diversity} in Appendix S1.

The code for training both critic networks on the QuickDraw-FS dataset is available online at \url{https://anonymous.4open.science/r/Diffusion_vs_human/}.

\clearpage
\newpage
\section{Effect of the normalization on the diversity metric}
\label{diversity_normalization}
The diversity is obtained by computing a dispersion metric in the feature space. More specifically we use a Bessel-corrected standard deviation, applied in the latent space of a SimCLR network~\cite{chen2020simple}. This is the exact same setting as the one described in \citealp{boutin2022diversity}. 

This way of computing the diversity metric has $2$ main drawbacks: i) it is unbounded, and ii) it depends on the image size and on the size of the feature space of the SimCLR network. If we compare models on the same dataset, with a unique setting of the SimCLR network, those limitations are not problematic. But those drawbacks prevent us to compare diversity values on different datasets (and thus different SimCLR settings). 

To circumvent these problems, we normalize the values of the feature vector so that its standard deviation (in the feature space) is set to one for each individual sample. This is important to note that this normalization is performed using a standard deviation computed along the features coordinate. In this case, the standard deviation quantifies the dispersion of the feature activation. In the calculation of the diversity value, we also use the standard deviation, but that one is computed along the sample axis. Consequently, the standard deviation used to compute the diversity quantifies the dispersion of the sample (in a given category).

\begin{figure*}[ht]
%\vskip 0.2in
\begin{center}
%\centerline{
\includegraphics[width=0.5\columnwidth]{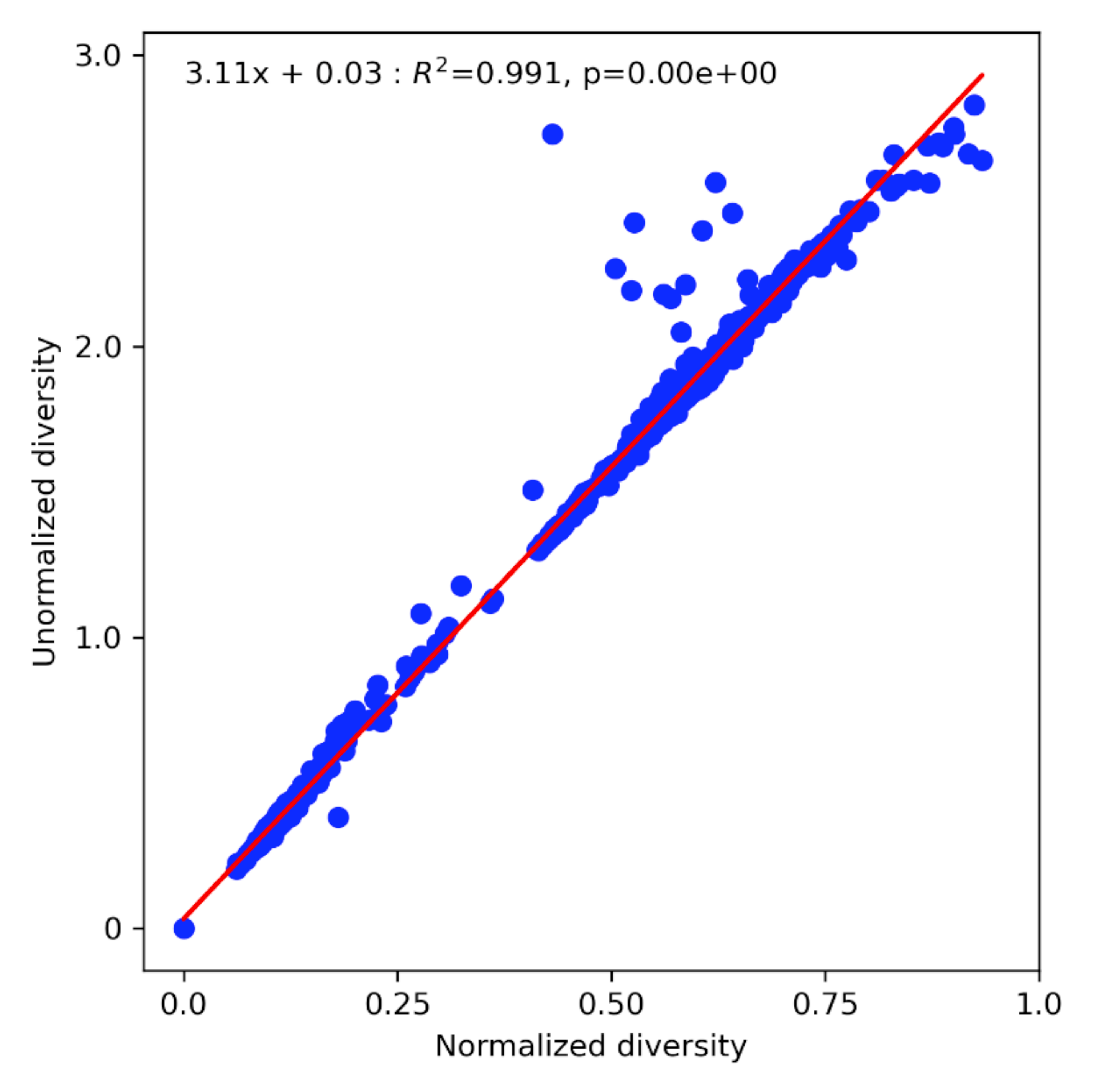}
%}
\caption{Control experiment to compare the effect of feature normalization in the computation of the diversity value. Each data point corresponds to the mean diversity for a single model. In this graph, we have included models trained on Omniglot and QuickDraw-FS.}
\label{fig_sup:diversity_normalization}
\end{center}
%\vskip -0.5in
\end{figure*}

We run a control experiment to verify that the proposed feature normalization is not
changing the model's relative position on the diversity axis. To do so, we plot the models' diversity when the features are normalized (x-axis) or not (y-axis) (see \cref{fig_sup:diversity_normalization}). We report a linear correlation of $R^2$ = 0.99 and a Spearman rank-order correlation of $\rho$ = $0.99$.

The high linear correlation and Spearman rank-order correlation suggest that the normalization operation is not changing the models' relative position on the diversity axis. Therefore, the proposed normalization method allows us to i) maintain the relative position of the model within a given SimCLR setting, and ii) compare models evaluated with different SimCLR settings.

\newpage
\section{Mathematics behind diffusion process}
\subsection{Diffusion process parametrization}
\label{sup:parametrization_diffusion}
Herein we detail the mathematics behind the diffusion models. Most of the demonstrations below are inspired by other works ~\cite{song2019generative, sohl2015deep, ho2020denoising} and are adapted to the few-shot image generation scenario. Even though those mathematical derivations are not crucial for a good understanding of our work, we include them to make sure our article is self-contained and complete.

\begin{figure}[ht]
\begin{center}
%\centerline{
\includegraphics[width=0.4\columnwidth]{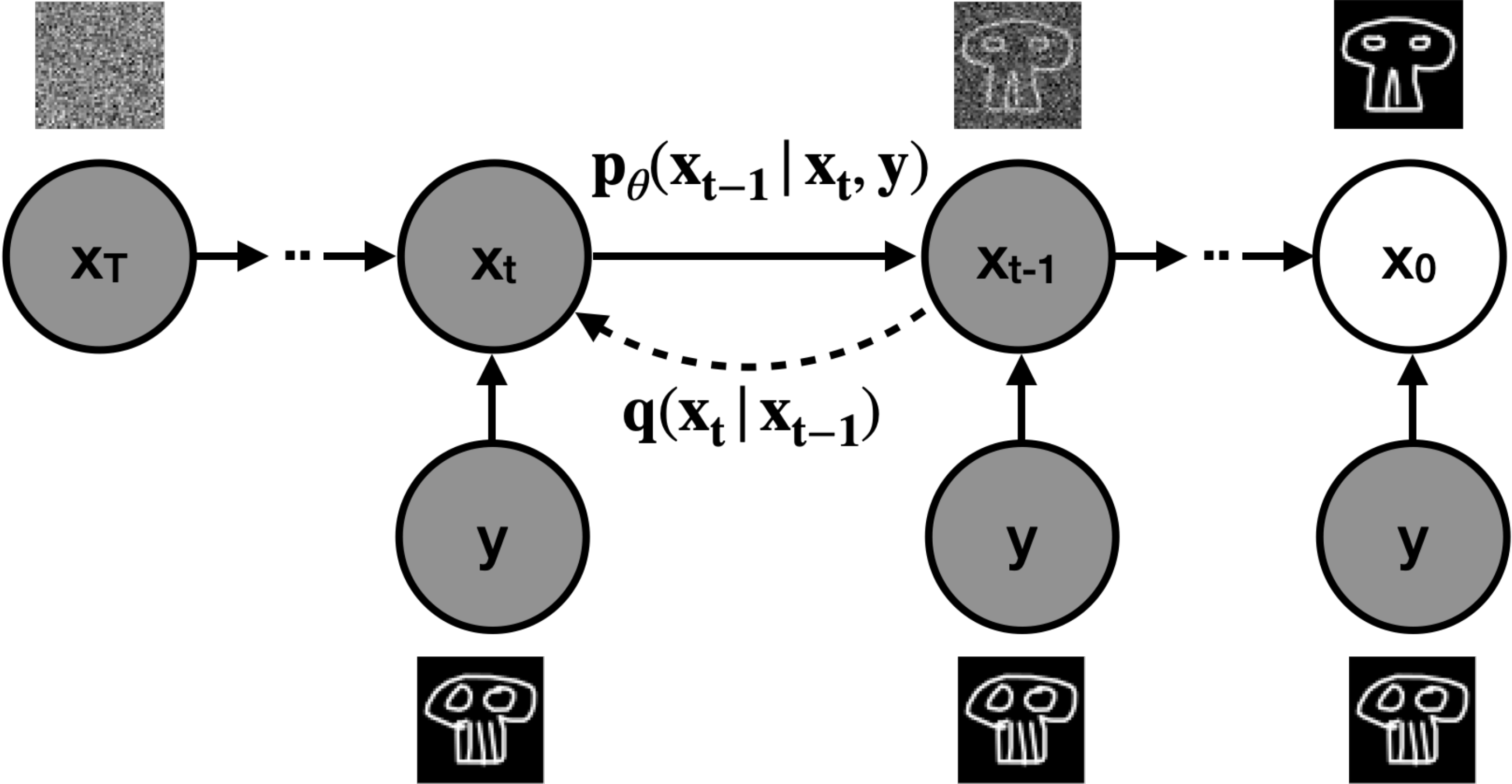}
%}
\caption{The directed graphical model considered in this work. Dotted and plain arrows represent the forward and reverse processes, respectively. The random variable $\mathbf{y}$ is exemplified by the skull exemplar (see bottom thumbnails), and the  $\mathbf{x_i}$ latent variables are exemplified using skull samples with varying noise levels (see top thumbnails).}
\label{fig:diffusion_process}
\end{center}
\vskip -0.1in
\end{figure}

A diffusion process describes the transformation of a pure noise $\mathbf{x}_{T} \in \mathbb{R}^{D}$ to an observed data $\mathbf{x_0} \in \mathbb{R}^{D}$ through a sequence of latent variables $\{\mathbf{x_i}\}_{i=1}^{T-1} \in \mathbb{R}^{D\times(T-1)}$. Diffusion models include a forward process modeling the transition probability $p_{\theta}(\mathbf{x_{t-1}}|\mathbf{x_{t}}, \mathbf{y})$ and a reverse process that parameterize $q(\mathbf{x_t}|\mathbf{x_{t-1}})$. The directed graphical model under consideration is shown in \cref{fig:diffusion_process}. 

The forward process is parametrized as follow~\cite{ho2020denoising}:
\begin{align} \label{sup:forward_diffusion_process}
q(\mathbf{x}_{1:T}|\mathbf{x}_{0}) = \prod_{t=1}^T q(\mathbf{x}_{t}| \mathbf{x}_{t-1}) \quad \text{with} \quad q(\mathbf{x}_{t}|\mathbf{x}_{t-1}) = \mathcal{N}(\mathbf{x}_{t};\sqrt{1-\beta_t}\mathbf{x}_{t-1}, \beta_{t}\textbf{I}) \quad \text{s.t.} \quad \{\beta_t \in (0,1)\}_{i=1}^{T}
\end{align}
In \cref{sup:forward_diffusion_process}, $\beta_t$ controls the step size of the diffusion process. Using the successive product of Gaussian, one can reparametrize $\mathbf{x}_t$ to express it without referring to the intermediate latent variables $\{\mathbf{x_i}\}_{i=1}^{t-1}$:
\begin{align} 
\mathbf{x}_{t} & = \sqrt{\alpha_t}\mathbf{x}_{t-1} + \sqrt{1 - \alpha_t}\mathbf{\epsilon} \quad \text{with} \quad \mathbf{\epsilon} \sim \mathcal{N}(\mathbf{0}, \textbf{I}) \nonumber \\
& = \sqrt{\alpha_{t}\alpha_{t-1}}\mathbf{x}_{t-2} \sqrt{1 - \alpha_{t} \alpha_{t-1}}\mathbf{\epsilon} \nonumber \\ 
& = \text{...}  \nonumber \\
& = \sqrt{\bar{\alpha}_t}\mathbf{x}_{0} + \sqrt{1 - \bar{\alpha}_t}\mathbf{\epsilon} \quad \text{with} \quad \alpha_t = 1 - \beta_t \quad \text{and} \quad \bar \alpha_t = \prod_{i=1}^t \alpha_t 
\label{sup:reparametrization}
\end{align}

Consequently, we have:
 
\begin{align} \label{sup:forward_diffusion_process_param} 
q(\mathbf{x}_{t}|\mathbf{x}_{0}) = \mathcal{N}(\mathbf{x}_t;\sqrt	{\bar{\alpha}_t}\mathbf{x}_0, (1-\bar{\alpha}_t)\textbf{I} ) 
\end{align}

The reverse process, also called the generative process, is conditioned on the exemplar $\mathbf{y}$ to recover the data from the noise~\cite{ho2020denoising}:
\begin{align} \label{sup:reverse_diffusion_process}
p_{\theta}(\mathbf{x}_{0:T}| \mathbf{y}) = p_{\theta}(\mathbf{x}_{T}| \mathbf{y}) \prod_{t=1}^{T} p_{\theta}(\mathbf{x}_{t-1}| \mathbf{x}_{t},  \mathbf{y}) \quad \text{with} \quad \left\{ \begin{array}
{ll}
p_{\theta}(\mathbf{x}_{t-1}|\mathbf{x}_{t}, \mathbf{y}) &= \mathcal{N}(\mathbf{x}_t;\mathbf{\mu}_{\theta}(\mathbf{x}_t,t,\mathbf{y}),\sigma^{2}_{t} \textbf{I}) \\
p_{\theta}(\mathbf{x}_{T}| \mathbf{y}) &= p(\mathbf{x}_{T}) = \mathcal{N}(\mathbf{0},\textbf{I})
\end{array}
\right.
\end{align}

\subsection{From variational lower bound to auto-encoder optimization}
\label{sup:ddpm_loss}
We first express the Variational Lower Bound of the diffusion model using Jensen's inequality~\cite{ho2020denoising}:
\begin{align}
\mathbb{E}_{\mathbf{x}_{0}\sim q(\mathbf{x}_{0})}\log p_{\theta}(\mathbf{x}_{0}| \mathbf{y}) &=  \mathbb{E}_{\mathbf{x}_{0}\sim q(\mathbf{x}_{0})} \log \big( \displaystyle \int p_{\theta}(\mathbf{x}_{0:T}| \mathbf{y}) d\mathbf{x}_{1:T}\big) \nonumber\\
&= \mathbb{E}_{\mathbf{x}_{0}\sim q(\mathbf{x}_{0})} \log \big( \displaystyle \int q(\mathbf{x}_{1:T}|\mathbf{x}_{0}) \frac{p_{\theta}(\mathbf{x}_{0:T}| \mathbf{y})}{q(\mathbf{x}_{1:T}|\mathbf{x}_{0}) } d\mathbf{x}_{1:T}\big) \nonumber \\
&= \mathbb{E}_{\mathbf{x}_{0}\sim q(\mathbf{x}_{0})} \log \Bigg( \mathbb{E}_{\mathbf{x}_{1:T}\sim q(\mathbf{x}_{1:T}|\mathbf{x}_{0})} \displaystyle \Big[\frac{p_{\theta}(\mathbf{x}_{0:T}| \mathbf{y})}{q(\mathbf{x}_{1:T}|\mathbf{x}_{0}) }\Big]\Bigg) \nonumber \\
&\leq \mathbb{E}_{\mathbf{x}_{0:T}\sim q(\mathbf{x}_{0:T})} \log \Big( \displaystyle \frac{p_{\theta}(\mathbf{x}_{0:T}| \mathbf{y})}{q(\mathbf{x}_{1:T}|\mathbf{x}_{0}) } \Big) = - L_{VLB} \nonumber
 \end{align}
 
 The Variational Lower Bound could be written as a sum of $KL$ terms~\cite{sohl2015deep}:
\begin{align}
L_{VLB} & = \mathbb{E}_q \Big[ \log \displaystyle \frac{q(\mathbf{x}_{1:T}|\mathbf{x}_{0}) }{p_{\theta}(\mathbf{x}_{0:T}| \mathbf{y})} \Big] \nonumber \\
&=\mathbb{E}_q \Big[ \log \displaystyle \frac{\prod_{t=1}^T q(\mathbf{x}_{t}| \mathbf{x}_{t-1})}{p(\mathbf{x}_{T}| \mathbf{y}) \prod_{t=1}^{T} p_{\theta}(\mathbf{x}_{t-1}| \mathbf{x}_{t}, \mathbf{y})} \Big] \quad \textnormal{using Eq.  (\ref{sup:forward_diffusion_process}) and~(\ref{sup:reverse_diffusion_process})} \nonumber \\
&= \mathbb{E}_q \Big[ -\log p_{\theta}(\mathbf{x}_{T}| \mathbf{y}) + \displaystyle \sum_{t=1}^{T} \log \frac{q(\mathbf{x}_{t}| \mathbf{x}_{t-1})}{p_{\theta}(\mathbf{x}_{t-1}| \mathbf{x}_{t}, \mathbf{y})} \Big] \nonumber \\
&= \mathbb{E}_q \Big[ -\log p_{\theta}(\mathbf{x}_{T}| \mathbf{y}) + \displaystyle \sum_{t=2}^{T} \log \frac{q(\mathbf{x}_{t}| \mathbf{x}_{t-1})}{p_{\theta}(\mathbf{x}_{t-1}| \mathbf{x}_{t}, \mathbf{y})} + \log \frac{q(\mathbf{x}_{1}| \mathbf{x}_{0})}{p_{\theta}(\mathbf{x}_{0}| \mathbf{x}_{1}, \mathbf{y})} \Big] \nonumber \\
&= \mathbb{E}_q \Big[ -\log p_{\theta}(\mathbf{x}_{T}| \mathbf{y}) + \displaystyle \sum_{t=2}^{T} \log \Big(\frac{q(\mathbf{x}_{t-1}| \mathbf{x}_{t},\mathbf{x}_{0})}{p_{\theta}(\mathbf{x}_{t-1}| \mathbf{x}_{t},  \mathbf{y})}\cdot \frac{q(\mathbf{x}_{t}| \mathbf{x}_{0})}{q(\mathbf{x}_{t-1}| \mathbf{x}_{0})} \Big)+ \log \frac{q(\mathbf{x}_{1}| \mathbf{x}_{0})}{p_{\theta}(\mathbf{x}_{0}| \mathbf{x}_{1}, \mathbf{y})} \Big] \nonumber \\
&= \mathbb{E}_q \Big[ -\log p_{\theta}(\mathbf{x}_{T}| \mathbf{y}) + \displaystyle \sum_{t=2}^{T} \log \frac{q(\mathbf{x}_{t-1}| \mathbf{x}_{t},\mathbf{x}_{0})}{p_{\theta}(\mathbf{x}_{t-1}| \mathbf{x}_{t},  \mathbf{y})} + \sum_{t=2}^{T} \frac{q(\mathbf{x}_{t}| \mathbf{x}_{0})}{q(\mathbf{x}_{t-1}| \mathbf{x}_{0})} + \log \frac{q(\mathbf{x}_{1}| \mathbf{x}_{0})}{p_{\theta}(\mathbf{x}_{0}| \mathbf{x}_{1}, \mathbf{y})} \Big] \nonumber \\
&= \mathbb{E}_q \Big[ -\log p_{\theta}(\mathbf{x}_{T}| \mathbf{y}) + \displaystyle \sum_{t=2}^{T} \log \frac{q(\mathbf{x}_{t-1}| \mathbf{x}_{t},\mathbf{x}_{0})}{p_{\theta}(\mathbf{x}_{t-1}| \mathbf{x}_{t},  \mathbf{y})} + \frac{q(\mathbf{x}_{T}| \mathbf{x}_{0})}{q(\mathbf{x}_{1}| \mathbf{x}_{0})} + \log \frac{q(\mathbf{x}_{1}| \mathbf{x}_{0})}{p_{\theta}(\mathbf{x}_{0}| \mathbf{x}_{1}, \mathbf{y})} \Big] \nonumber \\
&= \mathbb{E}_q \Big[ \log \displaystyle \frac{q(\mathbf{x}_{T}| \mathbf{x}_{0})}{p_{\theta}(\mathbf{x}_{T}| \mathbf{y})} + \displaystyle \sum_{t=2}^{T} \log \frac{q(\mathbf{x}_{t-1}| \mathbf{x}_{t},\mathbf{x}_{0})}{p_{\theta}(\mathbf{x}_{t-1}| \mathbf{x}_{t},  \mathbf{y})} - \log p_{\theta}(\mathbf{x}_{0}| \mathbf{x}_{1}, \mathbf{y}) \Big] \nonumber \\
&= \mathbb{E}_q \Bigg[KL\big[q(\mathbf{x}_{T}| \mathbf{x}_{0}) ||  p_{\theta}(\mathbf{x}_{T}| \mathbf{y}) \big] + \displaystyle \sum_{t=2}^{T} KL\big[ q(\mathbf{x}_{t-1}| \mathbf{x}_{t},\mathbf{x}_{0}) ||  p_{\theta}(\mathbf{x}_{t-1}| \mathbf{x}_{t},  \mathbf{y}) \big] - \log p_{\theta}(\mathbf{x}_{0}| \mathbf{x}_{1}, \mathbf{y}) \Bigg] \nonumber \\
&= \sum_{t=0}^{T} L_t \quad \textnormal{with} \quad \left\{ \begin{array}
{ll}
L_{0} &= - \mathbb{E}_q \Big[ \log p_{\theta}(\mathbf{x}_{0}| \mathbf{x}_{1}, \mathbf{y}) \Big] \\
L_{t} &= \mathbb{E}_q \Big[ KL\big[ q(\mathbf{x}_{t-1}| \mathbf{x}_{t},\mathbf{x}_{0}) ||  p_{\theta}(\mathbf{x}_{t-1}| \mathbf{x}_{t},  \mathbf{y}) \big] \Big] \\
L_{T} & = \mathbb{E}_q \Big[ KL\big[q(\mathbf{x}_{T}| \mathbf{x}_{0}) ||  p_{\theta}(\mathbf{x}_{T}| \mathbf{y}) \big] \Big]
\end{array}
\right. \label{eq:sup_vlb_to_kl}
 \end{align}

To keep notation concise, we consider $ \mathbb{E}_{\mathbf{x}_{0:T}\sim q(\mathbf{x}_{0:T})}=\mathbb{E}_q$ in the previous serie of equation. In \cref{eq:sup_vlb_to_kl}, $L_{T}$ could be ignored because it doesn't depend on $\theta$. $L_{0}$ is modeled by \citealp{ho2020denoising} using a separate neural network. $L_{t}$ is a KL between $2$ Gaussians distribution, so it could be calculated with a closed form. 

We observe that the probability distribution $q(\mathbf{x}_{t-1}| \mathbf{x}_{t},\mathbf{x}_{0})$ is actually tractable~\cite{ho2020denoising}: 
\begin{align}
 q(\mathbf{x}_{t-1}| \mathbf{x}_{t},\mathbf{x}_{0}) = \mathcal{N}(\mathbf{x}_{t-1};\tilde{\mathbf{\mu}}_t (\mathbf{x}_{t} ,\mathbf{x}_{0}), \tilde{\beta}_{t}\textbf{I}) \quad \textnormal{with}  \quad \left\{ \begin{array} {ll}
 \tilde{\mathbf{\mu}}_t (\mathbf{x}_{t} ,\mathbf{x}_{0}) &= \displaystyle \frac{\sqrt{\bar{\alpha}_{t-1}}\beta_{t}}{1 - \bar{\alpha}_{t}}\mathbf{x}_{0} + \frac{\sqrt{\bar{\alpha}_{t}}(1 - \bar{\alpha}_{t-1})}{1 - \bar{\alpha}_{t}}\mathbf{x}_{t}\\
 \tilde{\beta}_{t} &=  \displaystyle \frac{1 - \bar{\alpha}_{t-1}}{1 - \bar{\alpha}_{t}}\beta_{t}
 \end{array}
\right. \label{eq:sup_tractable_q} 
\end{align}

With $\tilde{\mathbf{\mu}}_t (\mathbf{x}_{t} ,\mathbf{x}_{0})$ and $\tilde{\beta}_{t}\textbf{I}$ the mean and the variance of $q(\mathbf{x}_{t-1}| \mathbf{x}_{t},\mathbf{x}_{0})$, respectively. Using \cref{sup:reparametrization} we can express $\mathbf{x}_0$ in a convenient way:
\begin{align}
\label{eq:sup_x0}
\mathbf{x}_0 = \frac{1}{\sqrt{\bar{\alpha}}} (\mathbf{x}_t - \sqrt{1 - \bar{\alpha}_t}\mathbf{\epsilon})
\end{align}

We can then simplify $\tilde{\mathbf{\mu}}_t (\mathbf{x}_{t} ,\mathbf{x}_{0})$ in \cref{eq:sup_tractable_q}:
\begin{align}
\tilde{\mathbf{\mu}}_t (\mathbf{x}_{t} ,\mathbf{x}_{0}) = \tilde{\mathbf{\mu}}_t = \frac{1}{\sqrt{\alpha_t}}\Big(\mathbf{x}_t - \frac{1- \alpha_t}{\sqrt{1 - \bar{\alpha}_t}}\mathbf{\epsilon}\Big)
\label{eq:sup_mu_t}
\end{align}

Similarly, we can re-parameterize $p_{\theta}(\mathbf{x}_{t-1}|\mathbf{x}_{t}, \mathbf{y})$ because $\mathbf{x}_{t}$ is available as input at training time:
\begin{align}
\mathbf{\mu}_{\theta} (\mathbf{x}_{t} , t) = \frac{1}{\sqrt{\alpha_t}}\Big( \mathbf{x}_t - \frac{1 - \alpha_t}{\sqrt{1-\bar{\alpha}_t}}\epsilon_{\theta}(\mathbf{x}_t, t) \Big)
\label{eq:sup_mu_theta}
\end{align}

We compute $L_{t}$ (from \cref{eq:sup_vlb_to_kl}) by using the closed-form formula of the KL between $2$ Gaussian distributions:  
\begin{align}
L_t &= \mathbb{E}_q \Bigg[ \frac{1}{2\left\| \sigma^{2}_{t} \right\|_{2}^{2}} \left\| \tilde{\mathbf{\mu}}_t (\mathbf{x}_{t} ,\mathbf{x}_{0}) - \mathbf{\mu}_{\theta} (\mathbf{x}_{t} , t) \right\|_{2}^{2} \Bigg] \nonumber \\
& =   \mathbb{E}_q \Bigg[ \frac{1}{2\left\| \sigma^{2}_{t} \right\|_{2}^{2}} \left\|  \frac{1}{\sqrt{\alpha_t}}\Big(\mathbf{x}_t - \frac{1- \alpha_t}{\sqrt{1 - \bar{\alpha}_t}}\mathbf{\epsilon}\Big) -  \frac{1}{\sqrt{\alpha_t}}\Big( \mathbf{x}_t - \frac{1 - \alpha_t}{\sqrt{1-\bar{\alpha}_t}}\epsilon_{\theta}(\mathbf{x}_t, t) \Big) \right\|_{2}^{2} \Bigg]\quad \text{using Eqs. \ref{eq:sup_mu_t} and \ref{eq:sup_mu_theta} } \nonumber \\
&= \mathbb{E}_q \Bigg[ \frac{(1 - \alpha_{t})^2}{2\alpha_{t}(1 - \bar{\alpha}_t)\left\| \sigma^{2}_{t} \right\|_{2}^{2}} \left\| \mathbf{\epsilon} - \mathbf{\epsilon}_{\theta}(\sqrt{\bar{\alpha}_t}\mathbf{x}_{0} + \sqrt{1 - \bar{\alpha}_t}\mathbf{\epsilon}, t) \right\|_{2}^{2} \Bigg]
\label{eq:sup_loss_complex}
\end{align}

One could simplify the loss shown in \cref{eq:sup_loss_complex} \cite{ho2020denoising}:
\begin{align}
L_t & = \mathbb{E}_q \Big[ \left\| \mathbf{\epsilon} - \mathbf{\epsilon}_{\theta}(\sqrt{\bar{\alpha}_t}\mathbf{x}_{0} + \sqrt{1 - \bar{\alpha}_t}\mathbf{\epsilon}, t) \right\|_{2}^{2} \Bigg] \\
& = \mathbb{E}_q \Big[ \left\| \mathbf{\epsilon} - \mathbf{\epsilon}_{\theta}(\mathbf{x}_t, t) \right\|_{2}^{2} \Bigg]
\end{align}

\newpage
\section{Details on the \DDPM{} trained on Omniglot}
\label{sup:DDPM_omniglot}

\subsection{Architecture}
The \DDPM{} and \CFGDM{} models are leveraging a U-Net~\citep{ronneberger2015u} to model $\mathbf{\epsilon}_\theta$. The U-Net is made of an encoder and a decoder. The architecture is described in detail in \cref{SI:table_Conv_Next}.
\begin{table}[h!]
  \caption{Description of U-Net architecture of the \DDPM{} and \CFGDM{}}
  \centering
  \begin{tabular}{ccc}
    \toprule
    Network & Layer & \# params \\
    \midrule
    \multirow{8}{*}{ConvNext(In$_{c}$, Out$_{c}$)} & Conv2d(In$_{c}$, In$_{c}$, 7, padding=3)    &    In$_{c}$ $\times$ In$_{c}$ $\times$ 7 $\times$ 7 +  Inc$_{c}$ \\
    & GroupNorm(In$_{c}$)  & 2 x In$_{c}$ \\
    & Conv2d(In$_{c}$, 3*In$_{c}$, 3, padding=3)  & 3*In$_{c}$ $\times$ In$_{c}$ $\times$ 3 $\times$ 3 +  3*Inc$_{c}$ \\
    & GeLU & - \\
    & GroupNorm(In$_{c}$)  & 6 x Inc$_{c}$ \\
    & Conv2d(3*In$_{c}$, Out$_{c}$, 3, padding=3)  & 3*In$_{c}$ $\times$ Out$_{c}$ $\times$ 3 $\times$ 3 +  Out$_{c}$ \\
    & \textcolor{gray}{ \#\# Shortcut connection} & \\
    & Conv2d(In$_{c}$, Out$_{c}$, 3, padding=3)  & In$_{c}$ $\times$ Out$_{c}$ $\times$ 3 $\times$ 3 +  Out$_{c}$ \\
    \midrule
    \multirow{2}{*}{TimeEmbedding(In$_{c}$, Out$_{c}$)} & 
    GeLU & - \\
    & Linear(In$_{c}$, Out$_{c}$)    &    Out$_{c}$ $\times$ In$_{c}$ \\
    \midrule
    \multirow{4}{*}{LinearAttention(In$_{c}$)} & Conv2d(In$_{c}$, 8*In$_{c}$, 1, padding=0)    &    In$_{c}$ $\times$ 8*In$_{c}$ +  8*In$_{c}$ \\
    & Conv2d(3*In$_{c}$, In$_{c}$, 1, padding=0)  & 3*In$_{c}$ $\times$ In$_{c}$ +  In$_{c}$ \\
    & GroupNorm(In$_{c}$)  & Inc$_{c}$ \\
    & Conv2d(In$_{c}$, In$_{c}$, 4, padding=1)  & In$_{c}$ $\times$ In$_{c}$ $\times$ 4 $\times$ 4 +  In$_{c}$ \\
    \midrule
    \multirow{4}{*}{DS\_U-Net\_Block(Inc$_{c}$, Out$_{c}$)} & 
    ConvNext(Inc$_{c}$, Out$_{c}$)&     \\
    & TimeEmbedding(192, Out$_{c}$)  &   \\
    & ConvNext(Out$_{c}$, Out$_{c}$)  &   \\
    & TimeEmbedding(192, Out$_{c}$)  &   \\
    & LinearAttention(Out$_{c}$)  &   \\
    & DownSampling(2)  &   \\
    \midrule
    \multirow{4}{*}{US\_U-Net\_Block(Inc$_{c}$, Out$_{c}$)} & ConvNext(Inc$_{c}$, Out$_{c}$)&     \\
    & TimeEmbedding(192, Out$_{c}$)  &   \\
    & ConvNext(Out$_{c}$, Out$_{c}$)  &   \\
    & TimeEmbedding(192, Out$_{c}$)  &   \\
    & LinearAttention(Out$_{c}$)  &   \\
    & UpSampling(2)  &   \\
    \midrule
    \multirow{10}{*}{U-Net Omniglot} &  Conv2d(2, 32, 7, padding=3) &    2 K \\
    & DS\_U-Net\_Block(32, 48) &    167 K \\
    & DS\_U-Net\_Block(48, 96)  &  577 K \\
    & DS\_U-Net\_Block(96, 192)  &  1 771 K \\
    & ConvNext(192, 192)  &  956 K \\
    & LinearAttention(192)  &  82 K \\
    & ConvNext(192, 192)  &  956 K \\
    & US\_U-Net\_Block(2*192, 96)  &  1 062 K \\
    & US\_U-Net\_Block(2*96, 48)  &  297 K \\
    &  ConvNext(48, 1) &    60K \\
    %\midrule
    
  \end{tabular}
\label{SI:table_Conv_Next}
\end{table}

The encoder of the U-Net is made with 4 layers: a first convolution and 3 down-sampling layers (called DS\_U-Net\_Block). These down-sampling layers are made with $2$ ConvNext layers~\citep{liu2022convnet}
 followed by one Linear Attention layer~\citep{li2020linear}. Each of the feature maps of the ConvNext layer is conditioned with time through the TimeEmbedding Block.

 The information bottleneck is composed of $3$ layers: a ConvNext layer followed by a Linear attention layer followed by another ConvNext layer.

After the bottleneck, there are $2$ up-sampling layers (called US\_U-Net\_Block). The up-sampling layers are very similar to the down-sampling ones except that they increase the size of the feature maps by a factor of $2$. Similarly to the DS\_U-Net\_Block, each ConvNext layer in the US\_U-Net\_Block is time-conditioned using the TimeEmbedding layer. In the end, we use a ConvNet layer to equate the number of channels and the size of the output image.
 
Overall, the base architectures of the \DDPM{} and the \CFGDM{} on Omniglot have $5.9$ million parameters.

\subsection{Training details}
We schedule the $\beta_t$ coefficient in \cref{sup:forward_diffusion_process}. $\beta_0$ is equal to $1.10^{-4}$ and $\beta_T$ to $0.02$. The $\beta_T$ are linearly spanning the time space between $1.10^{-4}$ and $0.02$. In the base architecture $T$ is set to $600$

For the training of the parameters of the U-Net model, we use an Adam Optimizer~\citep{kingma2014adam} with a learning rate of $1.10^{-4}$. We train the network for $300$ epochs, with a batch size of $128$

\subsection{Explored hyper-parameters}
To obtain the scatter plot in \cref{fig:fig1a}, we have varied certain hyper-parameters:
\begin{itemize}
    \item The $T$ hyper-parameter, ranging from $200$ to $900$ with steps of $100$ ($8$ values overall).
    \item The number of features of the First ConvNext Layer ($48$ in the base architecture), ranging from $36$ to $120$ with steps of $12$ ($8$ values overall). Note that this hyper-parameter has a strong impact on the total number of parameters of the U-Net network because the number of features of the subsequent ConvNext blocks depends on the number of features of the first ConvNext layer (it is multiplied by $2$ at every layer).
\end{itemize}
Overall we have plotted the diversity and the accuracy of $64$ \DDPM{} models in \cref{fig:fig1a}.

\newpage
\subsection{\DDPM{} samples on Omniglot}
\begin{figure}[h!]
\centering
\includegraphics[width=0.8\textwidth]{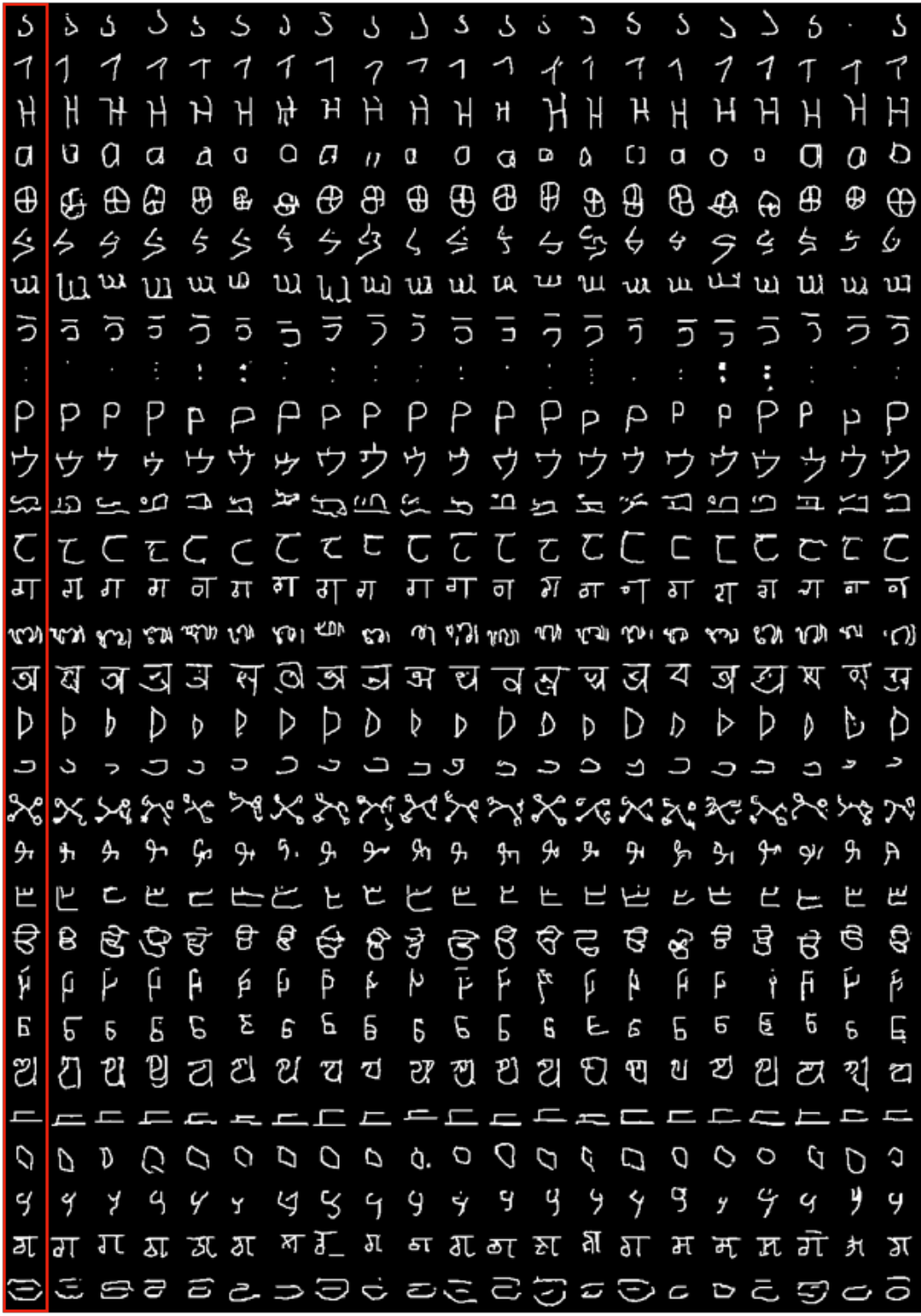}
%\vspace{-20pt}
         \caption{Samples generated by the \DDPM{} on Omniglot. All the exemplars used to condition the generative model are in the red frame. The $30$ concepts have been randomly sampled (out of $150$ concepts) from the Omniglot test set. Each line is composed of $20$ \DDPM{} samples that represent the same visual concept.}
         \label{SI:DDPM_samples_o}
\end{figure}

\newpage
\section{Details on the \CFGDM{} trained on Omniglot}
\label{sup:CFGDM_omniglot}

\subsection{Loss of the \CFGDM{}}
\label{sup:CFGDM_loss}
\citealp{dhariwal2021diffusion} proposed to improve the conditioning signal of the \DDPM{} using a classifier. To do so, the authors suggest the following form of conditional probability distribution:
\begin{align}
%p_{\theta, \gamma}(\mathbf{x}|\mathbf{y})  \propto p_{\theta}(\mathbf{x}) \cdot p_{\theta}(\mathbf{y}|\mathbf{x})^{\gamma}
p_{\theta, \gamma}(\mathbf{x}|\mathbf{y})  \propto p_{\theta}(\mathbf{x}) \cdot p_{\theta}(\mathbf{y}|\mathbf{x})^{1 + \gamma}
\label{eq:guidance1}
\end{align}
In this equation, $p_{\theta}(\mathbf{y}|\mathbf{x})$ is a classifier (trained separately). The score function of the corresponding diffusion model is then:
\begin{align}
%\nabla_{\mathbf{x}_t} \log p_{\theta, \gamma}(\mathbf{x}_t|\mathbf{y}) = \nabla_{\mathbf{x}_t} \log p_{\theta}(\mathbf{x}_t) + \gamma  \nabla_{\mathbf{x}_t} \log p_{\theta}(\mathbf{y}|\mathbf{x}_t) %p_{\theta, \gamma}(\mathbf{x}|\mathbf{y})
\nabla_{\mathbf{x}_t} \log p_{\theta, \gamma}(\mathbf{x}_t|\mathbf{y}) = \nabla_{\mathbf{x}_t} \log p_{\theta}(\mathbf{x}_t) + (1+\gamma)  \nabla_{\mathbf{x}_t} \log p_{\theta}(\mathbf{y}|\mathbf{x}_t) %p_{\theta, \gamma}(\mathbf{x}|\mathbf{y})
\label{eq:sup_log_classif}
\end{align}
The second term of the RHS of \cref{eq:sup_log_classif} requires to train a classifier ($\log p_{\theta}(\mathbf{y}|\mathbf{x}_t)$). Training such a classifier is not convenient because it has to be trained to recognize degraded samples (the $\mathbf{x}_t$ are the degraded versions of the original image). To circumvent this issue, \citealp{ho2022classifier} apply the Bayes' rule to replace $p_{\theta}(\mathbf{y}|\mathbf{x}_t)$:
\begin{align}
p_{\theta}(\mathbf{y}|\mathbf{x}_t)  = \displaystyle \frac{p_{\theta}(\mathbf{x}_t | \mathbf{y}) p_{\theta}(\mathbf{y}) }{p(\mathbf{x}_t)} 
\end{align}
\cref{eq:sup_log_classif} now becomes:
\begin{align} \label{eq:sup_guidance2}
%\nabla_{\mathbf{x}_t} \log p_{\theta, \gamma}(\mathbf{x}_{t}|\mathbf{y})  = (1- \gamma) \nabla_{\mathbf{x}_t} \log p_{\theta}(\mathbf{x}_t) + \gamma \nabla_{\mathbf{x}_t} \log p_{\theta}(\mathbf{x}_t|\mathbf{y})
\nabla_{\mathbf{x}_t} \log p_{\theta, \gamma}(\mathbf{x}_{t}|\mathbf{y})  =  (1+ \gamma) \nabla_{\mathbf{x}_t} \log p_{\theta}(\mathbf{x}_t|\mathbf{y}) - \gamma \nabla_{\mathbf{x}_t} \log p_{\theta}(\mathbf{x}_t)
\end{align}

\subsection{Practical considerations for $\nabla_{\mathbf{x}_t} \log p_{\theta}(\mathbf{x}_t)$ and $\nabla_{\mathbf{x}_t} \log p_{\theta}(\mathbf{x}_t|\mathbf{y})$}
\label{sup:CFGDM_practical}
The loss in \cref{eq:sup_guidance2} is particularly convenient as one can train a single model to evaluate both $\nabla_{\mathbf{x}_t} \log p_{\theta}(\mathbf{x}_t)$ and $\nabla_{\mathbf{x}_t} \log p_{\theta}(\mathbf{x}_t|\mathbf{y})$. In the section \cref{sup:ddpm_loss}, we have shown that $\nabla_{\mathbf{x}_t} \log p_{\theta}(\mathbf{x}_t|\mathbf{y})$ could be modeled with an auto-encoder $\mathbf{\epsilon}_{\theta}$ ($\epsilon_{\theta} : \mathbb{R}^{D} \times \mathbb{R}^{D} \to {R}^{D}$). We can actually use the same auto-encoder, with a non-informative conditioning signal, to model also $\nabla_{\mathbf{x}_t} \log p_{\theta}(\mathbf{x}_t)$. In practice, if we want to model $\nabla_{\mathbf{x}_t} \log p_{\theta}(\mathbf{x}_t|\mathbf{y})$, we fed the auto-encoder  $\mathbf{\epsilon}_{\theta}$ with a concatenation of the noisy input image ($\mathbf{x}_t$) and the corresponding exemplars ($\mathbf{y}$). In this case, we use the notation $\mathbf{\epsilon}_{\theta}(\mathbf{x}_t,\mathbf{y})$. To model $\nabla_{\mathbf{x}_t} \log p_{\theta}(\mathbf{x}_t)$, we fed the network with the noisy image ($\mathbf{x}_t$), concatenated with a black image. In this case, we use the notation $\mathbf{\epsilon}_{\theta}(\mathbf{x}_t,\mathbf{\varnothing})$. In practice, we use a drop-out function, to randomly drop some of the informative exemplars and replace them with a black image (following a Bernoulli distribution). We set the drop-out probability to $0.1$.

\subsection{Architecture and training}

The architecture and the training details of the \CFGDM{} model on the Omniglot dataset are exactly the same as those of the \DDPM{} on the Omniglot dataset (see \cref{sup:DDPM_omniglot}).

\subsection{Explored hyper-parameters}
To obtain the scatter plot in \cref{fig:fig1a}, we have varied certain hyper-parameters:
\begin{itemize}
    \item The $T$ hyper-parameter, ranging from $200$ to $900$ with steps of $100$ ($8$ values overall).
    \item The number of features of the First ConvNext Layer ($48$ in the base architecture), ranging from $12$ to $96$ with steps of $12$ ($8$ values overall). Note that this hyper-parameter has a strong impact on the total number of parameters of the U-Net network because the number of features of the subsequent ConvNext blocks depends on the number of features of the first ConvNext layer (it is multiplied by $2$ at every layer).
    \item The guidance scale with the following values : ($0.5$, $1$, $2$, $3$, $5$). Note, that we do not have to retrain the model to change the guidance scale, as the change is occurring only during sampling.
\end{itemize}
Overall we have plotted the diversity and the accuracy of $320$ \CFGDM{} models in \cref{fig:fig1a}.

\newpage
\subsection{\CFGDM{} samples on Omniglot}
\begin{figure}[h!]
\centering
\includegraphics[width=0.8\textwidth]{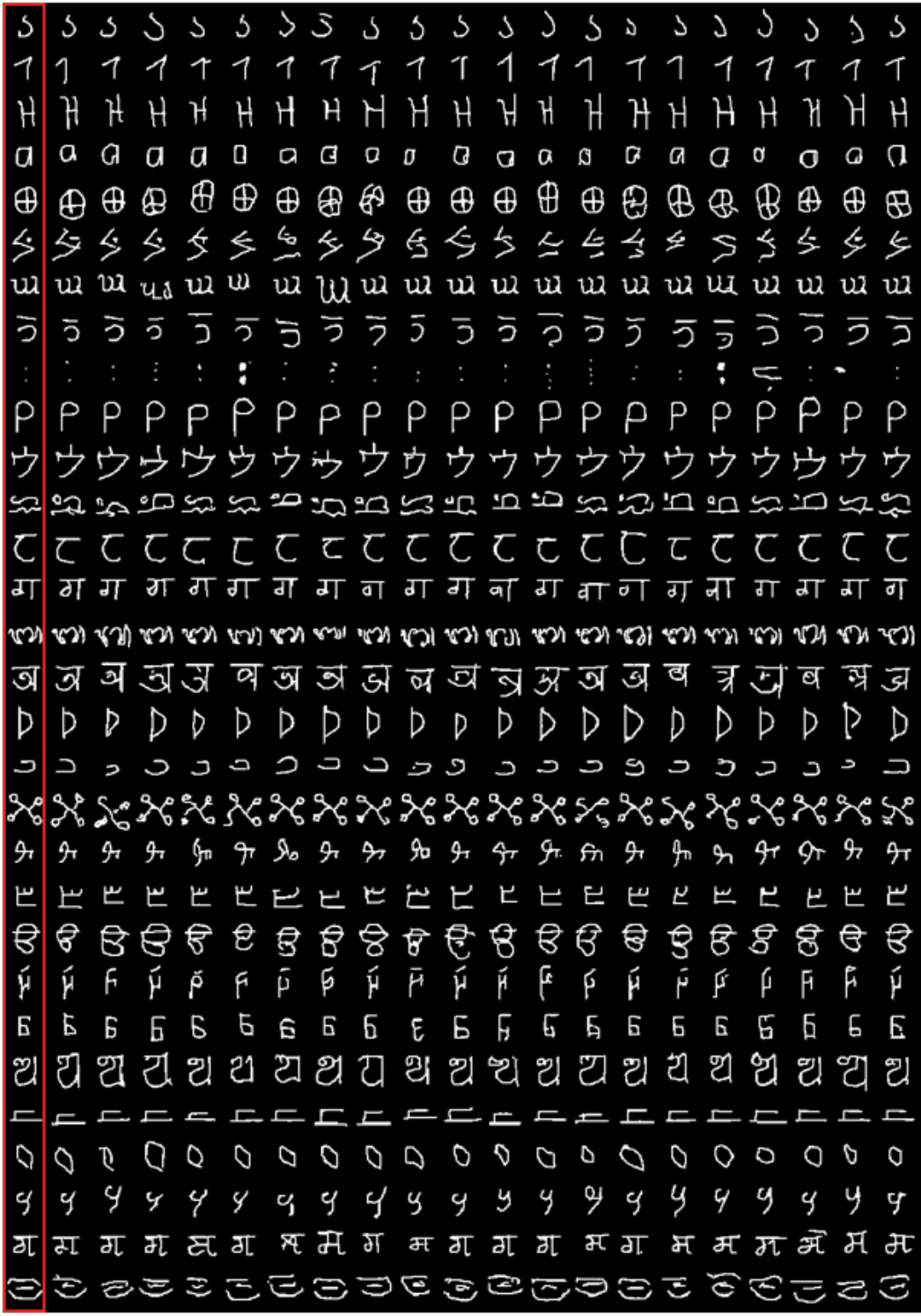}
%\vspace{-20pt}
         \caption{Samples generated by the \CFGDM{} on Omniglot. All the exemplars used to condition the generative model are in the red frame. The $30$ concepts have been randomly sampled (out of $150$ concepts) from the Omniglot test set. Each line is composed of $20$ \CFGDM{} samples that represent the same visual concept.}
         \label{SI:CFGDM_samples_o}
\end{figure}

%The context net utilized for QuickDraw-FS is a sViT similar to \citealp{seung2021svit} where we handle small sets (1-10) of images. The architecture is described in \cref{SI:table_sViT}.

\newpage
\section{Details on the \FSDM{} trained on Omniglot}
\label{sup:FSDM_omniglot}

\subsection{Conditioning}
The \FSDM{} offers an alternative to condition the \DDPM{} models. Instead of conditioning the U-Net network of the \DDPM{} with a single image, the \FSDM{} proposed to condition it with a context vector that aggregates the information from a context set. When the \FSDM{} is trained on Omniglot we condition it using a mechanism similar to FiLM (\citealp{ethan2018film}). We use a U-Net to extract a context vector from the set of samples presented to the model. The context is in the form of a vector, which is used to condition the intermediate feature maps $\mathbf{u}_t$ in the \DDPM{} U-Net. Note that the feature map $\mathbf{u}_t$ is obtained when the U-Net input is $\mathbf{x}_t$.
We can represent the conditioning as: 
\begin{align}
\mathbf{p}_t = m(\mathbf{c})\mathbf{u}_t + b(\mathbf{c})
\label{eq:conditioningFiLM}
\end{align}
Here, $m$ and $b$ are learnable and context-dependent neural networks. Additionally, we merge together $\mathbf{c}$ with the time-step embedding, and using that we  define a generic per-step conditioning mechanism for each layer.

\subsection{Architecture}

As the backbone, \FSDM{} utilizes the same architecture as the \DDPM{} trained on Omniglot. 

For the context net, as mentioned above we utilize a U-Net. The architecture of the Encoder U-Net is described in detail in Table~\ref{SI:table_param_FSDM_omniglot}. For the described architecture we consider the number of residual blocks to be 2.

The size of the model is 6.6 million parameters out of which 2.5 million parameters are for the encoder and 4.1 million are for the generative model.

\begin{table}[h!]
  \caption{Description of the Encoder U-Net architecture}
  \centering
  \begin{tabular}{ccc}
    \toprule
    Network & Layer & \# params \\
    \midrule
    \multirow{1}{*}{DownSample()} & AvgPool2D(kernel\_size=$2$, stride=$2$) & 0\\ 
    \midrule
    \multirow{3}{*}{AttnPool2D(Sp$_d$,Emb$_d$,NumHeads$_c$,Out$_d$)} & Conv1D(Emb$_d$.3*Emb$_d$,$1$,$1$) & $49.5$ K\\
    & QKVAttention(Emb$_d$ // NumHeads$_c$) & 0\\
    & Conv1D(Emb$_d$,Out$_d$,$1$,$1$) & $16.5$ K\\
    \midrule
    \multirow{12}{*}{ResBlock(In$_c$, Out$_c$, Emb$_c$, down=$False$)} & GroupNorm($32$,In$_c$) \\
    & \textcolor{blue}{if (down) : DownSample()} & \textcolor{blue}{0}\\
    & \textcolor{blue}{if (down) : DownSample()} & \textcolor{blue}{0}\\
    & SiLU & 0\\
    & Conv2D(In$_c$,Out$_c$,$3$,$1$,$1$) \\
    & SiLU & 0\\
    & Linear(Emb$_c$, $2 \times$ Out$_c$)\\
    & GroupNorm($32$,Out$_c$)  \\
    & SiLU & 0\\
    & Dropout & 0\\
    & Conv2D(Out$_c$,Out$_c$,$3$,$1$,$1$)\\   
    & \textcolor{blue}{if (In$_c$ != Out$_c$) : Conv2D(In$_c$,Out$_c$,$1$,$1$})\\
    \midrule
    \multirow{4}{*}{AttnBlock(In$_c$, NumHeads, NumHeads$_c$)} & GroupNorm($32$, In$_c$) &  \\
    & Conv1D(In$_c$,$3 \times$ In$_c$,$1$,$1$) & \\
    & QKVAttentionLegacy(In$_c$ // NumHeads$_c$) & 0\\
    & Conv1D(In$_c$,In$_c$,$1$,$1$) & \\
    \midrule
    \multirow{14}{*}{InputBlock()} & Conv2D($1$,In$_c$,$3$,$1$,$1$) & $320$\\
    & ResBlock(In$_c$=$32$,Out$_c$=$32$, Emb$_c$=$128$) & $26.9$ K\\
    & ResBlock(In$_c$=$32$,Out$_c$=$32$, Emb$_c$=$128$) & $26.9$ K\\
    & ResBlock(In$_c$=$32$,Out$_c$=$32$, Emb$_c$=$128$, down=$True$) & $26.9$ K\\
    & ResBlock(In$_c$=$32$,Out$_c$=$64$, Emb$_c$=$128$) & $74.2$ K\\
    & ResBlock(In$_c$=$64$,Out$_c$=$64$, Emb$_c$=$128$) & $90.6$ K\\
    & ResBlock(In$_c$=$64$,Out$_c$=$64$, Emb$_c$=$128$, down=$True$) & $90.6$ K\\
    & ResBlock(In$_c$=$64$,Out$_c$=$96$, Emb$_c$=$128$) & $169.8$ K\\
    & ResBlock(In$_c$=$96$,Out$_c$=$96$, Emb$_c$=$128$) & $191.2$ K\\
    & ResBlock(In$_c$=$96$,Out$_c$=$96$, Emb$_c$=$128$, down=$True$) & $191.2$ K\\
    & ResBlock(In$_c$=$96$,Out$_c$=$128$, Emb$_c$=$128$) & $304.2$ K\\
    & AttnBlock(In$_c$=$128$, NumHeads=$1$, NumHeads$_c$=$64$) & $66.3$ K\\
    & ResBlock(In$_c$=$128$,Out$_c$=$128$, Emb$_c$=$128$) & $328.7$ K\\
    & AttnBlock(In$_c$=$128$, NumHeads=$1$, NumHeads$_c$=$64$) & $66.3$ K\\
    \midrule
    \multirow{3}{*}{MiddleBlock()} & ResBlock(In$_c$=$128$,Out$_c$=In$_c$, Emb$_c$=$128$) & $328.7$ K\\
    & AttnBlock(In$_c$=$128$, NumHeads=$1$, NumHeads$_c$=$64$) & $66.3$ K\\
    & ResBlock(In$_c$=$128$,Out$_c$=In$_c$, Emb$_c$=$128$) & $328.7$ K\\
    \midrule
    \multirow{8}{*}{Encoder U-Net} & Linear($32$, $128$) & $4.2$ K \\
    & SiLU & 0\\
    & Linear($128$, $128$) & $16.5$ K \\
    & InputBlock() & $1653.8$ K\\
    & MiddleBLock() & $723.7$ K\\ 
    & GroupNorm($32$,$128$) & $256$ \\
    & SiLU & 0\\
    & AttnPool2D(Sp$_d$=6,Emb$_d$=128,NumHeads$_c$=64,Out$_d$=128) & $66.0$ K \\
    \bottomrule
    \\
    \multicolumn{3}{l}{Note : \textcolor{blue}{Blue} layers represent variable layers dependent on a certain parameter}
    
  \end{tabular}
\label{SI:table_param_FSDM_omniglot}
\end{table}

\subsection{Training details}
The attention resolution, learning rate, optimizer, and batch size we use are the same as that implemented in  \url{https://github.com/georgosgeorgos/few-shot-diffusion-models}. The size of the context channels and hidden dimensions is set to 128. 

We train the model for 300 epochs.
%\textcolor{red}{@Rishav: Here you should describe the training details of the network (learning rate, batch size...). If all these hyperparameters are similar to those in the original repo, don't need to repeat, just specify it. Also you should refer to the original repo we have used, using \url{https://XXX.xxx}.}

\subsection{Explored hyper-parameters} 
To obtain the scatter plot in \cref{fig:fig1a}, we varied the following hyper-parameters:
\begin{itemize}
\item The sample size, ranging from 2 to 6 with steps of 1 (5 values overall) 
\item The number of time-steps or diffusion steps ranging from 200 to 100 with steps of 200 (5 values overall)
\item The number of residual blocks per downsample, ranging from 1 to 4 (4 values overall)
\end{itemize}

We have trained $100$ different \FSDM{} models and plotted the diversity and the accuracy in \cref{fig:fig1a}.
%\textcolor{red}{@Rishav: Describe here the hyper-parameters we have varied to obtain the \FSDM points in Fig1a. Please present it in a similar way that what is done section G.4. }

\clearpage
\newpage
\subsection{\FSDM{} samples on Omniglot}
\begin{figure}[h!]
\centering
\includegraphics[width=0.8\textwidth]{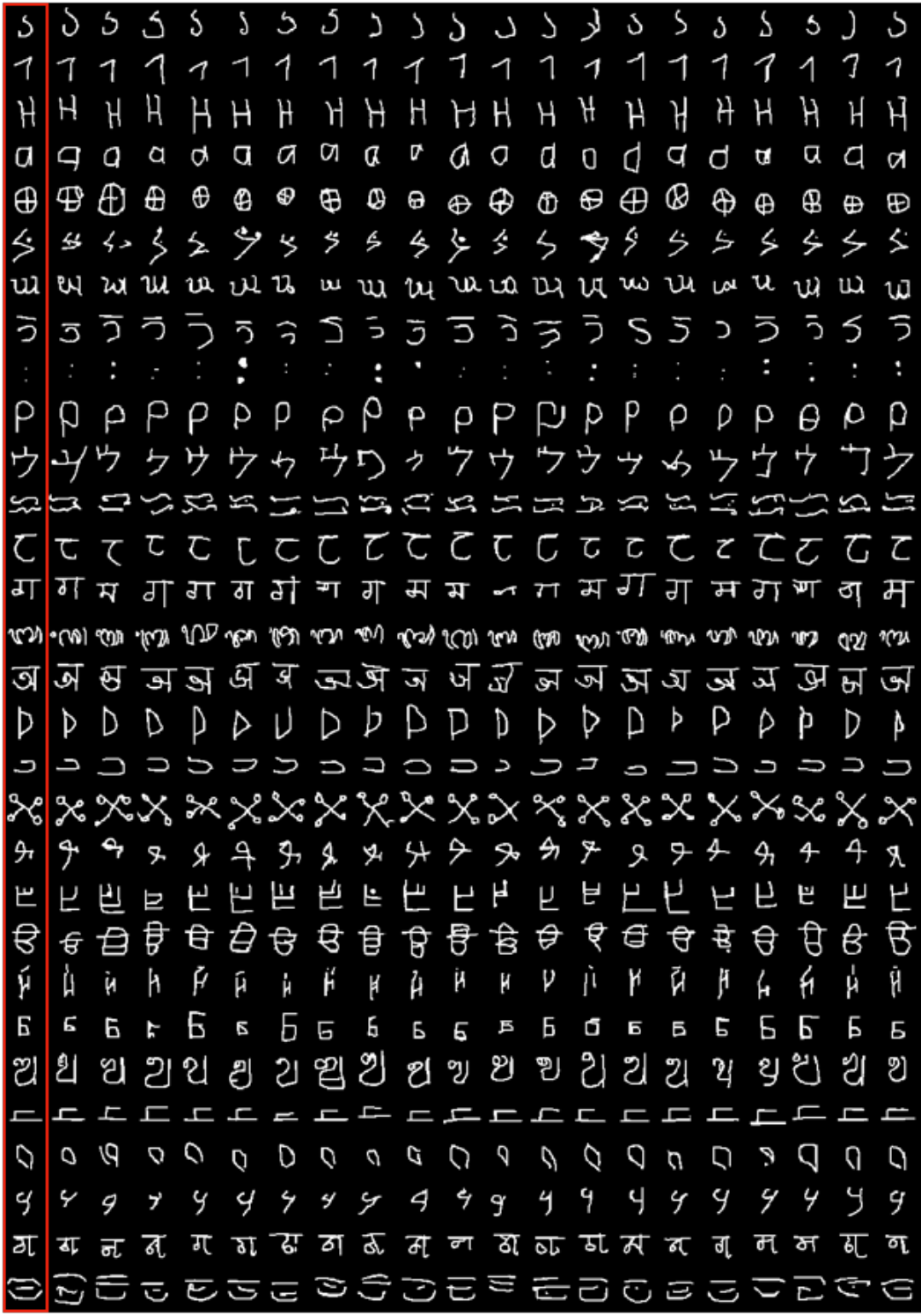}
%\vspace{-20pt}
         \caption{Samples generated by the \FSDM{} on Omniglot. All the exemplars used to condition the generative model are in the red frame. The $30$ concepts have been randomly sampled (out of $150$ concepts) from the Omniglot test set. Each line is composed of $20$ \FSDM{} samples that represent the same visual concept.}
         \label{SI:FSDM_samples_o}
\end{figure}

\newpage
\section{Details on the \DDPM{} trained on QuickDraw-FS}
\label{sup:DDPM_qd}

\subsection{Architecture and training}
The \DDPM{} trained on QuickDraw-FS has a similar architecture to the \DDPM{} trained on Omniglot (see \cref{sup:DDPM_omniglot}). The only difference is that the base architecture we use on QuickDraw-FS has 60 channels in the first ConvNext block. The total number of parameters of the base architecture on QuickDraw-FS is 13,1 million.

All the training hyper-parameters (batch size, learning rate, beta scheduling) are the same as the \DDPM{} on Omniglot.

\subsection{Explored hyper-parameters}
To obtain the scatter plot in \cref{fig:fig1b}, we have varied certain hyper-parameters:
\begin{itemize}
    \item The $T$ hyper-parameter, ranging from $200$ to $900$ with steps of $100$ ($8$ values overall).
    \item The number of features of the First ConvNext Layer ($48$ in the base architecture), ranging from $24$ to $108$ with steps of $12$ ($8$ values overall). Note that this hyper-parameter has a strong impact on the total number of parameters of the U-Net network because the number of features of the subsequent ConvNext blocks depends on the number of features of the first ConvNext layer (it is multiplied by $2$ at every layer).
\end{itemize}
We have repeated all this hyper-parameters exploration for $2$ different random seeds (i.e., different weight initialization).
Overall we have plotted the diversity and the accuracy of $128$ models in \cref{fig:fig1b}.

\newpage
\subsection{\DDPM{} samples on QuickDraw-FS}
\begin{figure}[h!]
\centering
\includegraphics[width=0.8\textwidth]{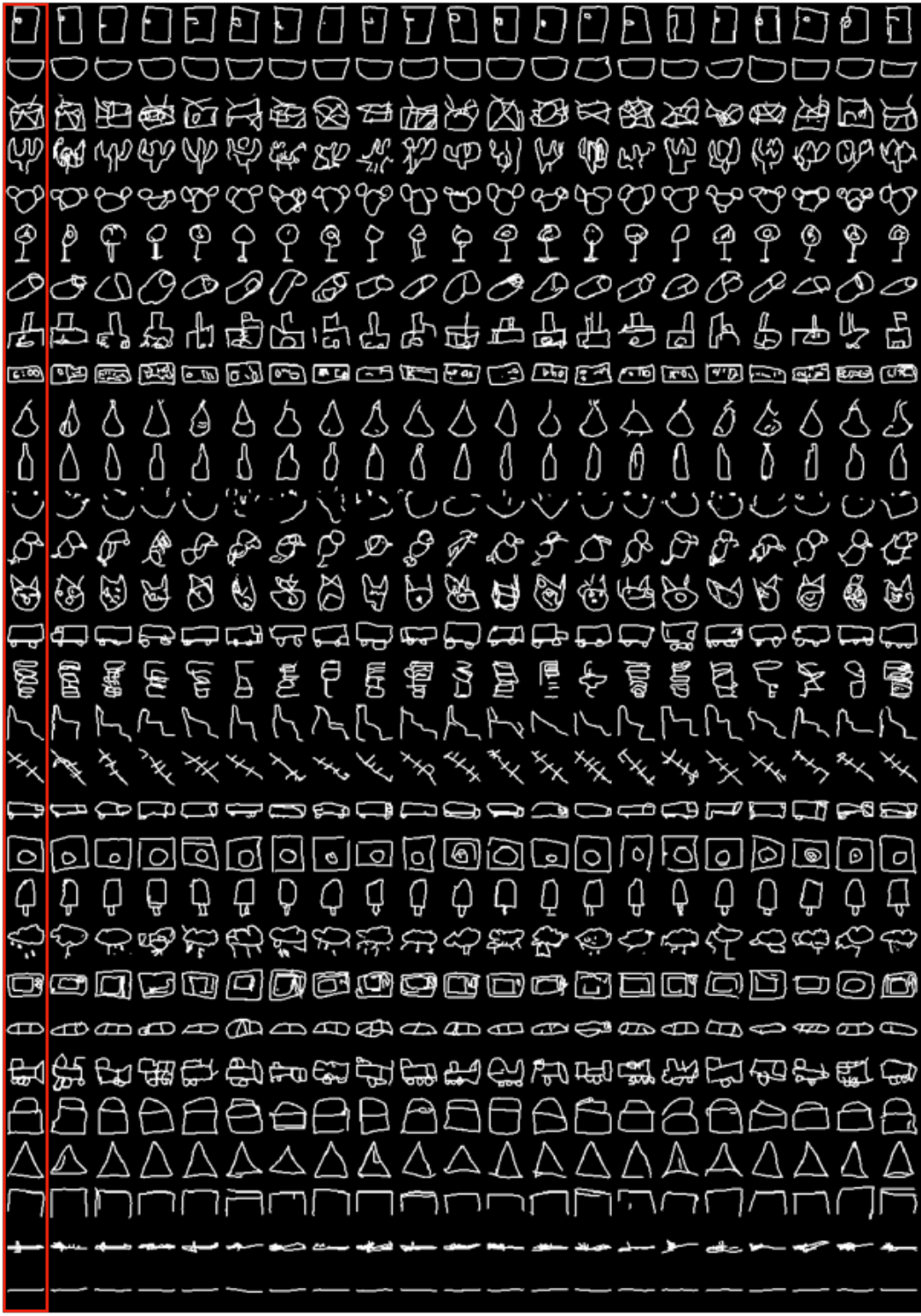}
%\vspace{-20pt}
         \caption{Samples generated by the \DDPM{} on QuickDraw-FS. All the exemplars used to condition the generative model are in the red frame. The $30$ concepts have been randomly sampled (out of $115$ concepts) from the QuickDraw-FS test set. Each line is composed of $20$ \DDPM{} samples that represent the same visual concept.}
         \label{SI:DDPM_samples_q}
\end{figure}

\newpage
\section{Details on the \CFGDM{} trained on QuickDraw-FS}
\label{sup:CFGDM_qd}
\subsection{Architecture and training}
The base architecture and the training details of the \CFGDM{} on QuickDraw-FS are exactly the same as those of the \DDPM{} on the QuickDraw-FS dataset (see \cref{sup:DDPM_qd}).

\subsection{Explored hyper-parameters}
To obtain the scatter plot in \cref{fig:fig1b}, we have varied certain hyper-parameters:
\begin{itemize}
    \item The $T$ hyper-parameters, ranging from $200$ to $900$ with steps of $100$ ($8$ values overall).
    \item The number of features of the First ConvNext Layer ($60$ in the base architecture), ranging from $24$ to $108$ with steps of $12$ ($8$ values overall). Note that this hyper-parameter has a strong impact on the total number of parameters of the U-Net network, as the number of features of the subsequent ConvNext block is equal to the previous number of features multiplied by 2...
    \item The guidance scale with the following values : ($0.5$, $1$, $2$, $3$, $5$). Note, that we do not have to retrain the model to change the guidance scale, as the change is occurring only during sampling.
\end{itemize}
We have repeated all this hyper-parameters exploration for $2$ different random seeds (i.e., different weight initialization).
Overall we have plotted the diversity and the accuracy of $320$ models in \cref{fig:fig1b}.

\newpage
\subsection{\CFGDM{} samples on QuickDraw-FS}
\begin{figure}[h!]
\centering
\includegraphics[width=0.8\textwidth]{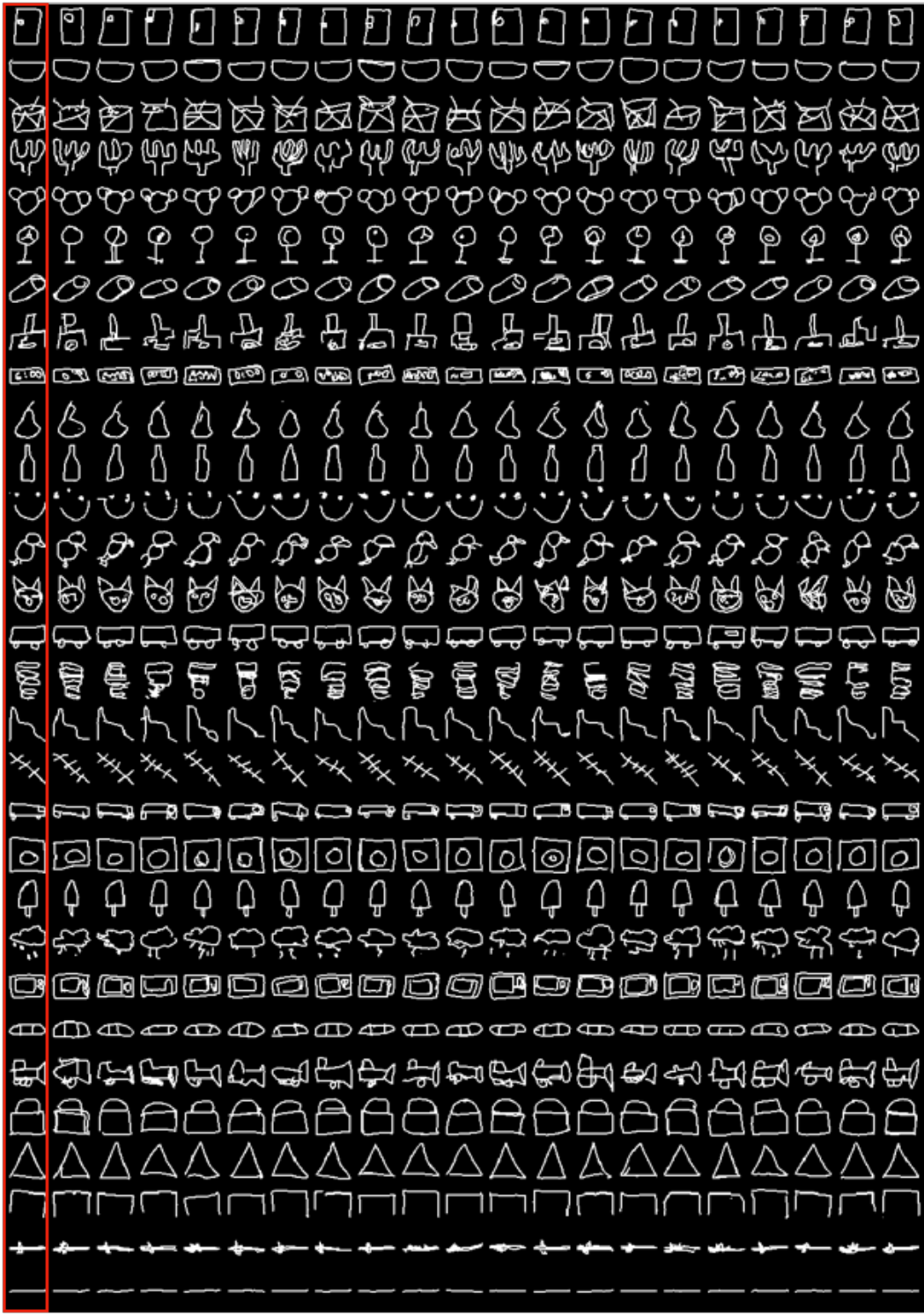}
%\vspace{-20pt}
         \caption{Samples generated by the \CFGDM{} on QuickDraw-FS. All the exemplars used to condition the generative model are in the red frame. The $30$ concepts have been randomly sampled (out of $115$ concepts) from the QuickDraw-FS test set. Each line is composed of $20$ \CFGDM{} samples that represent the same visual concept.}
         \label{SI:CFGDM_samples_q}
\end{figure}

\newpage
\section{Details on the \VAENS{} trained on QuickDraw-FS}
\label{sup:VAENS_qd}
\subsection{Architecture}
For the \VAENS{} on QuickDraw-FS we use a similar architecture to the \VAENS{} on Omniglot (described in S9 of \citealp{boutin2022diversity}). Here, we remind the reader of the main properties of the \VAENS{} network.

The \VAENS{} network is composed of different sub-networks:
\begin{itemize}

\item Shared encoder $x \mapsto h$: An instance encoder $E$ that takes each individual datapoint $x_i$ to a feature representation $h_i$ = $E(x_i)$.

\item Statistic network $q(c|D,\phi):h_1,...,h_k \mapsto \mu_c, {\sigma^2}_c$: A pooling layer that aggregates the matrix $(h_1,...,h_k)$ to a single pre-statistic vector $v$. ~\citealp{edwards2016towards} uses sample mean for their experiments. Which is followed by a post-pooling network that takes $v$ to a parametrization of a Gaussian.

\item Inference network $q(z|x, c, \phi) : h, c \mapsto \mu_z , {\sigma^2}_z$: Inference network gives an approximate posterior over latent variables.

\item Latent decoder network $p(z|c; \theta) : c \mapsto \mu_z , {\sigma^2}_z$

\item Observation decoder network $p(x|c, z; \theta) : c, z \mapsto \mu_x$

\end{itemize}

Compared to the Omniglot version, we have increased the number of parameters to fit the higher complexity of the QuickDraw-FS dataset. In particular, we have increased the number of stochastic layers in the inference network and the observation decoder network from $1$ to $6$. The number of stochastic layers controls the number of hierarchical latent variables we use in the encoder and the decoder. To keep the number of parameters reasonable we have decreased the dimension of the context vector, i.e., the output size of the static network, from $512$ to $128$. In practice, we have found that reducing this dimension has little impact on the performance while reducing drastically the number of parameters. We have set the dimension of the latent variable to $256$. For the base architecture, the size of the last latent variable is set to $80$, the number of context samples is equal to $5$, and the number of layers (per stochastic layer) is set to $6$. 

The base architecture of the \VAENS{} has 12.4 million parameters.

\subsection{Training details}
We have trained the \VAENS{} for $300$ epochs, using a batch size of $32$. We use the Adam optimizer to update the weights of the network, with a learning rate of $1.10^{-3}$.

\subsection{Explored hyper-parameters}
To obtain the scatter plot in \cref{fig:fig1b}, we have varied certain hyper-parameters:
\begin{itemize}
    \item The number of dimensions of the last latent variable, from $40$ to $120$, by steps of $40$ (so 3 different values overall)
    \item the number of sub-layers that composed each stochastic layer, from $2$ to $10$ with step $4$ (3 values overall.
    \item the $\beta$ coefficient with values: ($0.1$, $0.5$, $1$, $2$). We remind the reader that the $\beta$ coefficient in the VAE is used to increase (or decrease if $\beta\leq1$) to weight of the KL in the ELBO loss~\citep{higgins2016beta}.
    \item the number of context samples with values ($2$, $5$, $10$). In the \VAENS{}, the context samples are used to evaluate the statistics of a specific category (through the statistic network). 
    
\end{itemize}
Overall we have plotted the diversity and the accuracy of $108$ models in \cref{fig:fig1b}.

\newpage
\subsection{\VAENS{} samples on QuickDraw-FS}
\begin{figure}[h!]
\centering
\includegraphics[width=0.8\textwidth]{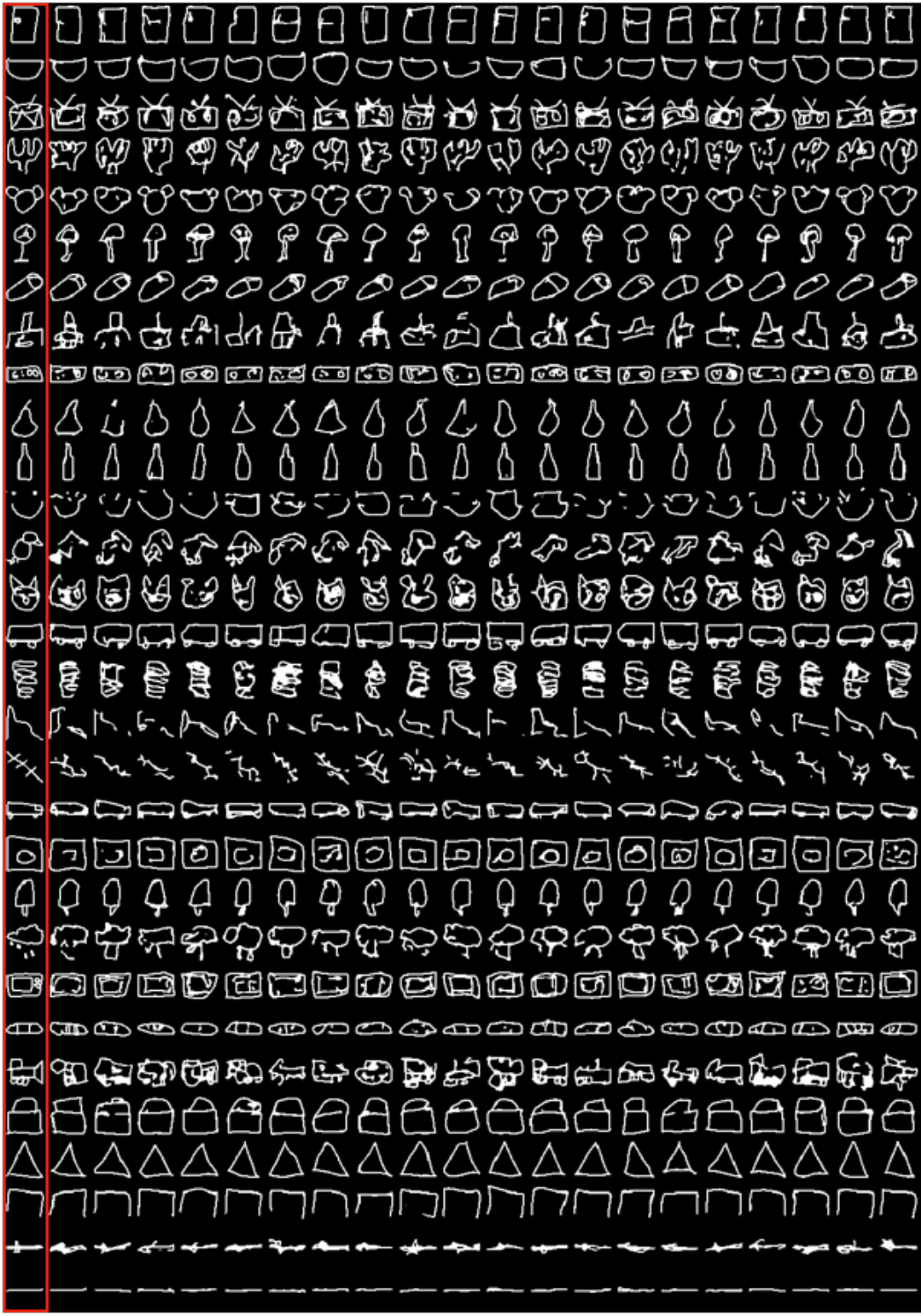}
%\vspace{-20pt}
         \caption{Samples generated by the \VAENS{} on QuickDraw-FS. All the exemplars used to condition the generative model are in the red frame. The $30$ concepts have been randomly sampled (out of $115$ concepts) from the QuickDraw-FS test set. Each line is composed of $20$ \VAENS{} samples that represent the same visual concept.}
         \label{SI:VAENS_samples_q}
\end{figure}

\newpage
\section{Details on the \VAESTN{} trained on QuickDraw-FS}
\label{sup:VAESTN_qd}

The \VAESTN{} is a sequential VAE that allows for the iterative construction of a complex image~\citep{rezende2016one}. At each iteration, the algorithm focuses its attention on a specific part of the image ($\boldsymbol{x}$), the prototype ($\boldsymbol{\tilde{x}}$), and the residual image ($\boldsymbol{\hat{x}}$) using the Reading Spatial Transformer Network (STN${_r}$). Then the extracted patch is passed to an encoding network (EncBlock) to transform it into a latent variable. This latent variable is concatenated to a patch extracted from the prototype and then passed to the RecBlock network. The produced hidden state is first passed to DecBlock to recover the original patch, and then to the STN$_{w}$ to replace and rescale the patch into the original image. The LocNet network is used to learn the parameter of the affine transformation we used in the STN. Note that the affine parameters used in STN$_{w}$ are simply the inverse of those used in STN$_{r}$.

The STN modules take $2$ variables in input: an image (or a patch in the case to the STN$_{w}$) and a matrix (3$\times$2) describing the parameters of the affine transformation to apply to the input image~\citep{jaderberg2015spatial}. All other modules are made with MLPs networks and are described in Table~\ref{SI:table_param_VAE_STN}. In the Table~\ref{SI:table_param_VAE_STN} we use the following notations:
\begin{itemize}
    \item $s_z$: This is the size of the latent space. In the base architecture, we set $s_z=80$.
    \item $s_{LSTM}$: This is the size of the output of the Long-Short Term Memory (LSTM) unit. In the base architecture, we set $s_{LSTM}=400$
    \item $s_r$: This is the resolution of the patches extracted by the Spatial Transformer Net (STN) during the reading operation. In the base architecture, we set $s_r=15$.
    \item $s_{loc}$: This is the number of neurons used at the input of the localization network. In the base architecture, we set $s_{loc}=150$
    \item $s_{w}$: This is the resolution of the patch passed to the STN network for the writing operation. In the base architecture $s_w=12$.
\end{itemize}

For the base architecture, we used $N_{steps}=80$. The base architecture of the \VAESTN{} has $11.9$ million parameters. For more details on the loss function, please refer to \cite{rezende2016one}.
\begin{table}[h!]
  \caption{Description of the VAE-STN architecture}
  \centering
  \begin{tabular}{ccc}
    \toprule
    Network & Layer & \# params \\
    \midrule
    \multirow{11}{*}{EncBlock(s$_r$, s$_{LSTM}$, s$_z$)} & Linear($3$ $\times$ s$_{r}^2$ + s$_{LSTM}$ , $2048$) &   \thead{($3$ $\times$ s$_{r}^2$ + s$_{LSTM}$) $\times$ $2048$)\\ + $2048$}\\
    & ReLU \\
    & Linear($2048$, $1024$) & $2097$ K \\
    & ReLU \\
    & Linear($1024$, $1024$) & $1050$ K \\
    & ReLU \\
    & Linear($1024$, $512$) & $524$ K \\
    & ReLU  & - \\
    & Linear($512$, $128$) & $65$ K \\
    & ReLU  & - \\
    & Linear($128$, $2\times$ s$_{z}$) & $256\times$s$_{z}$ + 2$\times$s$_{z}$ \\
    \midrule
    \multirow{5}{*}{LocNet(s$_{loc}$)} & Linear(s$_{loc}$, $64$) &   s$_{loc}\times64$ + $64$\\
    & ReLU & - \\
    & Linear($64$, $32$) & $2$ K \\
    & ReLU  & - \\
    & Linear($32$, $6$) & $0.2$ K \\
    \midrule
    \multirow{9}{*}{DecBlock(s$_{LSTM}$, s$_{loc}$, s$_{w}$)} & Linear(s$_{LSTM}$ - s$_{loc}$, $2048$) & (s$_{LSTM}$ - 
    s$_{loc}$)$\times2048$ + $2048$ \\
    & ReLU & - \\
    & Linear($2048$, $1024$) & $2097$ K \\ 
    & ReLU & - \\
    & Linear($1024$, $512$) & $525$ K \\ 
    & ReLU & - \\
    & Linear($512$, $256$) & $131$ K \\ 
    & ReLU & - \\
    & Linear($256$, $4*s_{w}^2$) & $256\times4\times$s$_{w}^2 + $4*s$_{w}^2$ \\ 
    \midrule
    \multirow{1}{*}{RecBlock(s$_z$, s$_r$, s$_{LSTM}$)} & LSTMCell(s$_z$ + s$_{r}^2$, s$_{LSTM}$) & \thead{$4\times\big($s$_z$ + s$_{r}^2$)$\times$s$_{LSTM}$ \\ + s$_{LSTM}^2$ + s$_{LSTM}$\big)} \\
    \midrule
    \multirow{4}{*}{VAE-STN} & EncBlock($15$ , $400$, $80$) & $5,4$ K \\
    & RecBlock($80$, $12$, $400$) &  $2,998$ K \\
    & DecBlock($400$, $150$, $15$) & $3,552$ K \\ 
    & LocNet($150$) & $11.9 K$ \\
    \bottomrule
  \end{tabular}
\label{SI:table_param_VAE_STN}
\end{table}

All other training details are similar to the version trained on Omniglot (see Section S8 in the supplementary information of \citealp{boutin2022diversity}).

\clearpage
\newpage
\subsection{\VAESTN{} samples on QuickDraw-FS}
\begin{figure}[h!]
\centering
\includegraphics[width=0.8\textwidth]{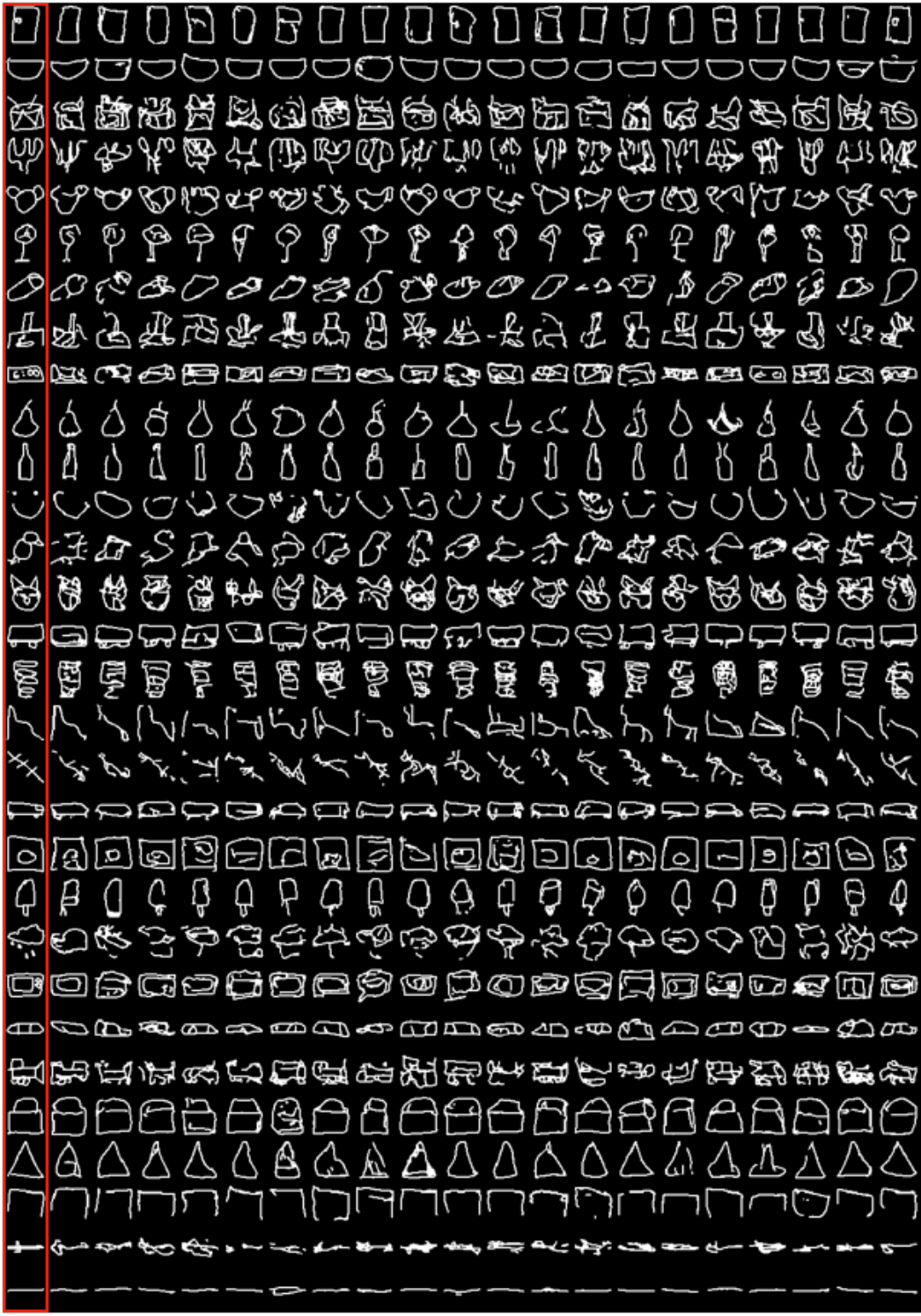}
%\vspace{-20pt}
         \caption{Samples generated by the \VAESTN{} on QuickDraw-FS. All the exemplars used to condition the generative model are in the red frame. The $30$ concepts have been randomly sampled (out of $115$ concepts) from the QuickDraw-FS test set. Each line is composed of $20$ \VAESTN{} samples that represent the same visual concept.}
         \label{SI:VAESTN_samples_q}
\end{figure}

\newpage
\section{Details on the \DAGANUN{} and \DAGANRN{} trained on QuickDraw-FS}
\label{sup:DAGAN_qd}

\subsection{Architecture}
% \textcolor{red}{@Lakshya: You should at least remind what's the difference between DAGAN-UN and DAGAN-RN. You should also remind the reader how the blocks are interconnected, and what is the function of each block. }

The DA-GAN architecture used for the Omniglot dataset is derived from \citealp{boutin2022diversity}. For QuickDraw-FS dataset, the same architecture has been extended as shown in Table \ref{SI:table_param_DAGAN-UN}. \DAGANUN{} and \DAGANRN{} model refers to the version whose generator is based on the U-Net and ResNet architecture respectively. Therefore, the difference between the two models is due to the presence of skip connections in \DAGANUN{} model. Following are the details on the DA-GAN's generator model:

\begin{itemize}
    \item $s_{z}$: It represents the size of the latent space. In the base architecture, we have used $s_z$ = 120.
    \item $G(x,z)$: This is the generator of DAGAN model, which takes an exemplar $x$ and gaussian noise $z$ as input to generate new samples. 
\end{itemize}

The base architecture of DAGAN's generator has 10.5 million parameters. All the DAGAN training details are similar to its Omniglot version (refer to Sections S10 and S11 in the supplementary section of \citealp{boutin2022diversity}.

\begin{table}[h!]
 \caption{Description of the Data Augmentation GAN Architecture}
 \centering
 \begin{tabular}{ccc}
    \toprule
    Network & Layer & \# params \\
    
    \midrule
    \multirow{2}{*}{ConvBlock(In$_{c}$, Out$_{c}$, s$_{l}$)}  & 
    Conv2d(In$_{c}$, Out$_{c}$, 3, stride=s$_{l}$, padding=1)   &    Out$_{c}$ $\times$ (In$_{c}$ $\times$ 3 $\times$ 3 + 1)    \\
    & LeakyReLU(0.2), BatchNorm2d(Out$_{c}$)   & 2 x Out$_{c}$ \\
    
\midrule

    \multirow{2}{*}{DeConvBlock(In$_{c}$, Out$_{c}$, s$_{l}$)}  & 
    ConvTranspose2d(In$_{c}$, Out$_{c}$, 3, stride=s$_{l}$, padding=1)   &    Out$_{c}$ $\times$ (In$_{c}$ $\times$ 3 $\times$ 3 + 1)    \\
    & LeakyReLU(0.2), BatchNorm2d(Out$_{c}$)   & 2 x Out$_{c}$ \\
    
\midrule
    
    \multirow{4}{*}{EncoderBlock(In$_{p}$, In$_{c}$, Out$_{c}$)}  & 
    ConvBlock(In$_{p}$, In$_{p}$)   &      \\
    % & LeakyReLU(0.2), BatchNorm2d(In$_{p}$)   & 2 x In$_{p}$ \\
    & ConvBlock(In$_{c} $ + In$_{p}$, Out$_{c}$)   &    \\
    % & LeakyReLU(0.2), BatchNorm2d(Out$_{c}$)   & 2 x Out$_{c}$ \\
    & Conv2d(In$_{c} $+ Out$_{c}$, Out$_{c}$)   &    \\
    % & LeakyReLU(0.2), BatchNorm2d(Out$_{c})$   & 2 x Out$_{c}$ \\
    & Conv2d(In$_{c} $+ 2 $\times$ Out$_{c}$, Out$_{c}$)   &  \\
    & Conv2d(In$_{c} $+ 3 $\times$ Out$_{c}$, Out$_{c}$)   &  \\
    % & LeakyReLU(0.2), BatchNorm2d(Out$_{c})$   & 2 x Out$_{c}$ \\

\midrule

\multirow{8}{*}{DecoderBlock(In$_{p}$, In$_{c}$, Out$_{c}$)}  & 
    DeConvBlock(In$_{p}$, In$_{p}$, 1)   & \\
    % & LeakyReLU(0.2), BatchNorm2d(In$_{p}$)   & 2 $\times$ In$_{p}$ \\
    & ConvBlock(In$_{c}$+In$_{p}$, Out$_{c}$, 1)   &    \\
    % & LeakyReLU(0.2), BatchNorm2d(Out$_{c}$)   & 2 $\times$ Out$_{c}$ \\
    & DeConvBlock(In$_{p}$, In$_{p}$, 1)   &  \\
    % & LeakyReLU(0.2), BatchNorm2d(In$_{p}$)   & 2 $\times$ In$_{p}$ \\
    & ConvBlock(In$_{c}$ + In$_{p}$ + Out$_{c}$, Out$_{c}$, 1)   & \\
    % & LeakyReLU(0.2), BatchNorm2d(Out$_{c}$)   & 2 x Out$_{c}$ \\
    & DeConvBlock(In$_{p}$, In$_{p}$, 1)   &  \\
    % & LeakyReLU(0.2), BatchNorm2d(In$_{p}$)   & 2 $\times$ In$_{p}$ \\
    & ConvBlock(In$_{c}$ + In$_{p}$ + 2$\times$ Out$_{c}$, Out$_{c}$, 1)   & \\
    & DeConvBlock(In$_{c}$ + 2 $\times$ Out$_{c}$, Out$_{c}$, 1)   &     \\
    % & LeakyReLU(0.2), BatchNorm2d(In$_{c}$)   & 2 x In$_{c}$ \\
    
\midrule

\multirow{18}{*}{Generator(s$_z$)}  & 
    ConvBlock(1, 64, 2)   &  \multirow{17}{*}{10,567,811}  \\
    % & LeakyReLU(0.2), BatchNorm2d(64)  \\
    & EncoderBlock(1, 64, 64)  \\
    & EncoderBlock(64, 64, 128)  \\
    & EncoderBlock(128, 128, 128)  \\
    & Linear(s$_z$, 4$\times$4$\times$8) \\
    & DecoderBlock(0, 136, 64)\\
    & Linear(s$_z$, 7$\times$7$\times$4)  \\
    & DecoderBlock(128, 260, 64) &\\
    & Linear(s$_z$, 13$\times$13$\times$2) \\
    & DecoderBlock(128, 194, 64) \\
    & DecoderBlock(64, 128, 64)  \\
    & DecoderBlock(64, 65, 64) \\
    & ConvBlock(64, 64, 1)   & \\
    % & LeakyReLU(0.2), BatchNorm2d(64)   & \\
    & ConvBlock(64, 64, 1)   & \\
    % & LeakyReLU(0.2), BatchNorm2d(64)\\
    & Conv2d(64, 1, 3, stride=1, padding=1)    \\
    
    \bottomrule
\end{tabular}
\label{SI:table_param_DAGAN-UN}
\end{table}

\newpage
\subsection{\DAGANRN{}{} samples on QuickDraw-FS}
\begin{figure}[h!]
\centering
\includegraphics[width=0.8\textwidth]{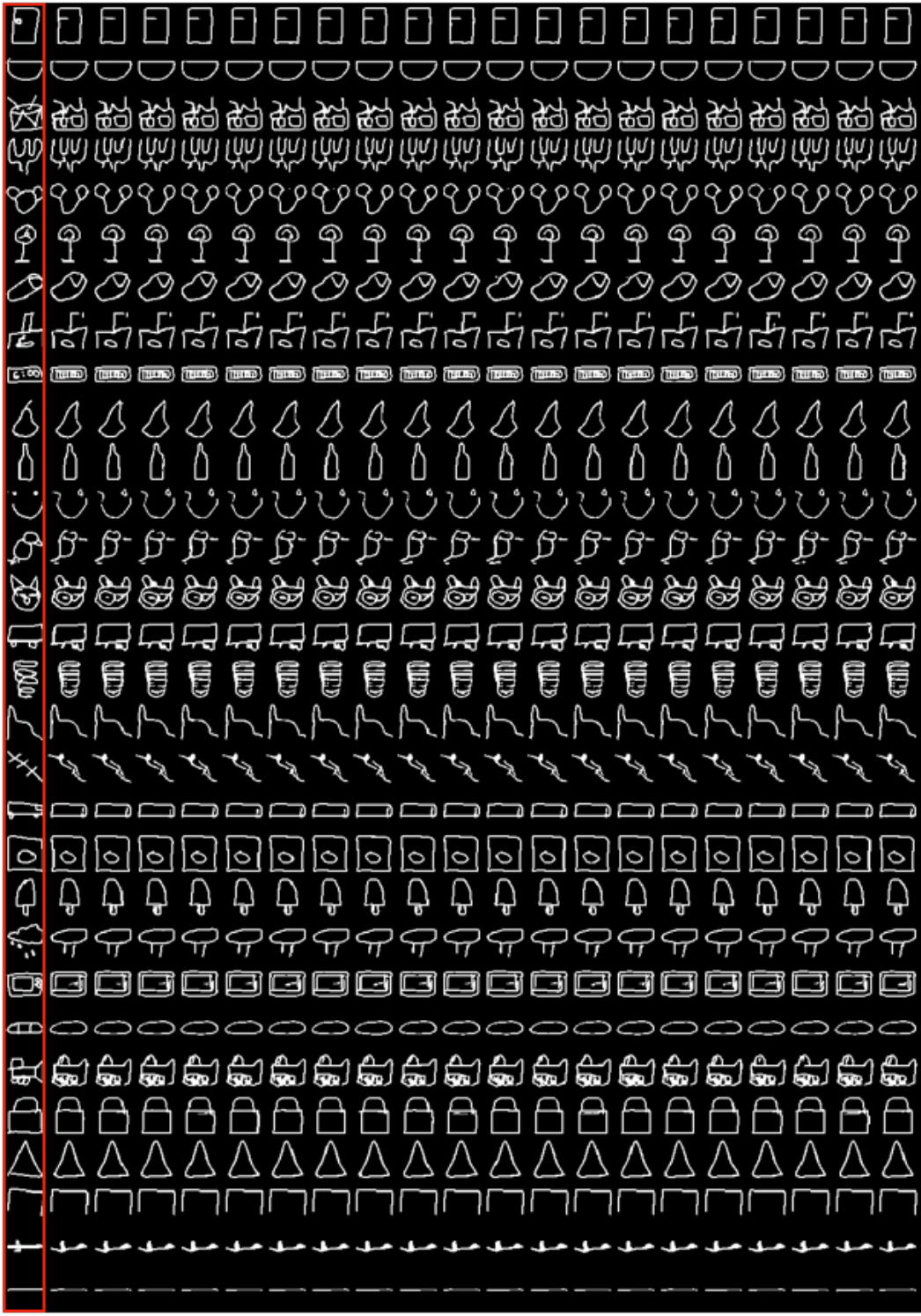}
%\vspace{-20pt}
         \caption{Samples generated by the \DAGANRN{} on QuickDraw-FS. All the exemplars used to condition the generative model are in the red frame. The $30$ concepts have been randomly sampled (out of $115$ concepts) from the QuickDraw-FS test set. Each line is composed of $20$ \DAGANRN{} samples that represent the same visual concept.}
         \label{SI:DAGAN_RN_samples_q}
\end{figure}

\newpage
\subsection{\DAGANUN{}{} samples on QuickDraw-FS}
\begin{figure}[h!]
\centering
\includegraphics[width=0.8\textwidth]{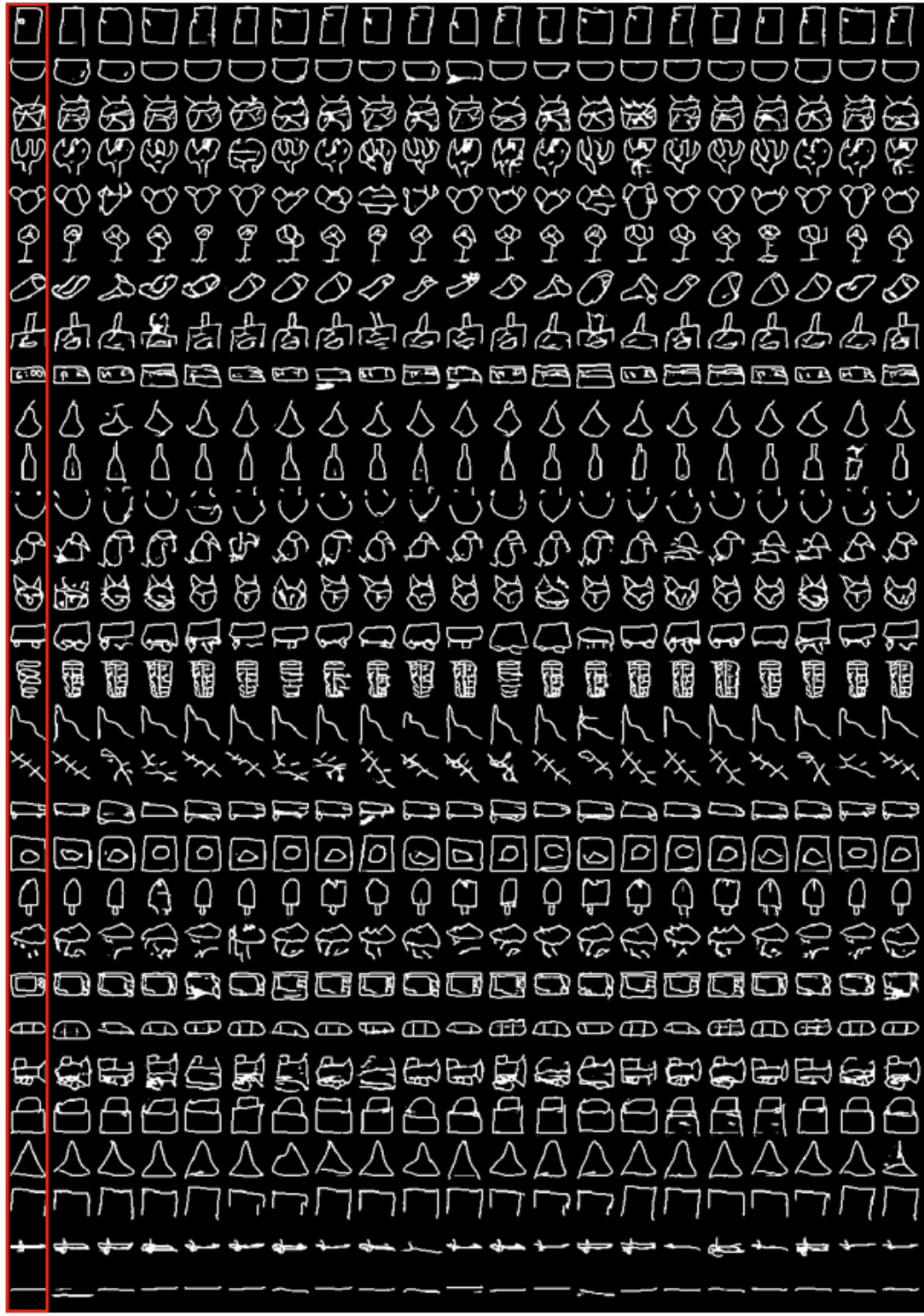}
%\vspace{-20pt}
         \caption{Samples generated by the \DAGANUN{} on QuickDraw-FS. All the exemplars used to condition the generative model are in the red frame. The $30$ concepts have been randomly sampled (out of $115$ concepts) from the QuickDraw-FS test set. Each line is composed of $20$ \DAGANUN{} samples that represent the same visual concept.}
         \label{SI:DAGAN_UN_samples_q}
\end{figure}

%\begin{itemize}
% 
%\end{itemize}

\newpage

\section{Details on the \FSDM{} trained on QuickDraw-FS}
\label{sup:FSDM_qd}
\subsection{Conditioning}
Unlike \FSDM{} trained on Omniglot, for QuickDraw-FS the context net we use to extract context from the set of samples is an sViT similar to \citealp{seung2021svit}. Furthermore, unlike \FSDM{} trained on Omniglot, the context is not calculated on the entire set of images, but rather we utilize a per-patch aggregation, wherein we take a mean value of each patch over the entire set and use that to generate the context vector. We can process any sample size using per-patch aggregation without increasing the number of tokens needed to condition the U-Net, and more crucially, we are able to composite information from various different samples simultaneously.

The conditioning mechanism to combine the context with the generator U-Net is the same as that in \FSDM{} trained on Omniglot (see \cref{sup:FSDM_omniglot}).

%\textcolor{red}{@Rishav: You should explain the conditioning mechanism of the VIT-set version of the FSDM. }
%\citealp{giannone2022few} put forward a framework to leverage conditional DDPMs for few-shot generation. The generative process in FSDM is conditioned on a small set of images from a given class by aggregating image patch information using a set-based Vision Transformer (ViT). FSDM can also use a U-Net based  architecture to extract context.

%\citealp{giannone2022few} proposed two mechanisms to fuse the context into the generation process; a mechanism based on \citealp{ethan2018film} and Learnable Attentive Conditioning (LAC), inspired by \citealp{robin2021highresimgsynth}. Our experiments are carried out on the FiLM-based mechanism.

\subsection{Architecture}

As mentioned in \cref{sup:FSDM_omniglot}, for the backbone, \FSDM{} utilizes the same architecture as the \DDPM{}. 

The context net utilized for QuickDraw-FS as mentioned above is an sViT similar to \citealp{seung2021svit} where we handle small sets (1-10) of images. The architecture is described in \cref{SI:table_sViT}.

The size of the model is 12.9 million parameters out of which 5.7 million parameters are for the encoder and 7.2 million are for the generative model.

\begin{table}[h!]
  \caption{Description of the sViT architecture}
  \centering
  \begin{tabular}{ccc}
    \toprule
    Network & Layer & \# params \\
    \midrule
    \multirow{2}{*}{SPT(patch\_dim, dim)} & LayernNorm(patch\_dim)  & \\
    & Linear(patch\_dim, dim)  & \\
    \midrule
    \multirow{2}{*}{PreNorm(dim, fn)} & 
    LayerNorm(dim) &  \\ 
    & \textcolor{blue}{fn} &\\
    \midrule
    \multirow{5}{*}{FeedForward(dim, hidden\_dim)} & Linear(dim, hidden\_dim)    & \\
    & GeLU  & $0$\\
    & Dropout() & $0$\\
    & Linear(hidden\_dim, dim)  &  \\
    & Dropout() & $0$\\
    \midrule
    \multirow{5}{*}{LSA(dim, heads, dim\_head)} & 
    Softmax(dim=-1)& $0$\\
    & Dropout()  &  $0$\\
    & Linear(dim, dim\_head $\times$  heads $\times3$, bias = $False$)  & \\
    & Linear(dim\_head $\times$  heads, dim)  &   \\
    & Dropout()  & $0$\\
    \midrule
    \multirow{15}{*}{sViT} & Linear(128,256) & $33.0$ K\\
    & SPT(patch\_dim = $320$, dim=$256$) &  $82.8$ K\\
    & PreNorm(dim = $256$, fn = LSA(dim = $256$, heads = $12$, dim\_head = $64$) & $787.2$ K\\
    & PreNorm(dim = $256$, fn = FeedForward(dim = $256$, hidden\_dim = $256$) & $132.1$ K\\
    & PreNorm(dim = $256$, fn = LSA(dim = $256$, heads = $12$, dim\_head = $64$) & $787.2$ K\\
    & PreNorm(dim = $256$, fn = FeedForward(dim = $256$, hidden\_dim = $256$) & $132.1$ K\\
    & PreNorm(dim = $256$, fn = LSA(dim = $256$, heads = $12$, dim\_head = $64$) & $787.2$ K\\
    & PreNorm(dim = $256$, fn = FeedForward(dim = $256$, hidden\_dim = $256$) & $132.1$ K\\
    & PreNorm(dim = $256$, fn = LSA(dim = $256$, heads = $12$, dim\_head = $64$) & $787.2$ K\\
    & PreNorm(dim = $256$, fn = FeedForward(dim = $256$, hidden\_dim = $256$) & $132.1$ K\\
    & PreNorm(dim = $256$, fn = LSA(dim = $256$, heads = $12$, dim\_head = $64$) & $787.2$ K\\
    & PreNorm(dim = $256$, fn = FeedForward(dim = $256$, hidden\_dim = $256$) & $132.1$ K\\
    & PreNorm(dim = $256$, fn = LSA(dim = $256$, heads = $12$, dim\_head = $64$) & $787.2$ K\\
    & PreNorm(dim = $256$, fn = FeedForward(dim = $256$, hidden\_dim = $256$) & $132.1$ K\\
    & mean(dim=1) & $0$\\
    & LayerNorm(256) & $512$\\
    & Linear(256,256) & $65.8$ K\\
 \bottomrule
    \\
    \multicolumn{3}{l}{Note : \textcolor{blue}{Blue} layers represent variable layers dependent on a certain parameter}
    
  \end{tabular}
\label{SI:table_sViT}
\end{table}

\subsection{Training details}
The attention resolution, learning rate, optimizer, and batch size we use are the same as that implemented in  \url{https://github.com/georgosgeorgos/few-shot-diffusion-models}. The size of the context channels and hidden dimensions is set to 256. 

We train the model for 200 epochs.
%\textcolor{red}{@Rishav: Here you should describe the training details of the network (learning rate, batch size...). If all these hyperparameters are similar to those in the original repo, don't need to repeat, just specify it. Also you should refer to the original repo we have used, using \url{https://XXX.xxx}.}

\subsection{Explored hyper-parameters} 
To obtain the scatter plot in \cref{fig:fig1a}, we varied the following hyper-parameters:
\begin{itemize}
\item The sample size, ranging from 3 to  6 with steps of 1 (4 values overall) 
\item The number of timesteps or diffusion steps ranging from 200 to 100 with steps of 200 (5 values overall)
\item The number of residual blocks per downsample, ranging from 3 to 6 (4 values overall)
\end{itemize}

We have trained 80 different \FSDM{} models and plotted the diversity and the accuracy in \cref{fig:fig1a}.

\clearpage
\newpage
\subsection{\FSDM{}{} samples on QuickDraw-FS}
\begin{figure}[h!]
\centering
\includegraphics[width=0.8\textwidth]{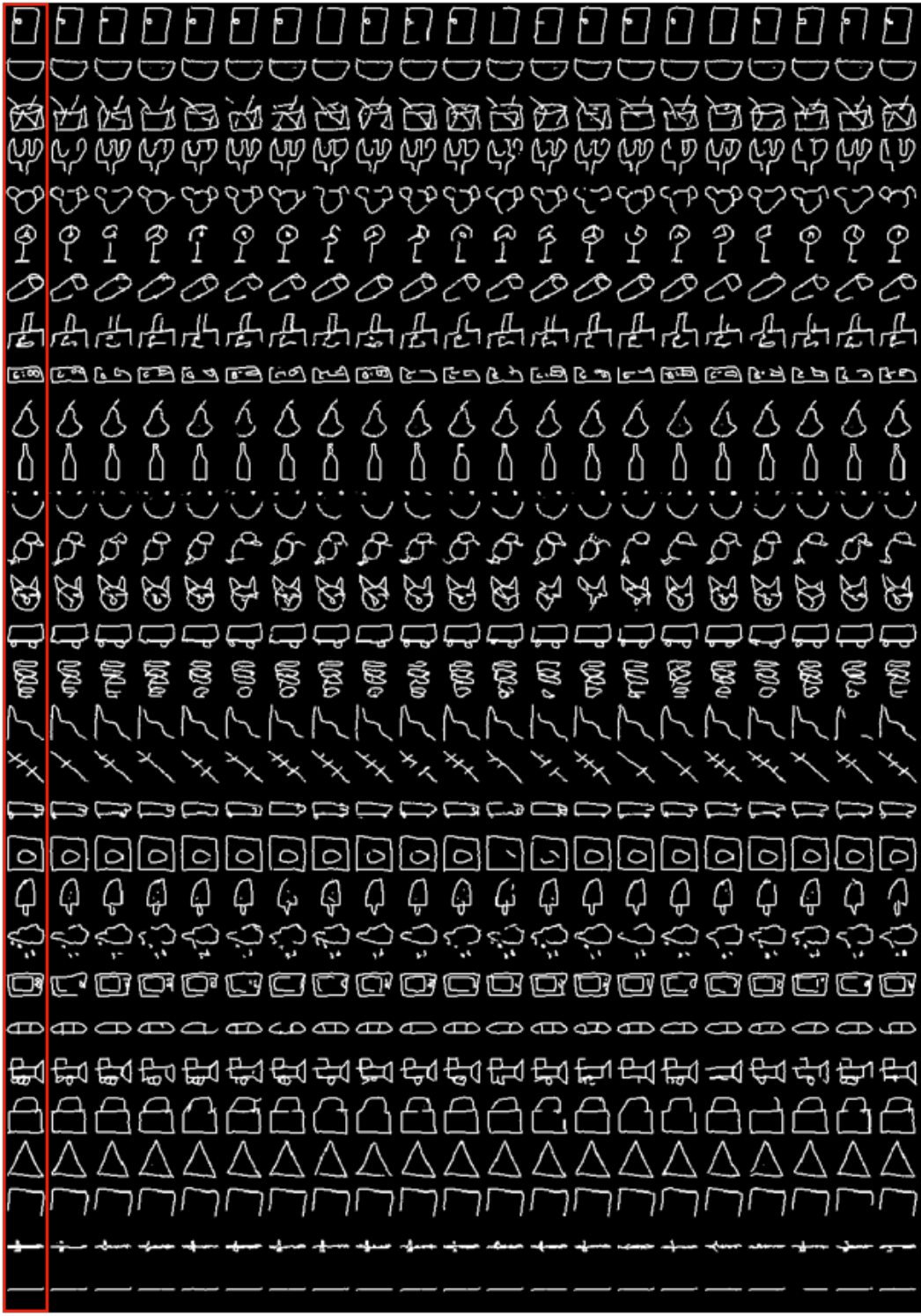}
%\vspace{-20pt}
         \caption{Samples generated by the \FSDM{}{} on QuickDraw-FS. All the exemplars used to condition the generative model are in the red frame. The $30$ concepts have been randomly sampled (out of $115$ concepts) from the QuickDraw-FS test set. Each line is composed of $20$ \FSDM{}{} samples that represent the same visual concept.}
         \label{SI:FSDM_samples_q}
\end{figure}

\clearpage
\newpage
\section{More details on the \textit{originality} metric}
\label{SI:originality_metric}
\begin{figure*}[h!]
%\vskip 0.2in
\begin{center}

\begin{tikzpicture}
\draw [anchor=north west] (0\linewidth, 1\linewidth) node {\includegraphics[width=0.28\linewidth]{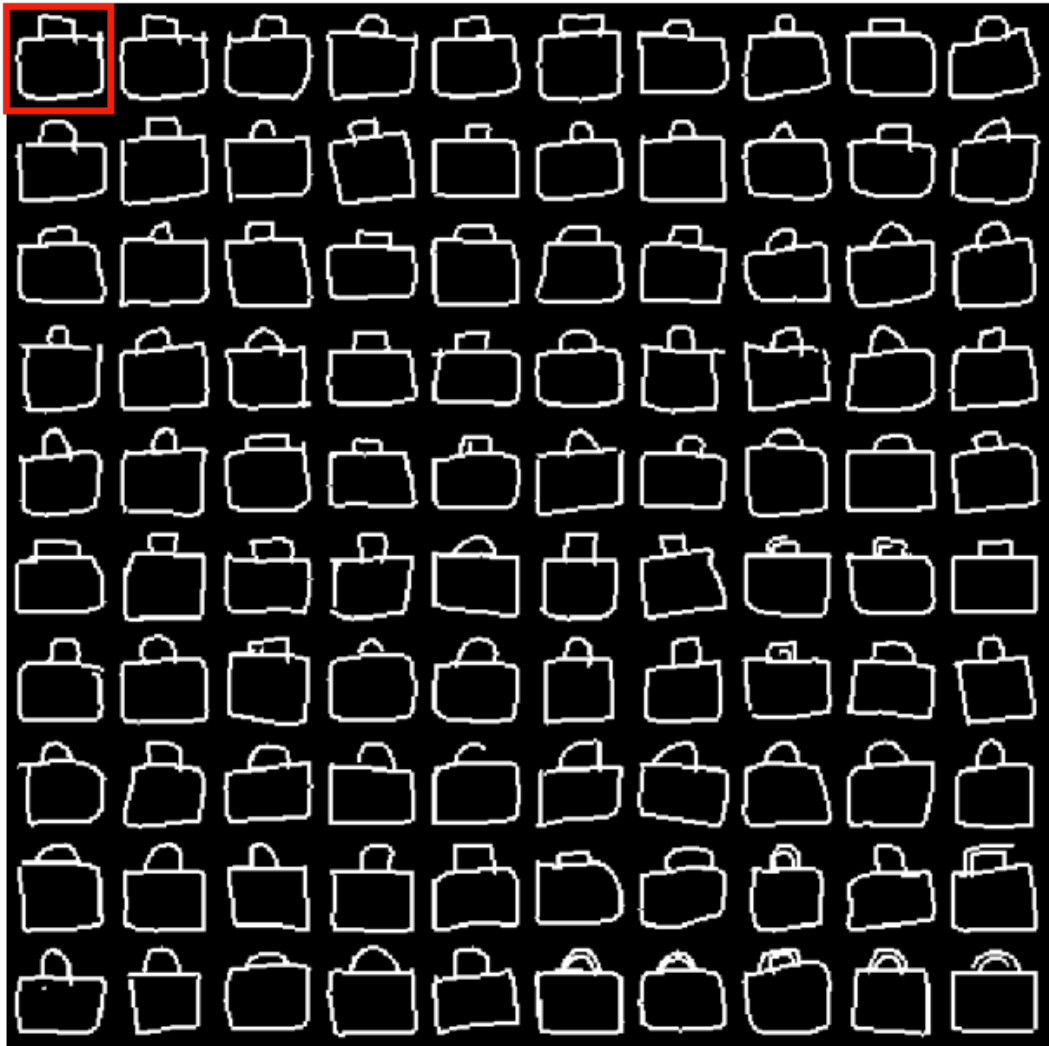}};
\draw [anchor=north west] (0.33\linewidth, 1\linewidth) node {\includegraphics[width=0.28\linewidth]{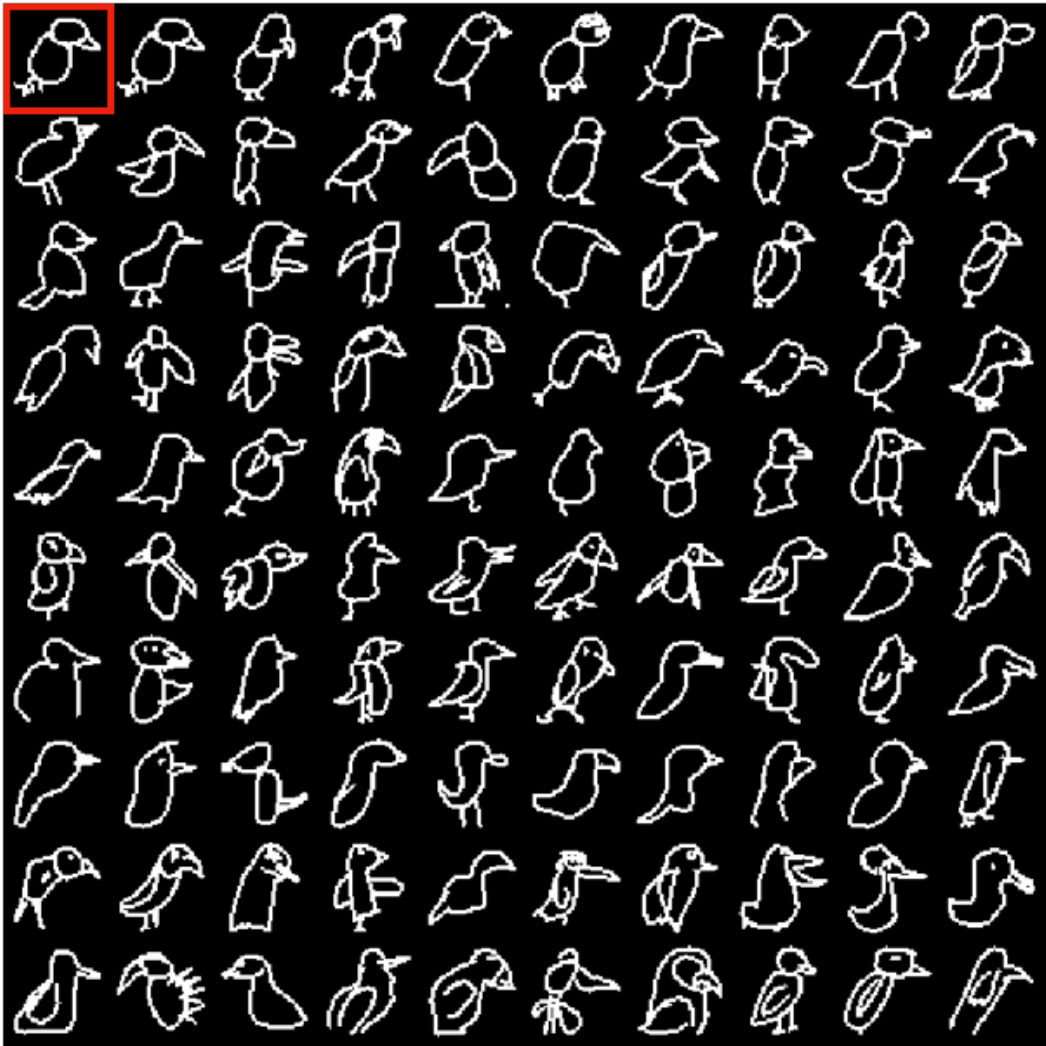}};
\draw [anchor=north west] (0.66\linewidth, 1\linewidth) node {\includegraphics[width=0.28\linewidth]{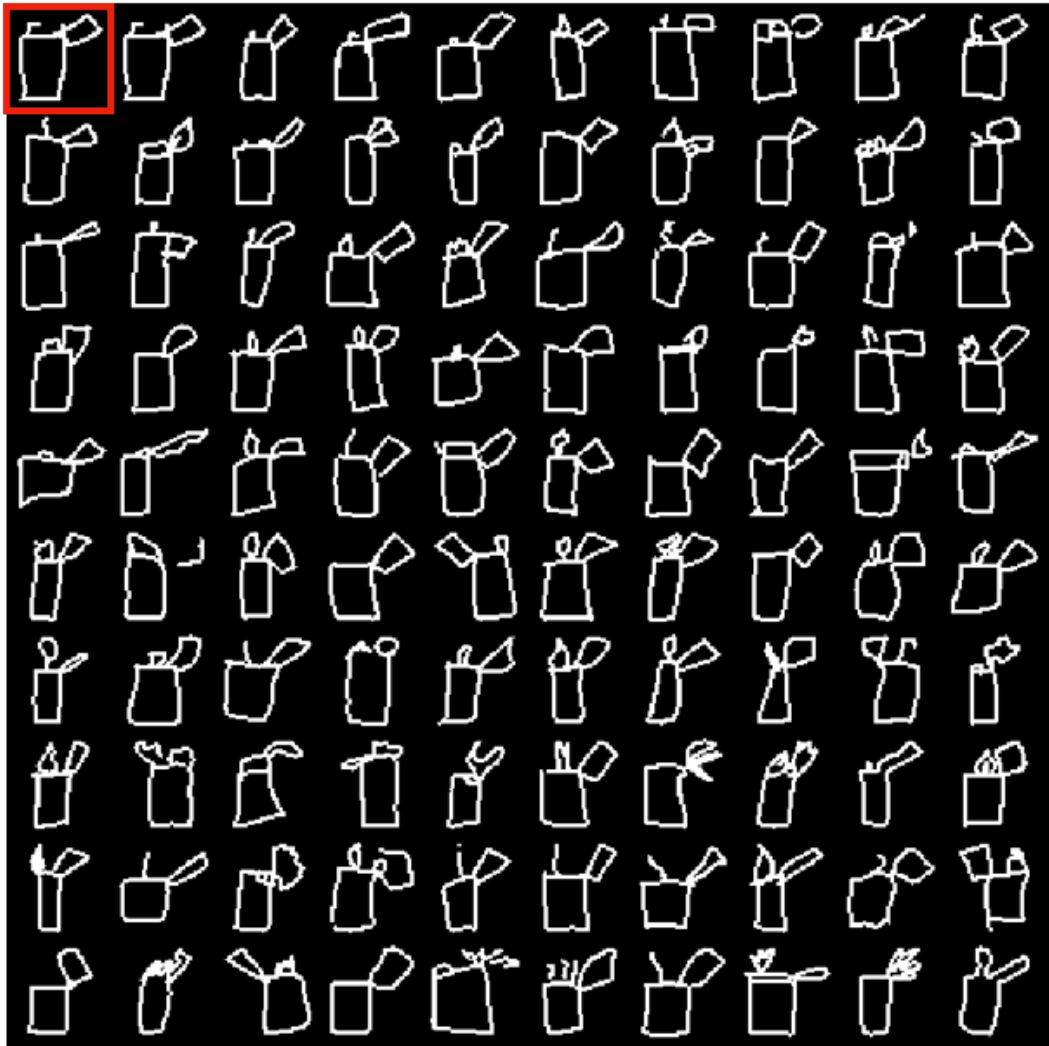}};
\draw [anchor=north west] (0\linewidth, 0.715\linewidth) node {\includegraphics[width=0.28\linewidth]{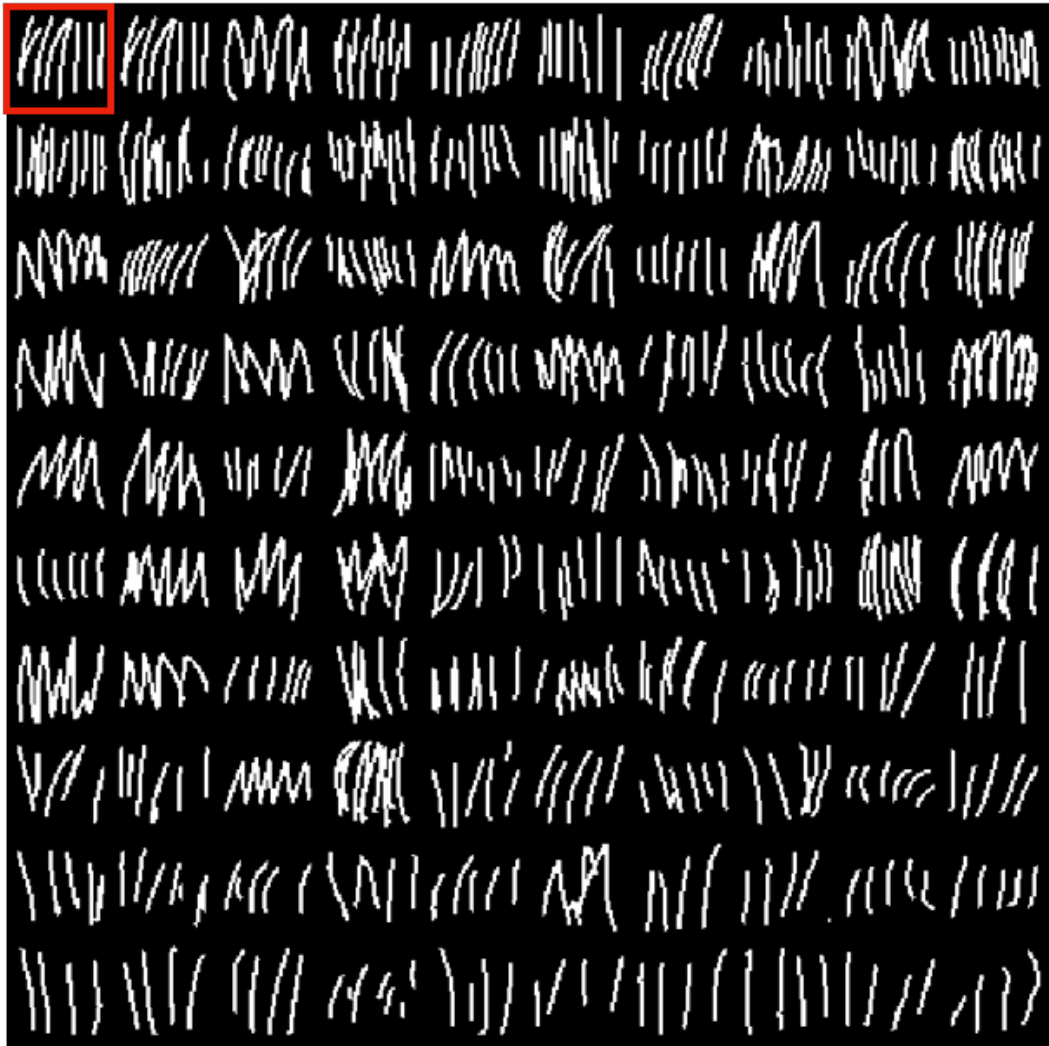}};
\draw [anchor=north west] (0.33\linewidth, 0.715\linewidth) node {\includegraphics[width=0.28\linewidth]{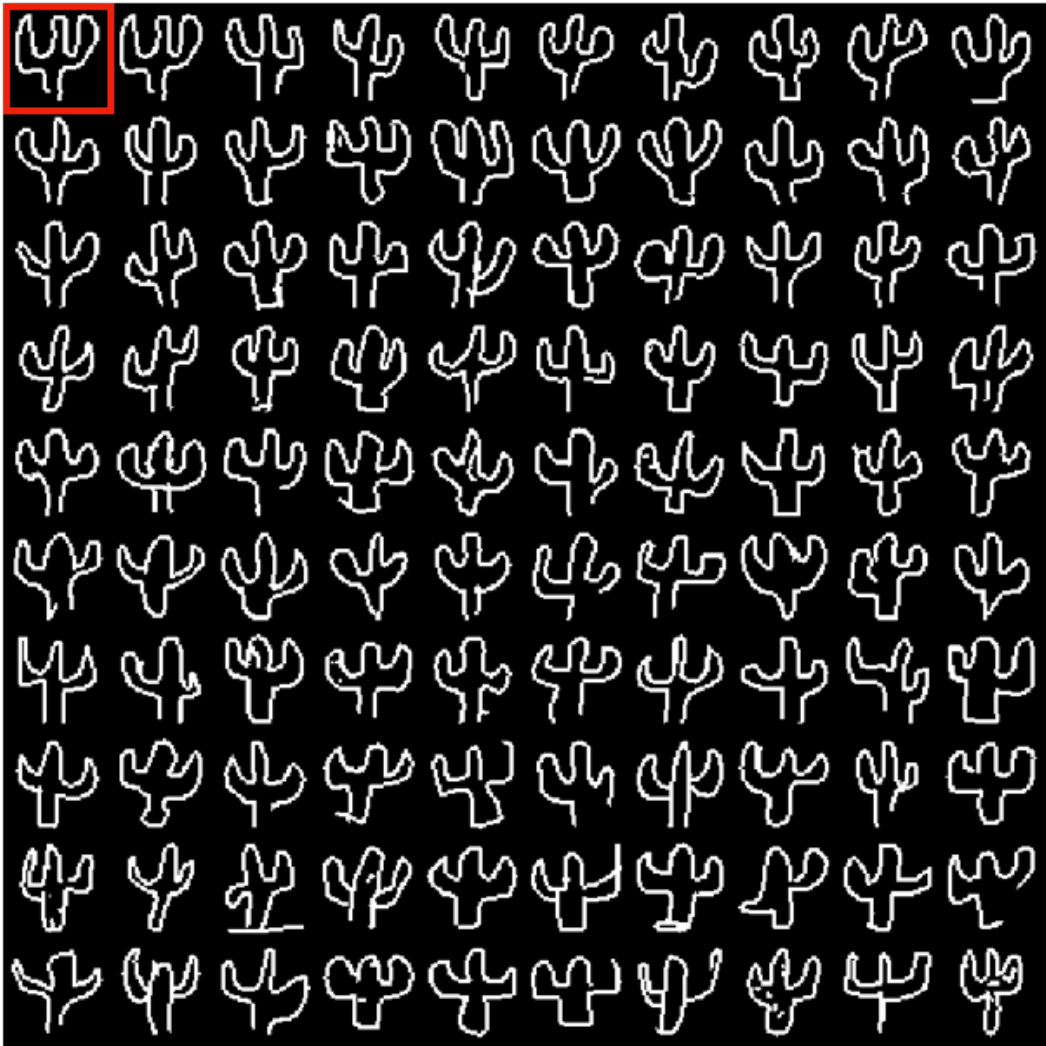}};
\draw [anchor=north west] (0.66\linewidth, 0.715\linewidth) node {\includegraphics[width=0.28\linewidth]{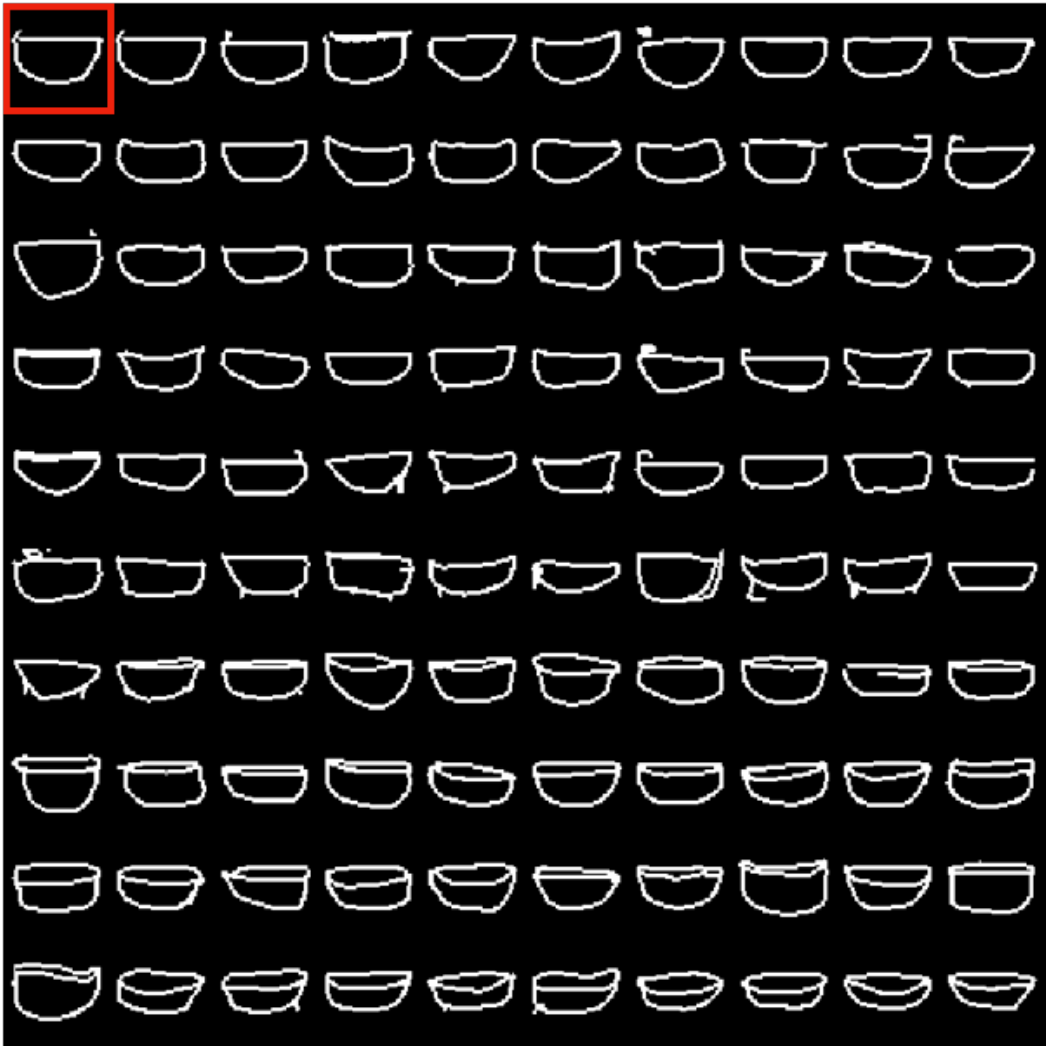}};
\end{tikzpicture}
\vspace{-16pt}
\caption{Randomly picked samples of the QuickDraw-FS test set. The samples are sorted (ascending order) according to their distance from the exemplar using the \textit{originality} metric (SimCLR feature extractor + $\ell_2$-norm). The exemplars are highlighted with a red square.}
\label{fig_sup:Samples_sorted_originality}
\end{center}
%\vskip -0.5in
\end{figure*}

To compute the \textit{originality}, we use the $\ell_2$ distance, in the SimCLR latent space, between the exemplar and the samples. We validate the originality metric by comparing it with alternative metrics in which we vary the feature extractor network and the distance metric. As an alternative to the SimCLR network, we consider the Prototypical Net~\citep{snell2017prototypical}. For the metric used to compute the distance between the samples and their corresponding exemplars, we have considered the cosine distance. For a given category j, composed with samples $v_{i}^{j}$ and exemplars $e^{j}$ we define the cosine distance in \cref{SI:Cosine_distance}. In this equation, $f$ denotes the feature extractor network.
\begin{align}
d(v_{i}^{j}, e^{j}) = \sqrt{2-2C(f(v_{i}^{j}), f(e^{j}))} \quad \text{s.t.} \quad \text{and} \quad C(u,v) = \frac{u \cdot v}{\rVert u \lVert \rVert v \lVert} 
\label{SI:Cosine_distance}
\end{align}
In \cref{SI_table:originality_metric}, we have computed the Spearman rank-order correlation between all possible combinations of distance metrics and feature extractor networks: 
\begin{table}[h!]
  \caption{Spearman rank-order correlation for different settings}
  \label{SI_table:originality_metric}
  \centering
  \begin{tabular}{cccc}
    \toprule
    Setting 1     & Setting 2     & Spearman correlation & p-value \\
    \midrule
     Proto. Net + $\ell_2$ distance  & Proto. Net + cosine distance  & $0.97$ & $4.23.\times10^{-41}$\\
     Proto. Net + $\ell_2$ distance  &  SimCLR + $\ell_2$ distance & $0.62$ & $1.14\times10^{-34}$ \\
     SimCLR + $\ell_2$ distance  & SimCLR + cosine distance  & $0.85$ & $8.2\times10^{-12}$ \\
     Proto. Net + cosine  & SimCLR + cosine distance  & $0.66$ & $1.02\times10^{-21}$ \\
    \bottomrule
  \end{tabular}
\end{table}
We observe that all combinations of feature extractor network + metrics have a strong correlation with each other ($\rho>0.5$) and this correlation is statistically significant ($p<1\times10^{-3}$). It suggests that the way we have defined distance to exemplar is compatible with the other definition. We prefer the SimCLR over the Prototypical feature extractor, because this is a fully unsupervised network~\citep{chen2020simple}, so it is more convenient to train (no need for labels). Similarly, we choose the $\ell_2$-norm to compute the distance between exemplars and samples because it is more natural. In \cref{fig_sup:Samples_sorted_originality}, we show some visual concepts, sorted by originality (in ascending order).

\newpage
\section{Link between originality and diversity}
\label{sup:creativity and diveristy}
We have defined diversity as the intra-class variability, computed with a standard deviation in the SimCLR feature space. By definition, the intra-class standard deviation is the square root of the mean squared distance between the center of the samples and the samples themselves. For a given category $j$, composed of N samples $v_{i}^{j}$ and a feature space f, the diversity $\sigma_j$ is defined as in \cref{eq:sup_diversity}.
\begin{align}
\sigma_{j} = \sqrt{\frac{1}{N-1}\sum_{i=1}^{N}\Big(f(v_{i}^{j}) -  \frac{1}{N}\sum_{i=1}^{N}f(v_{i}^{j})\Big)^2}
\label{eq:sup_diversity}
\end{align}

For a given sample, we have also defined the \textit{originality}. The originality is the $\ell_2$ distance, in the SimCLR feature space, between a sample and the corresponding category exemplar. In \cref{eq:sup_creativity}, we write down the formula of average category originality $c_j$, for a category $j$ represented by an exemplar $e^{j}$:

\begin{align}
c_{j} = \frac{1}{N} \sum_{i=1}^{N} c(v_{i}^{j}) \quad \text{s.t.} \quad c(v_{i}^{j}) = \Big\rVert f(v_{i}^{j}) - f(e^{j})\Big\lVert_{2}
%\sqrt{\frac{1}{N-1}\sum_{i=1}^{N}\Big(f(v_{i}^{j}) -  \frac{1}{N}\sum_{i=1}^{N}f(v_{i}^{j})\Big)^2}
\label{eq:sup_creativity}
\end{align}

\begin{figure}[ht]
\vskip 0.2in
\begin{subfigure}{0.5\textwidth}
\includegraphics[width=\columnwidth]{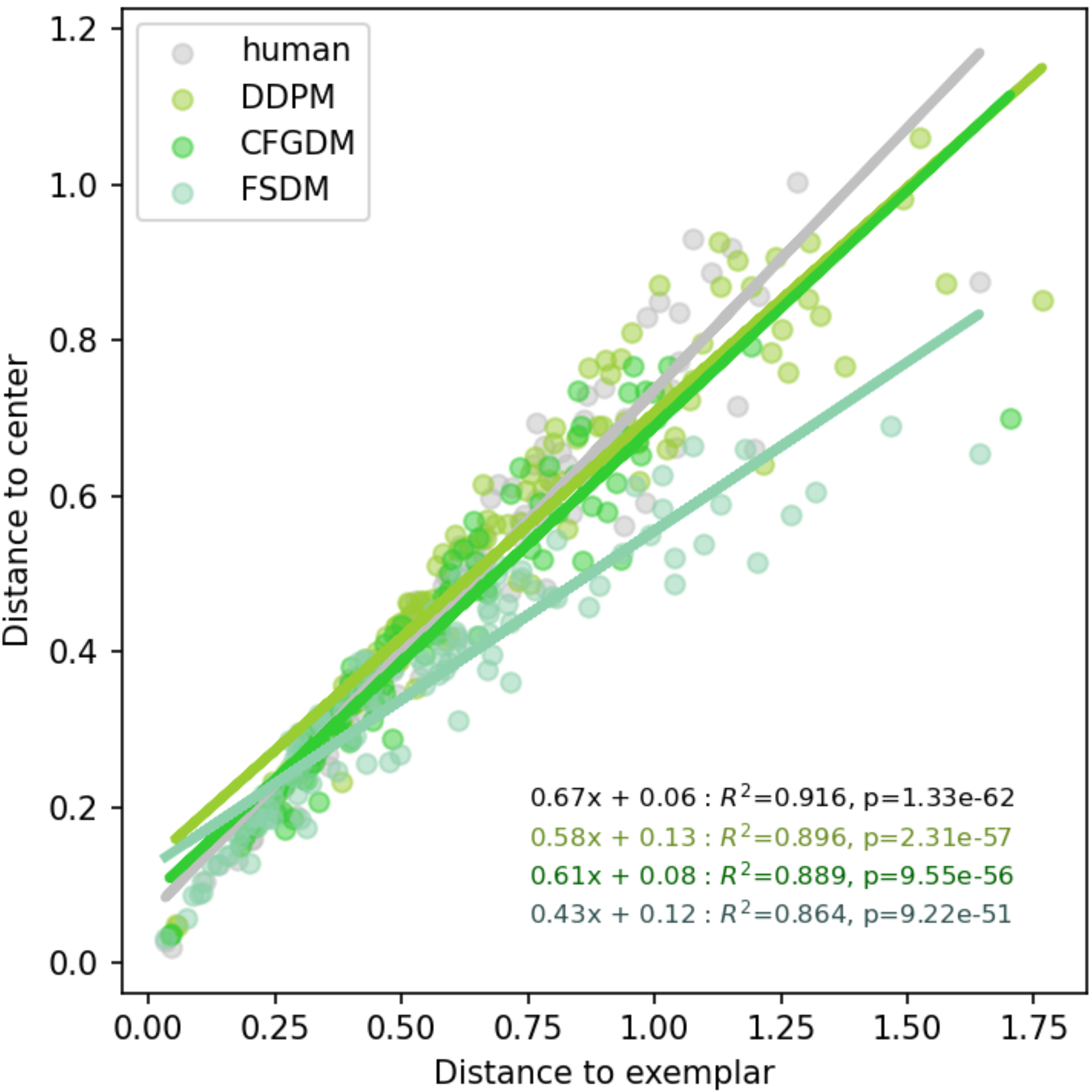}
\caption{averaged over object category}
\label{fig:crea_diver1}
\end{subfigure}  
\begin{subfigure}{0.5\textwidth}
\includegraphics[width=\columnwidth]{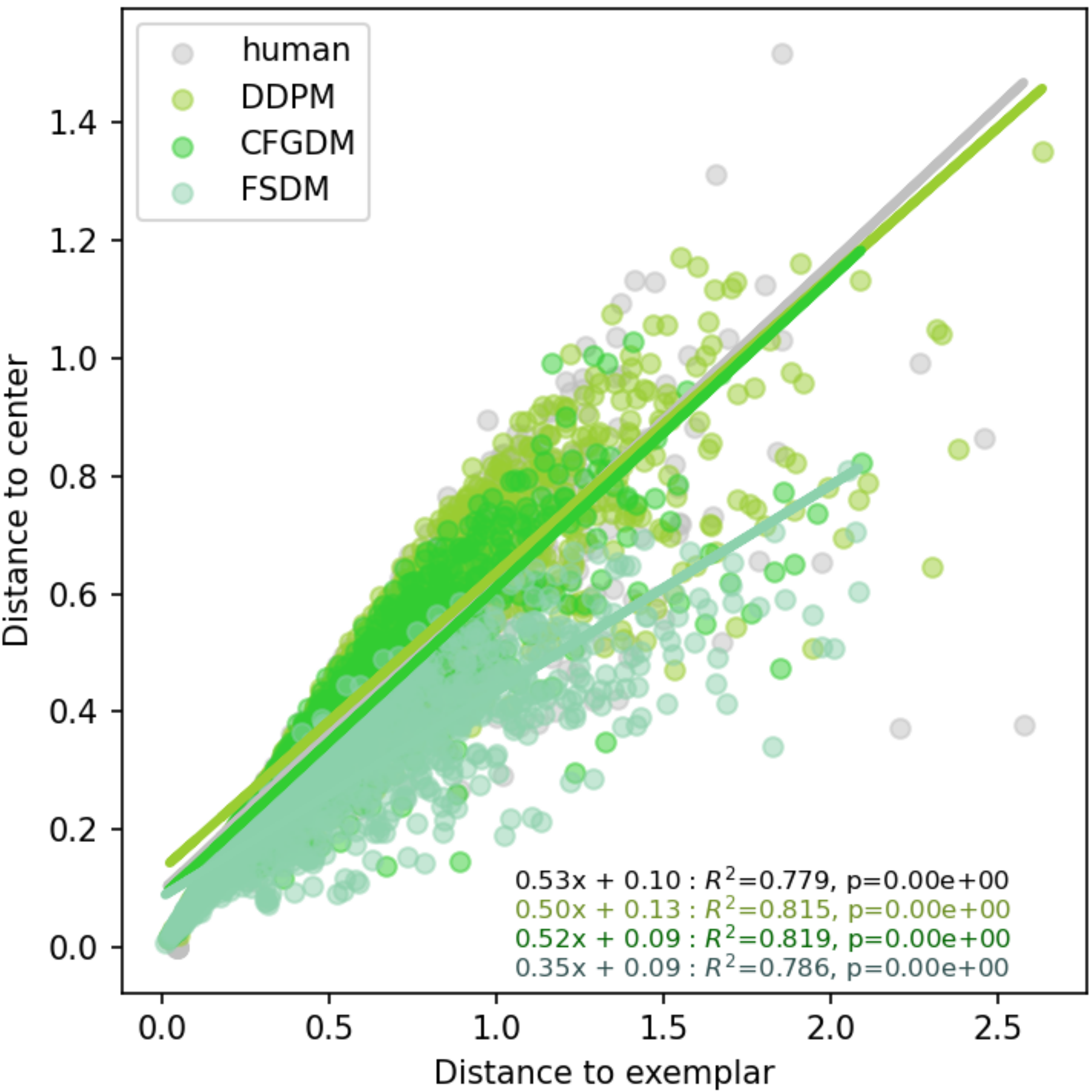}
\caption{averaged over originality bins}
\label{fig:crea_diver2}
\end{subfigure}
\caption{Scatter plot of the distances to the center as a function of the distances to exemplar for the \humans{}, the \CFGDM{}, the \FSDM{} and the \DDPM{}. \textbf{(a)}: is averaged over object category, and \textbf{(b)}: is averaged over originality bins.} 
\vskip -0.2in
\end{figure}

The QuickDraw-FS dataset is built such that the category exemplar is located as closely as possible to the center of the category cluster. We plot in \cref{fig:crea_diver1}, the average distance to the center as a function of the average distance to exemplar for each class and for \human{}, the \CFGDM{}, the \FSDM{} and the \DDPM{}. We observe a linear relationship between both distances (the lowest $R^2$ is $0.86$). We also observe that the slope of the linear regression is not equal to $1$ ($\approx 0.67$ for \human{}, $\approx0.61 $ for \CFGDM{}, $\approx 0.58$ for \DDPM{} and $\approx 0.43$ for \FSDM{}). It suggests that there is not an exact match between the center of the category and the exemplar. We observe similar behavior in \cref{fig:crea_diver2}, in which the distance value is averaged over the originality bins.

%\begin{figure}[ht]
%\begin{center}
%\centerline{
%\includegraphics[width=0.5\columnwidth]%{Figure/Supplementary/Fig_Crea_Div/Fig_Crea_Div.pdf}
%}

%\caption{ Scatter plot of the distance to center as a function of the distance to exemplar, averaged over each category, for the samples generated by \humans{}, the \CFGDM{}, the \FSDM{} and the \DDPM{}.}
%\label{fig:crea_diver}
%\end{center}
%\vskip -0.1in
%\end{figure}

\newpage
\section{Interpolation in the \textit{generalization curves}}
\label{sup:interp_error}
To obtain the data points of the generalization curves we compute the average originality and recognizability for all originality bins (see~\cref{generaliation_curves} for more information on the originality bins). We then interpolate between the data points using polynomial regression (degree 2). The fit of the polynomial curve is made using a least square error method. We report in~\cref{SI:error_generalization}.

\begin{table}[h!]
  \caption{regression errors of the generalization curves}
  \centering
  \begin{tabular}{ccc}
    \toprule
    Network & $\gamma$ & Least Square Error \\
    \midrule
    \human{} & - & $4.1\times10^{-7}$\\
    \FSDM{} & - & $9.4\times10^{-5}$\\
    \DDPM{} & $0$ & $3.7\times10^{-5}$\\
    \CFGDM{} & $1$ & $3.4\times10^{-6}$\\
    \midrule
    \CFGDM{} & $0.2$ & $3.7\times10^{-5}$ \\
    \CFGDM{} & $0.4$ & $2.2\times10^{-5}$ \\
    \CFGDM{} & $0.6$ & $1.3\times10^{-5}$ \\
    \CFGDM{} & $0.8$ & $7.3\times10^{-6}$\\
    \CFGDM{} & $1.5$ & $1.3\times10^{-6}$ \\
    \CFGDM{} & $2.0$ & $1.0\times10^{-6}$ \\

    \bottomrule
  \end{tabular}
\label{SI:error_generalization}
\end{table}

\newpage
\section{More \CFGDM{} importance feature maps}
\begin{figure}[h!]
\centering
\includegraphics[width=1\textwidth]{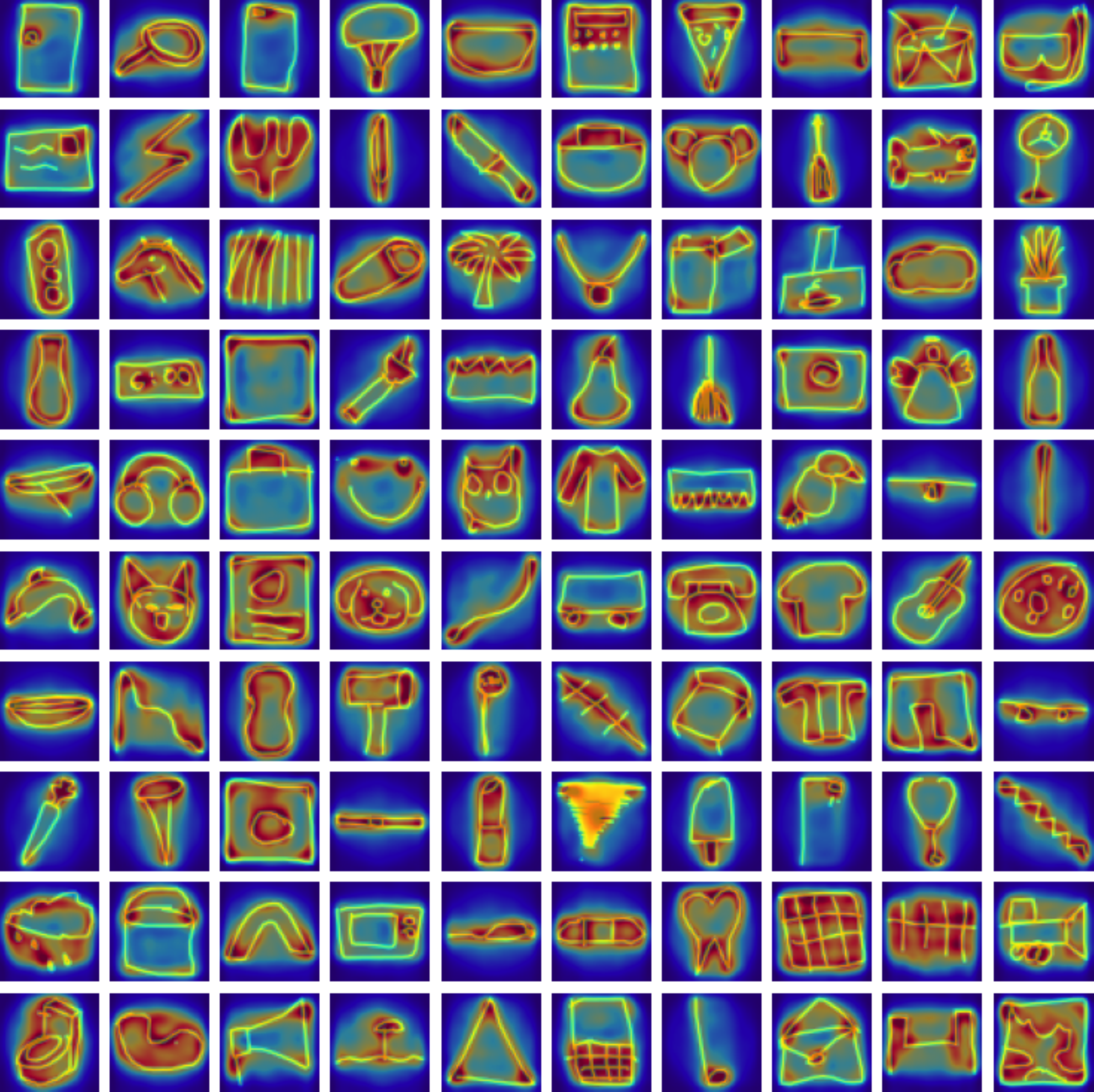}
%\vspace{-20pt}
         \caption{\CFGDM{} Importance feature map, for $100$ categories. The maps are obtained by averaging n=10 misalignment maps $\phi(\mathbf{x},\mathbf{y})$ as defined in \cref{eq:attribution_score}}
         \label{SI:additionalImportanceMap}
\end{figure}

Using the \cref{eq:attribution_score}, we have computed the features importance map for $100$ classes (see \cref{SI:additionalImportanceMap}). For each class, we have averaged the importance maps obtained for $10$ different samples generated by the \CFGDM{}. 

\newpage
\section{The ClickMe-QuickDraw Experimental Setup}
\label{SI:click}

\begin{figure}[h!]
\centering
\includegraphics[width=1\textwidth]{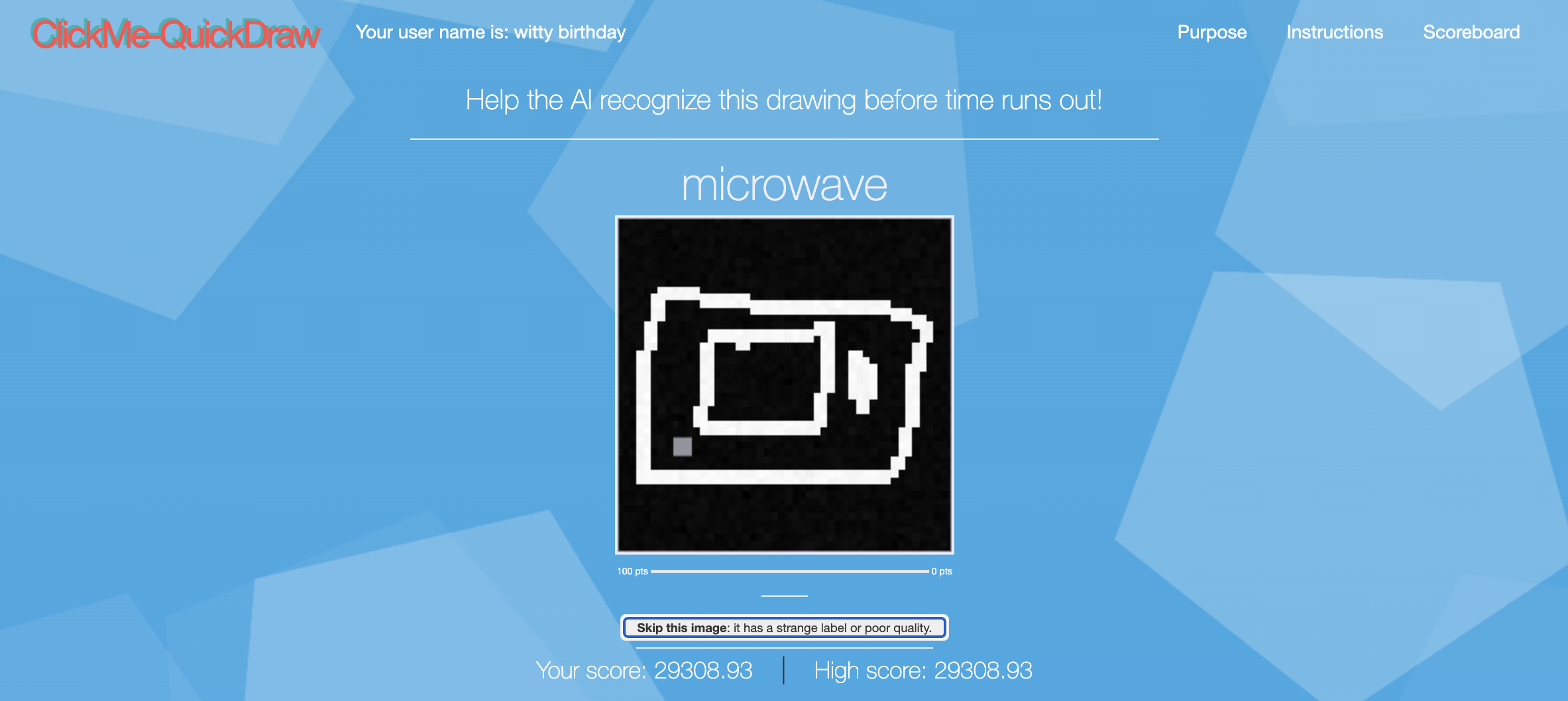}
%\vspace{-20pt}
         \caption{Screenshot of the ClickMe-QuickDraw web application}
         \label{SI:clickmeQuickdraw}
\end{figure}

ClickMe-QuickDraw is a web application on which the user (alongside an AI model) plays to win prizes and help us understand differences in how humans and machines perceive drawings. The goal of the game is to help the AI partner recognize as many object drawings with as much confidence as possible.

The user helps the AI model recognize objects by revealing parts of images to it. This is done by \textit{painting} over parts of the object image that help humans recognize it. When the user clicks on the image, the \textit{brush stroke} begins dragging the cursor over the other important parts of the image. These image parts will be revealed to the AI model as the user paints over them, which will try to classify the drawing based on the regions that were \textit{painted over}.

For every image guessed correctly, the user receives a score based on the time taken for the AI to recognize the image. The user will receive no points if the timer elapses before the AI model classifies correctly. The user can skip images that look strange or do not match their label by clicking the \textbf{Skip this image} button.

\subsection{Web Application Design Parameters}
Our ClickMe-QuickDraw experiment design followed the paradigm of ClickMe~\citealp{linsley2018learning}; images of object drawings are presented to the participant for 7 seconds before moving to the next image. We recruited 102 participants for this experiment, each instructed to annotate the most important parts of images of object drawings to complete the experiment. All the participants were informed of their participation in the experiment and the elementary drawing skill required. Participants were mostly selected among undergraduated students. The web app used for the experiment is built using Node.js and the production version of Python's Flask web server. As seen in \cref{SI:clickmeQuickdraw}, the user is presented with the correct class label and a $256\times256$ image of the drawing. The timer starts when the user clicks on the image and begins painting over the important parts of the image. As the user highlights parts of the image, the size of each click on the image is $21\times21$. The timer lasts for $5$ seconds, and if the user fails to highlight parts that help the AI correctly classify the drawing, the drawing is skipped, and another one is displayed.

\subsection{Classification Model}
The AI model that classifies the highlighted parts of each image is a Lipschitz-constrained network trained to classify drawing images. The robustness of the model was imperative since we are working with model-generated images from a single prototype. The Lipschitz-constrained network yielded a classification accuracy of $0.99$ on the entire database of images.

\subsection{Data}
The dataset for the web application contained $250$ images, with $25$ different classes and $10$ images per class. The experiment had 102 participants, with $1050$ correct annotations, giving us an average of $41.2$ annotations across all $25$ classes.

\subsection{Reliability Analysis}

\begin{figure}[h!]
\centering
\includegraphics[width=0.5\textwidth]{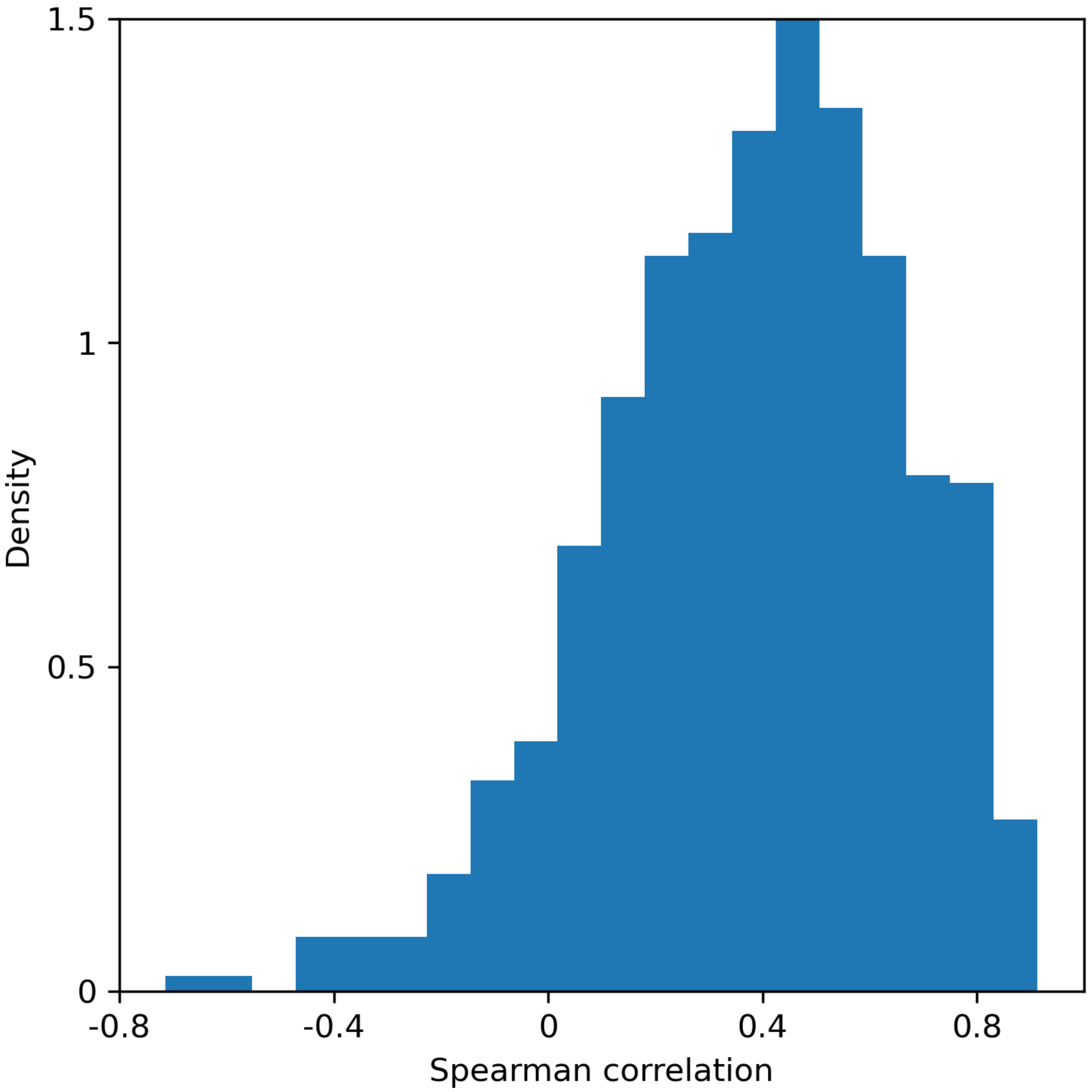}
%\vspace{-20pt}
         \caption{Distribution of the per-image rank-order Spearman correlation.}\label{SI:Reliability_analysis}
\end{figure}

We verify that the collected annotations (i.e., ClickMe maps) show a strong regularity and consistency across participants. We calculated the rank-order Spearman correlation between annotations from two randomly selected participants for an image. The annotations are blurred with a Gaussian kernel (of size $49\times49$). Such a blurring facilitates the comparison and reduces the noise related to the online application since the brush strokes are drawn with a computer mouse. We repeat this procedure $1\,0000$ times and average the per-image correlation. In \cref{SI:Reliability_analysis}, we plot the distribution of the per-image rank-order Spearman correlation.

We have filtered out the samples with a Spearman correlation $2$ standard deviations away from the mean, which corresponds to a p-value of 5\%. Such a filtering process allows us to remove inconsistent annotation maps. After removing the outliers, we have a mean per-image rank-order Spearman correlation of $0.47$ ($p<5e-2$). This is to be compared to the mean Spearman correlation between $2$ randomly selected annotations, which is $0.05$.
\end{document}